\documentclass{article} % For LaTeX2e
\usepackage{iclr2026_conference,times}

\usepackage{amsmath,amsfonts,bm}

\def\eqref#1{equation~\ref{#1}}
\def\1{\bm{1}}

\DeclareMathAlphabet{\mathsfit}{\encodingdefault}{\sfdefault}{m}{sl}
\SetMathAlphabet{\mathsfit}{bold}{\encodingdefault}{\sfdefault}{bx}{n}

\usepackage{url}
\usepackage{diagbox}
\usepackage{enumitem}
\usepackage{amsmath}
\usepackage{colortbl}
\usepackage{amssymb}
\usepackage{arydshln}
\usepackage[T1]{fontenc}    % use 8-bit T1 fonts
\usepackage{hyperref}       % hyperlinks

\usepackage{booktabs}       % professional-quality tables
\usepackage{amsfonts}       % blackboard math symbols
\usepackage{nicefrac}       % compact symbols for 1/2, etc.
\usepackage{microtype}      % microtypography
\usepackage{xcolor}         % colors  
\usepackage{graphicx}
\usepackage{multirow}
\usepackage{algpseudocode}
\usepackage{wrapfig}
\usepackage{subcaption}
\usepackage[ruled]{algorithm2e}
\usepackage{algorithmicx}

\usepackage{float}      % 提供 [H] 选项，强制固定位置
\usepackage{caption}    % 提供 \captionof 命令
\usepackage{makecell}
\usepackage{multicol}

\definecolor{xjwu}{RGB}{246, 145, 5}

\title{FASA: Frequency-Aware Sparse Attention}
\newcommand{\mymodel}{FASA\@\xspace}
\author{\textbf{Yifei Wang}$^{1}$
\quad
\textbf{Yueqi Wang}$^{2}$
\quad
\textbf{Zhenrui Yue}$^{3}$
\quad
\textbf{Huimin Zeng}$^{3}$
\quad
\textbf{Yong Wang}$^{1}$\thanks{Project lead and corresponding author.} \\
\textbf{Ismini Lourentzou}$^{3}$
\quad
\textbf{Zhengzhong Tu$^{4}$}
\quad
\textbf{Xiangxiang Chu$^{1}$} 
\quad
\textbf{Julian McAuley$^{2}$} \\
$^1$ AMAP, Alibaba Group
\quad
$^2$ UCSD
\quad
$^3$ UIUC
\quad
$^4$ Texas A\&M University
}
\iclrfinalcopy % Uncomment for camera-ready version, but NOT for submission.
\begin{document}

\maketitle

\begin{abstract}
The deployment of Large Language Models (LLMs) faces a critical bottleneck when handling lengthy inputs: the prohibitive memory footprint of the Key Value (KV) cache. To address this bottleneck, the token pruning paradigm leverages attention sparsity to selectively retain a small, critical subset of tokens. However, existing approaches fall short, with static methods risking irreversible information loss and dynamic strategies employing heuristics that insufficiently capture the query-dependent nature of token importance.
We propose \mymodel, a novel framework that achieves query-aware token eviction by dynamically predicting token importance.
\mymodel stems from a novel insight into RoPE: the discovery of functional sparsity at the frequency-chunk (FC) level. Our key finding is that a small, identifiable subset of "dominant" FCs consistently exhibits high contextual agreement with the full attention head. This provides a robust and computationally free proxy for identifying salient tokens.
%making them a powerful and efficient proxy for token importance.
Building on this insight, \mymodel first identifies a critical set of tokens using dominant FCs, and then performs focused attention computation solely on this pruned subset.
% Since accessing only a small fraction of the KV cache, \mymodel drastically lowers memory bandwidth requirements and computational cost.
Across a spectrum of long-context tasks, from sequence modeling to complex CoT reasoning, \mymodel consistently outperforms all token-eviction baselines and achieves near-oracle accuracy, demonstrating remarkable robustness even under constraint budgets. Notably, on LongBench-V1, FASA reaches nearly 100\% of full-KV performance when only keeping 256 tokens, and  achieves 2.56$\times$ speedup using just 18.9\% of the cache on AIME24 \footnote{Code is available at \href{https://github.com/AMAP-ML/FASA-ICLR2026}{https://github.com/AMAP-ML/FASA-ICLR2026}}.

%Building on this insight, we introduce \mymodel, a simple yet effective two-stage framework for efficient inference. \mymodel first leverages these dominant FCs to dynamically predict query-aware token importance. This allows for the precise identification of salient tokens, which are then exclusively used in a second stage for accurate, full-attention generation. Since accessing only a small fraction of the KV cache for the prediction, \mymodel significantly enhances both memory bandwidth and computational efficiency.
%Remarkably, \mymodel even achieves nearly 100\% of the full KV performance using 256 tokens on LongBench-V1 and 18.9\% of the KV cache on AIME24, leading to 6.6$\times$ throughput.
%Experimental results show that \mymodel consistently outperforms existing token-eviction baselines on both long-context and long-form generation tasks. 
\end{abstract}

\section{Introduction}
Despite recent advances in Large Language Models~\citep{dao2022flashattentionfastmemoryefficientexact,ainslie2023gqa,liu2024deepseek,wang-etal-2025-position} in long-context processing, requirements such as repository-level code analysis~\citep{chen2021evaluatinglargelanguagemodels,shi2025longcodezip,wang2026swe,chen2025swe} and document summarization~\citep{goyal2020evaluatingfactualitygenerationdependencylevel} pose both memory and computational challenges, especially the linear growth of the KV cache. As the sequences grow, each token generation requires accessing the entire KV cache, leading to increased memory I/O latency. This memory-bound process underutilizes high-performance GPUs, ultimately limiting the overall throughput.
To optimize KV cache management, previous studies have proposed mainly five directions:  \textit{token eviction} \citep{akhauri2025tokenbutlertokenimportancepredictable,fang2025attentionrag}, \textit{low-rank compression} \citep{chang2025palu,singhania2024loki,zhang2025lorc}, \textit{quantization} \citep{hooper2025kvquant10millioncontext,liu2024kivi},  \textit{KV  merging} \citep{wang2025model,wan2025textdtexto,liu2024minicache}, and \textit{budget allocation} \citep{cai2025pyramidkvdynamickvcache}. 

An intuitive and widely explored approach is \textit{token eviction}~\citep{li2025a,liu2023scissorhandsexploitingpersistenceimportance}.
The rationale is that only a small subset of tokens contributes significantly to outputs, enabling the selective removal of trivial ones. Existing token eviction methods can be classified into three types:  \textbf{(1)} \emph{Static strategies} remove tokens with fixed rules \citep{xiao2024efficient}, therefore risking irreversible information loss; \textbf{(2)} \emph{Adaptive strategies} either permanently evict less critical tokens \citep{zhang2023ho,li2024snapkv} or preserve the full cache while retrieving a subset of entries \citep{tang2024quest,ge2024model}. Yet such heuristic rankings provide an imperfect proxy for the truly dynamic nature of token importance; \textbf{(3)} \emph{Learning-based strategies} \citep{akhauri2025tokenbutlertokenimportancepredictable,yang2025attentionpredictortemporalpatternmatters,chen2025sepllmacceleratelargelanguage} rely on a trained token predictor, suffering from poor generalization on different datasets. \emph{Can a token predictor achieve \textbf{query-awareness} without resorting to costly training?} 

In response to this question, we introduce \mymodel (\underline{F}requency-\underline{A}ware \underline{S}parse \underline{A}ttention), a \textbf{training-free}, \textbf{high-granularity}, \textbf{query-aware} predictor designed to evaluate token significance during the decoding phase, in a training-free manner. The design of \mymodel is rooted in an intriguing observation that differential frequencies within RoPE~\citep{su2023roformerenhancedtransformerrotary} induce functional sparsity among frequency chunks (FCs). Only a sparse subset of FCs, termed as dominant FCs, contribute significantly to contextual awareness, while others construct robust positional patterns.
We empirically verify
% \sout{indicate}\il{verify? demonstrate?} 
that these dominant FCs are sparse, universal, and task-agnostic in Section~\ref{Detection of Contextual Frequency Chunks}, thereby providing a robust foundation for accurately predicting token importance. 

\begin{figure}[t]
\centering
\includegraphics[width=.95\textwidth]{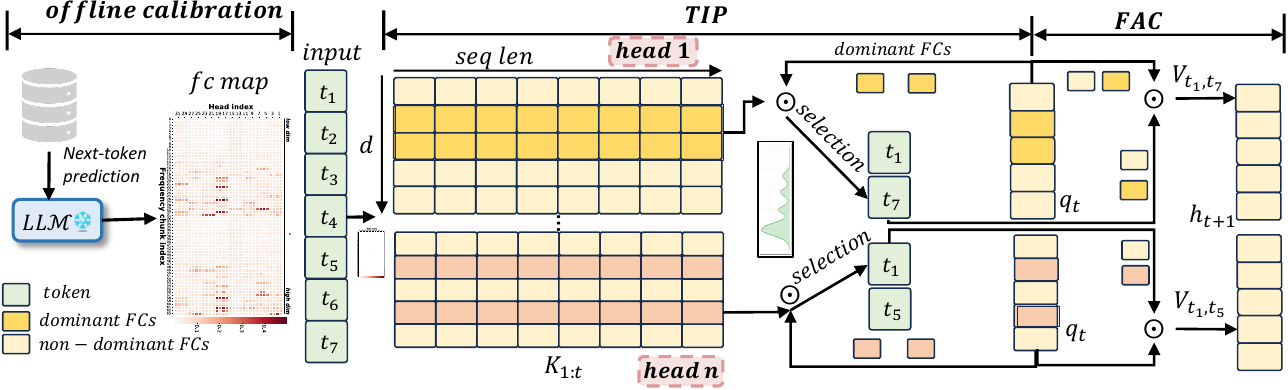} 
\vspace{-0.1in}
  \caption{\small Method Overview of \mymodel. First, the \textbf{TIP} stage leverages only dominant FCs to efficiently estimate token importance and select a critical subset of tokens. Then, the \textbf{FAC} stage performs full-dimensional attention exclusively on this reduced subset to generate the next token. See discussion about design in Appendix~\ref{appendix: design choices}.}
  \label{fig: method}
\vspace{-0.2cm}
\end{figure}

Building upon this insight, \mymodel employs a two-stage framework for efficient inference. The first stage, Token Importance Prediction, harnesses dominant FCs to dynamically estimate attention scores, obtaining critical tokens. At the second stage, Focused Attention Computation then performs precise and focused token generation on this reduced set. The overhead of \mymodel is minimal because the identification of dominant FCs is a one-time and task--invariant process. Ultimately, \mymodel achieves high efficiency by fetching only a small fraction of the KV cache, which significantly reduces the data transferred between memory and the processor and thereby lowers memory bandwidth consumption. The overview of \mymodel is in Figure~\ref{fig: method}.
Grounded on the same principles above, we introduce two variants of \mymodel: \textbf{\mymodel-M} and \textbf{\mymodel-C}. While they differ in implementation strategies, both \textit{achieve equivalent downstream task performance} while offering different efficiency profiles, specializing in memory and computation, respectively.
% \mymodel-M redesigns the cache architecture by offloading infrequently accessed non-dominant FCs of the key cache and the entire value cache to the CPU. This approach is well-suited for memory-constrained scenarios, ensuring efficient memory utilization.
% \mymodel-C retains the entire KV cache on the GPU.  It achieves substantial speedups by reducing memory bandwidth consumption, as it only needs to access a small fraction of the key states.
% \mymodel provides the flexibility to select the optimal variant based on specific hardware constraints. 
Crucially, despite \mymodel leverages a low-rank subspace, its primary objective is the dynamic prediction of token importance, not mere dimensionality reduction. This design makes \mymodel orthogonal to and compatible with most other KV cache compression methods. For example, it can be seamlessly integrated with layer-wise budget allocation schemes like PyramidKV~\citep{cai2025pyramidkvdynamickvcache}. 
% Moreover, quantization techniques~\citep{xiao2024smoothquantaccurateefficientposttraining} can be applied to the dominant FCs identified by \mymodel, yielding compounded savings in memory and latency.

% It is noteworthy although our method performs approximation in a low-rank space, the core motivation behind \mymodel remains the dynamic prediction of token importance, rather than dimensionality reduction. Meanwhile, \mymodel is orthogonal to other KV cache compression methodologies, allowing for synergistic integration with most existing techniques. For instance, \mymodel is fully compatible with layer-wise budget allocation strategies, such as PyramidKV~\citep{cai2025pyramidkvdynamickvcache}, which assigns a larger cache budget to shallower layers. Furthermore, quantization techniques~\citep{xiao2024smoothquantaccurateefficientposttraining} can be readily incorporated; quantizing the key states of the dominant FCs selected by \mymodel will yield additional reductions in both memory footprint and inference latency.

We evaluated \mymodel across a range of LLMs with varying KV cache budgets, concentrating on three core tasks: long-context benchmark, long-sequence modeling, and long chain-of-thought (LongCoT) reasoning. Our method achieves performance comparable to that of full KV cache, with reduction of less than 0.7\%, while consistently surpassing all baseline methods across these tasks. \mymodel-M provides an 8$\times$ compression of the KV cache, substantially optimizing memory usage. and \mymodel-C delivers 2.6$\times$ speedups, enhancing computational efficiency, with 25\% of FCs selected. Our contributions are summarized as follows:
\begin{itemize}[leftmargin=*,itemsep=.5pt,topsep=0pt,parsep=0pt]
  % \vspace{-.1in}
  \item We are the first to uncover an intriguing finding: functional sparsity at FC-level induced by RoPE.
  \item Leveraging the functional sparsity of FCs, we introduce \mymodel, a training-free framework for dynamically predicting token importance.
  \item We present two variants of \mymodel: \mymodel-M,  optimized for settings with memory constraints, and \mymodel-C, designed for scenarios with computational constraints.
  \item Extensive experiments across three paradigm tasks
  % \il{paradigm tasks}
  demonstrate that \mymodel consistently achieves near-oracle accuracy in both long-context and long-generation tasks.
\end{itemize}

\section{Related Works}
\textbf{Token Eviction.}  A central theme in recent KV cache optimization ~\citep{hooper2025squeezedattentionacceleratinglong,wang2025prefixkvadaptiveprefixkv} is the exploitation of inherent, query-dependent attention sparsity~\citep{liu2024retrievalattentionacceleratinglongcontextllm,liu2025clusterkvmanipulatingllmkv,behnam2025rocketkvacceleratinglongcontextllm}. Stream~\citep{xiao2024efficient} employs a rigid heuristic, preserving only initial and recent tokens, which invariably discards potentially crucial information from intermediate positions. SnapKV~\citep{li2024snapkv} improves on this by introducing a one-time, prefill-stage filtering based on empirically estimated attention scores. However, the static nature of this estimation cannot adapt to the evolving relevance of tokens as generation progresses. Quest~\citep{tang2024quest} offers a more dynamic solution by organizing the KV cache into pages and selectively fetching them. Despite its dynamism, its efficacy is hampered by a coarse, page-level granularity, which incurs significant overhead by forcing the retrieval of entire pages even when only a few tokens are needed.

\textbf{Low-rank Compression.}  Another prominent paradigm for KV cache compression is low-rank approximation~\citep{zhang2025lorc,dong2024lesssynthesizingrecurrencekv}, predicated on the observation that the cache's information content is concentrated in a low-dimensional subspace~\citep{sun2025shadowkvkvcacheshadows,saxena2024eigenattentionattentionlowrank,behnam2025rocketkvacceleratinglongcontextllm}. For instance, SparQ~\citep{ribar2024sparq} employs a heuristic that selects key dimensions based on high query-vector magnitudes, a strategy that proves suboptimal due to its head-agnostic nature and its simplistic reliance on magnitude as a proxy for importance. Similarly, LoKi~\citep{singhania2024loki} leverages Principal Component Analysis (PCA) to project key states into a compact subspace for efficient computation, but at the cost of significant memory overhead from storing the requisite projection matrices. In contrast, our proposed \mymodel circumvents these limitations by operating in-place on the KV cache, thereby incurring no auxiliary memory overhead.

\section{Observation}
% Heterogeneous Frequency Chunks
\subsection{Preliminary: Rotary Positional Encodings (RoPE)}
RoPE embeds relative position information into the self-attention computation.  
% At its core, RoPE leverages a key principle: by applying rotational transformations $\mathbf{R}$ to query and key vectors conditioned on their absolute positions, the resulting attention score becomes a function of only their relative displacement. 
Specifically, for a  query vector $\mathbf{q}_{t_1}$ and a key vector $\mathbf{k}_{t_2}$ at positions $t_1$ and $t_2$,  the attention score is formulated as $\mathbf{A}_{t_1, t_2}\!=\!(\mathbf{q}_{t_1} \mathbf{R}_{t_1})(\mathbf{k}_{t_2} \mathbf{R}_{t_2})^\top\!=\!\mathbf{q}_{t_1} \mathbf{R}_{\Delta t} \mathbf{k}_{t_2}^\top$. Due to the orthogonality,  the product of $\mathbf{R}_{t_1}$ and $\mathbf{R}_{t_2}$ elegantly simplifies to a single rotation matrix parameterized solely by the relative offset  $\Delta t = t_1 - t_2$.

\textbf{A Frequency-Chunk Perspective on RoPE.}  
From a frequency-domain perspective, the RoPE mechanism can be interpreted through the concept of ``frequency chunks'' (FCs).  This framework posits that any $d$-dimensional vector $\mathbf{v}\in \mathbb{R}^d$ (e.g., a query and key) is partitioned into  $d/2$ orthogonal 2D subspaces. We denote the $i$-th such subspace, or FC, as $\mathbf{v}^{[i]}=(v_{2i},v_{2i+1})^T$.
Each FC is associated with a unique base angular frequency, calculated as $\theta_i\!=\!B^{{-2(i-1)}/{d}}$ for $i \in \{ 1, \dots, d/2$\}, where $B$ is a predefined frequency base. This design establishes a direct mapping from a chunk's dimensional indices $(2i,2i+1)$ to its rotational frequency. \textit{Lower dimension indices ($i$) result in higher frequencies, which implies that the corresponding FCs rotate very quickly physically.}
For a token at absolute position $m$, its $i$-th FC is rotated by an angle $m\theta_i$ through a specific  $2 \times 2$ rotation matrix $\textbf{R}_{m,\theta_i}$.  
The global rotation matrix $\mathbf{R}_{\Delta t}$ is  block-diagonal, where each diagonal block is a $2\times2$ rotation matrix $\mathbf{R}_{\Delta t, \theta_i}$ and defined as $\mathbf{R}_{\Delta t} = \operatorname{Diag}(\mathbf{R}_{\Delta t, \theta_1}, \mathbf{R}_{\Delta t, \theta_2}, \dots, \mathbf{R}_{\Delta t, \theta_{d/2}}) = \bigoplus_{i=1}^{d/2} \mathbf{R}_{\Delta t, \theta_i}$.
\vspace{-.1in}
\begin{equation}
    \mathbf{v}_{m} = \bigoplus_{k = 1}^{d/2} \mathbf{v}_m^{[i]} = \bigoplus_{k = 1}^{d/2} (\mathbf{v}_{2i},\mathbf{v}_{2i+1})^T,   \mathbf{R}_{m, \theta_i} = \begin{pmatrix} \cos(m\theta_i) & -\sin(m\theta_i) \\ \sin(m\theta_i) & \cos(m\theta_i) \end{pmatrix}.\\
\label{frequency chunks defination} 
\end{equation}
\vspace{-.1in}
% \begin{equation}
%     \mathbf{R}_{\Delta t} = \operatorname{diag}(\rho(\Delta t \cdot u_1), \rho(\Delta t \cdot u_2), \dots, \rho(\Delta t \cdot u_{d/2})) = \bigoplus_{k=1}^{d/2} \rho(\Delta t \cdot u_k).
% \label{eq: the global rotation matrix}
% \end{equation}
% \begin{equation}
%     \mathbf{R}_m = \operatorname{diag}(\mathbf{R}_{m, \theta_1}, \mathbf{R}_{m, \theta_2}, \dots, \mathbf{R}_{m, \theta_{d/2}}) = \bigoplus_{i=1}^{d/2} \mathbf{R}_{m, \theta_i},
% \label{eq: the global rotation matrix}
% \end{equation}
\subsection{Motivation and Hypothesis}
\paragraph{Position vs. Semantics: Different Roles of FCs.}
The varying rotational velocities across FCs inherently lead to functional heterogeneity. This principle is substantiated by two key observations from prior literature. First, a distinct division of labor exists within RoPE~\citep{barbero2025round, wei2025videorope}, where high-frequency FCs (in low dimensions) are primarily responsible for constructing robust positional patterns, and in contrast, low-frequency counterparts specialize in carrying the semantic information and model long-range dependencies. Second, this functional specialization is structurally reflected by a RoPE-induced concentration of high-magnitude values within specific query and key dimensions~\citep{sun2024massive}, reinforcing the non-uniform functional importance of FCs.
This functional heterogeneity suggests that FCs can be grouped into two distinct categories:
\begin{enumerate}[itemsep=-.5pt, leftmargin=.17in]
\vspace{-.1in}
    \item \textbf{Contextual FCs:} A small, critical subset responsible for dynamic, context-specific attention. These FCs identify which tokens are semantically relevant to the current query.
    \item \textbf{Structural FCs:} The remaining majority primarily injects inherent, positional attention patterns, mainly recency bias~\citep{peysakhovich2023attention} and attention sinks~\citep{xiao2024efficient}.
\end{enumerate}
\textbf{Hypothesis:} \textit{The model's contextual awareness is overwhelmingly driven by the Contextual FCs. A few contextual FCs could replicate the contextual selection behavior of a full attention head.} If their index set is denoted as $\mathcal{I}_{\text{dom}} \subset \{1, \dots, d/2\}$, the full attention dot product can be effectively approximated by summing only over $\mathcal{I}_{\text{dom}}$, namely $\mathbf{A}_{t_1,t_2} = \mathbf{q}_{t_1} \mathbf{R}_{\Delta t} \mathbf{k}_{t_2}^T \sum_{i \in \mathcal{I}_{\text{dom}}} \mathbf{q}_{t_1}^{[i]} \textbf{R}_{\Delta
 t,\theta_i}{\mathbf{k}_{t_2}^{[i]}}^{\top}$.

\subsection{Quantifying Functional Sparsity}
\label{Detection of Contextual Frequency Chunks}
% 
% \paragraph{Contextual Agreement Metric} To verify the sparsity of functionality of frequency chunks \footnote{Only a few frequency chunks are enough to select contextual tokens.}, we devise a compatible metric named \textit{Contextual Agreement (CA)}, measuring the contextual agreement of each single frequency chunk applied to self-attention computation to the counterpart full head-dimension versions. Concetely, we compute the intersection of top-K tokens according to their respective attention weights.  We average $CA_K, K= 0,1,2,\dots ,d/2-1$, across all generated tokens at decoding stage for diverse models and tasks, for all model layers and all heads.

Quantifying our hypothesis of FC-level functional sparsity requires a metric to assess the ``dominance'' of individual FCs. Therefore, we propose the \textbf{Contextual Agreement (CA)} metric, which measures the alignment between the attention pattern from a single FC and that of the full attention head.

\begin{figure}[t!]
    \centering
    % --- 第一行 (4个子图) ---
    % \begin{subfigure}[b]{0.24\textwidth}
    %     \centering
    %     \includegraphics[width=\textwidth]{figures/sparsity_phenomenon/llama3.2-3b-256-22-qasper.pdf}
    % \end{subfigure}%
    \begin{subfigure}[b]{0.38\textwidth}
        \centering
        \includegraphics[width=\textwidth]{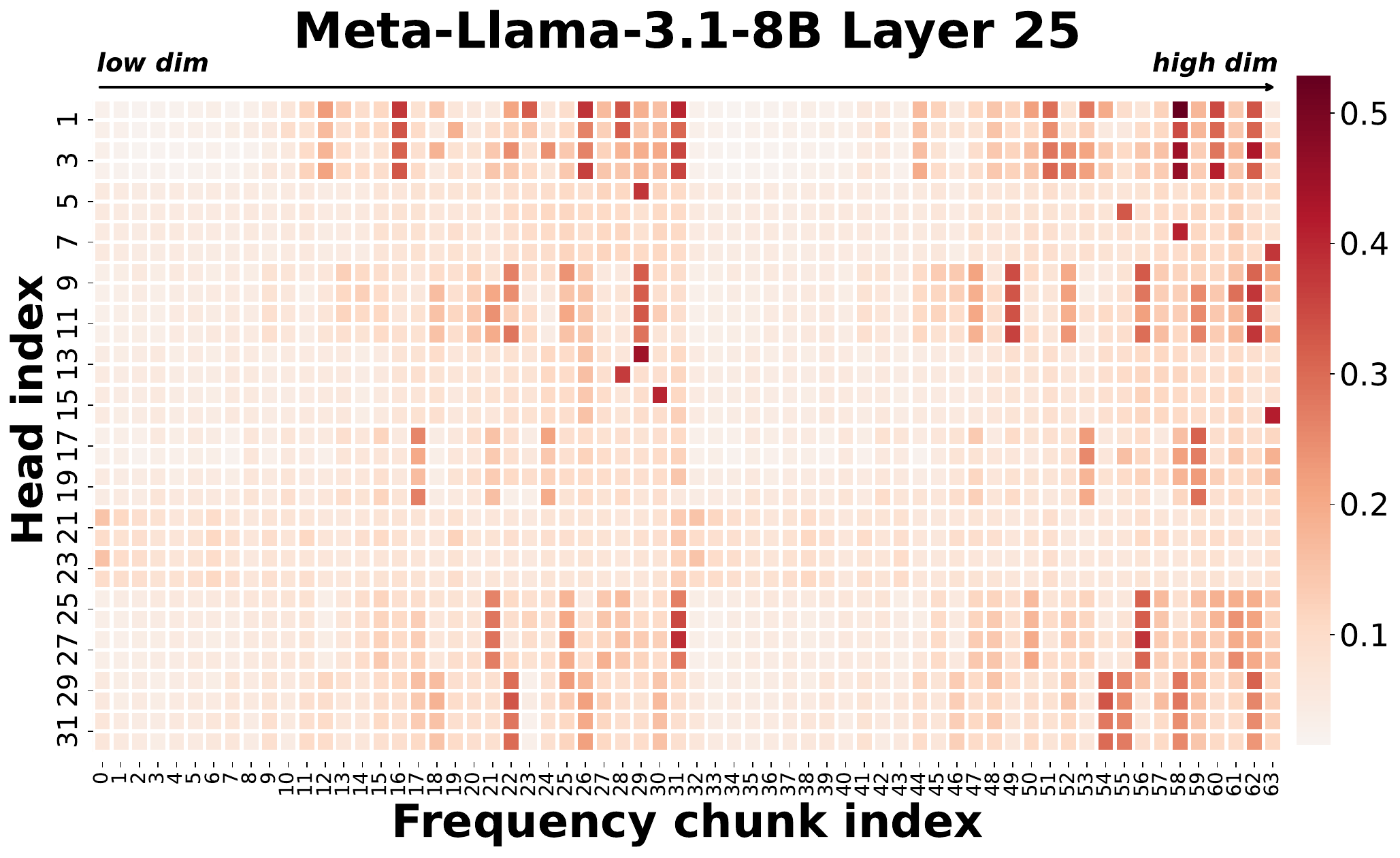}
    \end{subfigure}%
    \begin{subfigure}[b]{0.38\textwidth}
        \centering
        \includegraphics[width=\textwidth]{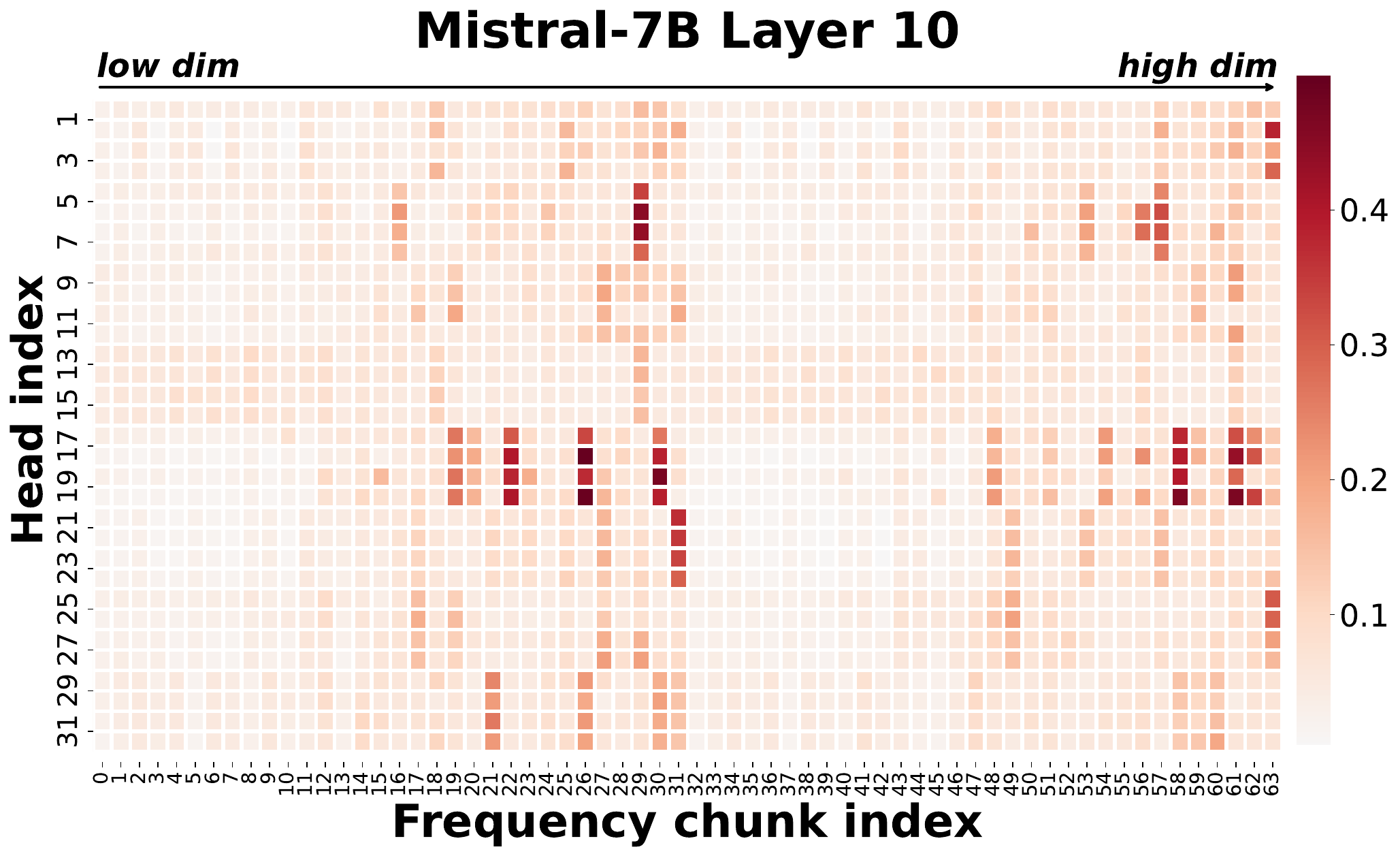}
    \end{subfigure}%
    % \begin{subfigure}[b]{0.3\textwidth}
    %     \centering
    %     \includegraphics[width=\textwidth]{figures/sparsity_phenomenon/qwen14b-qasper-layer5.pdf}
    % \end{subfigure}
    % \vspace{-0.1cm} % 减少行间距
    % --- 第二行 ---
    % \begin{subfigure}[b]{0.48\textwidth}
    %     \centering
    %     \includegraphics[width=\textwidth]{figures/sparsity_phenomenon/llama3b-gov_layer22.pdf}
    % \end{subfigure}%
    % \begin{subfigure}[b]{0.48\textwidth}
    %     \centering
    %     \includegraphics[width=\textwidth]{figures/sparsity_phenomenon/llama8b-gov_layer25.pdf}
    % \end{subfigure}%
    % \begin{subfigure}[b]{0.24\textwidth}
    %     \centering
    %     \includegraphics[width=\textwidth]{figures/sparsity_phenomenon/mistral-gov_layer10.pdf}
    % \end{subfigure}%
    % \begin{subfigure}[b]{0.24\textwidth}
    %     \centering
    %     \includegraphics[width=\textwidth]{figures/sparsity_phenomenon/qwen14b-qasper-layer5.pdf}
    % \end{subfigure}
    
    \vspace{-0.3cm}
    \caption{\small \textbf{Functional sparsity of FCs revealed by Contextual Agreement ($\overline{\text{CA}}$) heatmaps.} Each heatmap shows $\overline{\text{CA}}$ per FC ($x$-axis) across all heads ($y$-axis). A few ``dominant'' FCs (bright vertical bands) consistently capture contextual information across attention heads. Results on Qasper ($\mathcal{K}=256$); see Appendix~\ref{part1: sparse fcs investigations}.}
    % \caption{Heatmaps ($\overline{\text{CA}}_{\mathcal{K}=256}$) on Qasper dataset. Each heatmap visualizes $\overline{\text{CA}}$ for every FC ($x$-axis) across all heads ($y$-axis) within a given layer. Cell color indicates the corresponding $\overline{\text{CA}}$ value, with brighter colors indicating higher agreement with the full-head attention patterns. The consistent emergence of a few brightly-colored columns reveals a functional sparsity, where a small subset of FCs consistently dominates. Additional experiments across model scales and datasets in Appendix~\ref{part1: sparse fcs investigations}.}
\label{fig: sparse_frequency_qasper_gov}
\vspace{-.2cm}
\end{figure}

\textbf{Formal Setup.} For a query $\mathbf{q}_t \in \mathbb{R}^d$ and key matrix $\mathbf{K}_{1:t} \in \mathbb{R}^{d \times t}$ in an attention head $(l, h)$, we define two raw score vectors: the standard \textbf{full-head scores} $\boldsymbol{\alpha}_{l,h}$ and the \textbf{single-FC scores} $\boldsymbol{\alpha}_{l,h}^{(i)}$. The latter are computed using only the 2D components of the $i$-th FC. These are expressed as:
\begin{align}\label{equation of full attentions scores}
    \boldsymbol{\alpha}_{l,h}(\textbf{q}_t,\mathbf{K}_{1:t}) &= [\mathbf{q}_{t} 
        \, 
        \textbf{R}_{t-1} 
        \, 
        (\mathbf{k}_{0})^T, \cdots, \mathbf{q}_{t}
        \, 
        \textbf{R}_0 
        \, 
        (\mathbf{k}_{t})^T]^{T} \\
    \boldsymbol{\alpha}_{l,h}^{(i)}(\textbf{q}_t,\mathbf{K}_{1:t}) &= [\mathbf{q}_{t}^{[i]} 
        \, 
        \textbf{R}_{t-1,\theta_i} 
        \, 
        {\mathbf{k}_{0}^{[i]}}^T, \cdots, \mathbf{q}_{t}^{[i]} 
        \, 
        \textbf{R}_{0,\theta_i} 
        \, {\mathbf{k}_{t}^{[i]}}^T]^{T} 
\end{align}
% \begin{equation}
% \boldsymbol{\alpha}_{l,h}(q_t,\mathbf{K}_{1:t})= [\mathbf{q}_{t} 
%         \, 
%         R_t 
%         \, 
%         (\mathbf{k}_{0})^T, \cdots, \mathbf{q}_{t}
%         \, 
%         R_0 
%         \, 
%         (\mathbf{k}_{t})^T]^{T} 
% \end{equation}
% \begin{equation}
% \boldsymbol{\alpha}_{l,h}^{(i)}(q_t,\mathbf{K}_{1:t})= [\mathbf{q}_{t}^{[i]} 
%         \, 
%         R_{t,\theta_i} 
%         \, 
%         {\mathbf{k}_{0}^{[i]}}^T, \cdots, \mathbf{q}_{t}^{[i]} 
%         \, 
%         R_{0,\theta_i} 
%         \, {\mathbf{k}_{t}^{[i]}}^T]^{T} 
% \end{equation}
% We measure the alignment of a single FC with full-head version through measuring the alignment between $\boldsymbol{\alpha}_{l,h}$ and $\boldsymbol{\alpha}_{l,h}^{(i)} \in \mathbb{R}^{t}$.
\textbf{Metric Definition.}
The \textbf{CA} score, $\text{CA}_\mathcal{K}^{l,h,i}$, quantifies the agreement between the full-head $\boldsymbol{\alpha}_{l,h}$ and single-FC $\boldsymbol{\alpha}_{l,h}^{(i)}$ scores by measuring the normalized intersection of their top-$\mathcal{K}$ token index sets:
\vspace{-.05in}
\begin{equation} \label{eq:ca_k}
\text{CA}_\mathcal{K}^{l,h,i}(q_t,\mathbf{K}_{1:t}) = [\text{TopK-I}(\boldsymbol{\alpha}_{l,h}(q_t,\mathbf{K}_{1:t}), \mathcal{K}) \cap \text{TopK-I}(\boldsymbol{\alpha}_{l,h}^{(i)}(q_t,\mathbf{K}_{1:t}), \mathcal{K})]/\mathcal{K},
\end{equation}
% \vspace{-.05in}
where the operator $\text{TopK-I}(\boldsymbol{\alpha}, \mathcal{K})$ retrieves the top-$\mathcal{K}$ values of a vector $\boldsymbol{\alpha}$.
To assess an FC's importance robustly, we compute its mean CA score, by averaging across several samples from a specific dataset.   Figure~\ref{fig: sparse_frequency_qasper_gov} reveals the distinct functional contribution of each FC across all heads.
% As we are interested in the most salient tokens, we average from $f=0$ up to half the head dimension, $d/2-1$. The aggregated score for a single \textbf{FC} is:
% \begin{equation} \label{eq:ca_agg}
% \text{CA}(\boldsymbol{\alpha}, \boldsymbol{\alpha}^{(f)}) = \frac{1}{d/2 - 1} \sum_{K=1}^{d/2 - 1} \text{CA}_K(\boldsymbol{\alpha}, \boldsymbol{\alpha}^{(f)})
% \end{equation}
\begin{wraptable}{r}{0.47\textwidth}
  \vspace{-4mm}
  \renewcommand{\arraystretch}{0.7}
  \caption{Compound CA scores  under varying number of selected FCs ($F$) and KV cache budgets ($K$). Each head has 64 FCs in total.}
  \vspace{-0.2cm}
  \scalebox{0.8}{
    \begin{tabular}{@{}l!{\vrule width 0.4pt}*{6}{c}@{}}
      \toprule      \diagbox[width=6em,height=2em]{$|\mathcal{I}_{dom}|$}{$K$} & 64 & 256 & 512 & 768&1024 & 2048 \\
      \midrule
   Random & 2.0 &3.6&6.4&19.1&25.5&51.1 \\
    Stream& 34.4 &26.8&24.4&26.5&30.7&53.9\\
    SnapKV &  37.9  &  40.9    &41.9& 45.4 &49.5&66.6  \\
    \midrule
  $F=\textbf{8}$  \scriptsize $(1/8)$   & 43.0 & 49.4 & 54.3 &58.8& 62.6 & 76.1 \\
  $F=\textbf{10}$  & 46.4 & 52.1 & 56.6 &61.1& 64.8 & 77.5 \\
  $F=\textbf{12}$   & 49.7 & 54.7 & 58.9 & 63.4&66.8 & 79.0 \\
 $F=\textbf{14}$     & 52.4 & 56.9 & 60.9 &65.2& 68.5 & 80.2 \\
  \makecell{$F=\textbf{16}$  \scriptsize $(1/4)$}  & 55.3 & 59.7 & 62.8 &66.9& 70.1 & 81.4 \\
      \bottomrule
    \end{tabular}
  }
  
  \label{tab:sparse_fcs_table}
  \vspace{-0.35cm}
\end{wraptable}
\textbf{Sparse and Universal $\mathcal{I}_{\text{dom}}$.} \textbf{(1) \textit{Sparsity}:} A small subset of FCs (dominant FCs) exhibits disproportionately high agreement with full attention patterns, while the vast majority of other FCs have negligible CA scores (typically < 0.15). In Table \ref{tab: quantitative analysis about sparsity}, dominant FCs account for less than 1\% of all FCs, while non-dominant FCs with low CA scores comprise approximately 90\% or more;  \textbf{(2) \textit{Universality}:} The functional sparsity is widely observed across model architectures and scales (Appendix \ref{appendix: fucntional sparsity generalizes model scales and architechtures};Table \ref{tab: quantitative analysis about sparsity});  \textbf{(3) \textit{Task-Invariance:}} The identification of dominant FCs is largely task-agnostic. The saliency maps in Figure \ref{fig: parse_frequency_qasper task-agnotic} derived from tasks such as QA and summarization exhibit remarkable consistency. Quantitatively, the overlap of dominant FCs across different calibration datasets consistently exceeds 70\% in all tested models, as reported in Table \ref{tab:cross-task overlap}. This indicates that the functional roles of FCs are intrinsic to the underlying mechanics of RoPE, rather than being task-specific adaptations.

\textbf{Reconstructing Functionality from $\mathcal{I}_{\text{dom}}$.}
The analysis above supports that the functionality of a full attention head can be reconstructed using only its most dominant $F$ components $\mathcal{I}_{\text{dom}}^{l,h} =\text{TopK-I}(\{\text{CA}_\mathcal{K}^{l,h,i} \mid\!0\leq\!f<\!d/2\}, F)$. Therefore, we measure the collective efficacy of this subset using a compound CA score, $\text{CA}_K^{l,h,\mathcal{I}_{\text{dom}}}$, and present the results in Table~\ref{tab:sparse_fcs_table}. For comparison, we benchmark against token-eviction methods, which serve to emphasize the capability of predicting token importance. Our method demonstrates remarkable efficiency: with just 1/8 of the components selected under a tight budget 64, $\mathcal{I}_{\text{dom}}$ achieves an accuracy of 43\%, surpassing the strong baseline SnapKV~\citep{li2024snapkv} by an average of 10.3\% across all budget levels.  
We also present the predictive distribution of dominant FCs across tokens grouped by attention score quintiles in Table \ref{tab: predictive distribution of dominant fcs}. This analysis further substantiates the ability of dominant FCs to effectively capture both the relative ranking and true impact of context tokens.
% \begin{equation}
%     \mathcal{I}_{\text{dom}}^{l,h} =\text{TopK-Indices}(\{\text{CA}_K^{l,h,f} \mid f\in[0,\cdots,d/2]\}, F)
% \label{eq:defination of i_dom}
% \end{equation}
%paragraph 

% \noindent \textbf{A Unified View.} In prior work, two thoughts are most related to and consistent with the phenomenon identified in this paper.   
% % One view from ~\citep{jin2025massive}, 
% While prior work has pragmatically exploited the emergent low-rank properties of query and key states for efficient KV cache management~\citep{singhania2024loki,chang2025palu}, our analysis provides the first principled explanation for \textbf{the origin} of this structure. We demonstrate that it is an intrinsic property of RoPE's architecture, rooted in the functional dominance of a sparse set of FCs. This framework unifies previously disparate observations, such as the concentration of massive values in query and key states noted by \citeauthor{jin2025massive}, which treat these phenomena in isolation, we reveal the critical insight lies not in the values themselves, but in their synergistic interaction within specific FCs, an interaction that governs the attention distribution and ultimately generates the effective low-rank behavior of the head.
\section{Method}
Grounded in the functional sparsity of FCs, our training-free framework \mymodel employs a two-stage, coarse-to-fine strategy to circumvent the prohibitive cost of full self-attention. 
First, the \textbf{Token Importance Predictor (TIP)} stage utilizes a computationally frugal proxy, defined by a pre-calibrated set of dominant FCs, $\mathcal{I}_{\text{dom}}$, to efficiently identify a small subset of contextually salient tokens. 
Subsequently, the \textbf{Focused Attention Computation (FAC)} stage performs a full-fidelity attention computation exclusively on this salient subset, preserving high generation fidelity while drastically mitigating the computational and memory overhead of standard attention.
\subsection{Token Importance Predictor (TIP)}
% TIP is founded on the principle that the dominant frequencies can serve as a computationally efficient token importance predictor. Building on the foundations introduced in Section~\ref{Detection of Contextual Frequency Chunks}, we identify $\mathcal{I}_{dom}$ with a one-time offline calibration.
The TIP stage operates on the principle that dominant frequencies are an efficient proxy for token importance, where the dominant indices $\mathcal{I}_{dom}$ are identified via a one-time offline calibration.

\textbf{Offline Calibration: Identifying $\mathcal{I}_{dom}$.}
The objective of the offline calibration is to identify a small, head-specific set of \textit{dominant frequencies}, $\mathcal{I}_{\text{dom}}^{l,h}$, for each attention head $(l,h)$.
%, since the emergence of $\mathcal{I}_{dom}$ is highly dynamic.\il{contradicts task-agnostic before?} 
We formulate this process as a search problem over frequency indices. Given a small calibration dataset $\Omega$ and a target size $N_{tip}$, our goal is to find the subset of FCs of cardinality $N_{tip}$ that maximizes the expected average of CA scores.  The objective is defined as:
% \begin{align}
%     \mathcal{I}_{\text{dom}}^{l,h} = \underset{\substack{\mathcal{I} \subseteq \{0, \dots, d/2-1\} \\ |\mathcal{I}|=k}}{\text{argmax}} \ \mathbb{E}_{(q, \mathbf{K}) \sim \Omega}  \left[ \sum_{i \in \mathcal{I}} \text{CA}^{l,h,i}_{B}(q, \mathbf{K}) \right] 
%     \label{eq:dom_freq_selection_calibrated}
% \end{align}
\begin{equation}
    \mathcal{I}_{\text{dom}}^{l,h} = \underset{\substack{
        \mathcal{I} \subseteq \{0, \dots, d/2-1\}, |\mathcal{I}|=N_{tip}
    }}{\operatorname{argmax}} \,
    \mathbb{E}_{\textbf{q}, \mathbf{K} \sim \Omega}
    \left[ \sum_{i \in \mathcal{I}} \mathrm{CA}_{\mathcal{K}}^{l,h,i}(\textbf{q}, \mathbf{K}) \right].
\end{equation}
This calibration is a highly efficient, one-time offline process because the resulting $\mathcal{I}_{\text{dom}}$ is empirically found to be task-agnostic and can be robustly identified from a minimal number of samples. Its associated computational cost is negligible. The detailed algorithm is provided in Algorithm \ref{alg:offline_calibration}.

% \il{referring to Algo2 before Algo1. Move both to Appendix due to space limits.}
% \il{we didn't make that connection before in observation section?}
\textbf{Online Prediction: Importance Scoring via Frequency Subspace Aggregation.}
During the online prediction phase at a given decoding step $t$, we leverage the pre-calibrated set of dominant frequencies, $\mathcal{I}_{\text{dom}}^{l,h}$, to efficiently estimate token importance in a training-free manner.
Conceptually, the full attention score for a query $\textbf{q}_t$ and keys $\mathbf{K}_{1:t}$ can be decomposed into a sum of contributions from all $d/2$ frequency components: $ \boldsymbol{\alpha}^{l,h}(\textbf{q}_t, \mathbf{K}_{1:t})\!=\!\sum_{i=0}^{d/2-1} \boldsymbol{\alpha}^{l,h,i}(\textbf{q}_t, \mathbf{K}_{1:t})$.
Instead of performing this computationally expensive summation, our method constructs an \textit{importance score vector} $\mathbf{S}_t^{l,h}$, by exclusively aggregating the contributions from the pre-identified dominant frequencies, i.e., $\mathbf{S}_t^{l,h} \triangleq \sum_{i \in \mathcal{I}_{\text{dom}}^{l,h}} \boldsymbol{\alpha}^{l,h,i}(\textbf{q}_t, \mathbf{K}_{1:t})$.
This formulation strategically bypasses computation for non-dominant frequencies. Finally, based on these scores, we identify the set of top-$N_{fac}$ most important token indices, $\mathcal{T}_t$, for the subsequent FAC stage: $\mathcal{T}_t = \operatorname{TopK-I}(\mathbf{S}_t^{l,h}, N_{fac})$.
\subsection{Focused Attention Computation (FAC)}
Following the identification of the contextually important token set $\mathcal{T}_t$ by the TIP module, this stage executes an attention computation on $\mathcal{T}_t$, enabling the model to concentrate its computational resources on the most salient parts of the context. Specifically, for the current query vector $\mathbf{q}_t$ at decoding step $t$, instead of using the full key and value matrices ($\mathbf{K}_{1:t}, \mathbf{V}_{1:t}$) from the entire past context, we first gather the keys and values corresponding to the indices in $\mathcal{T}_t$:
% \vspace{-.05in}
\begin{equation}
    \mathbf{K}_{\mathcal{T}_t} = \operatorname{Gather}(\mathbf{K}_{1:t}, \mathcal{T}_t), \quad \mathbf{V}_{\mathcal{T}_t} = \operatorname{Gather}(\mathbf{V}_{1:t}, \mathcal{T}_t)
\end{equation}
% \vspace{-.1in}
where the $\text{Gather}(\cdot)$ operation selects the rows from the original matrices specified by the index set $\mathcal{T}_t$. 
The attention scores for each head $(l,h)$ are then computed using only these selected keys. The final output vector for the head is subsequently produced by weighting the selected value vectors:
\begin{align}
    \hat{\boldsymbol{\alpha}}_{\text{FAC}}^{l,h} = \operatorname{Softmax}\left(\mathbf{q}_t {\mathbf{K}_{\mathcal{T}_t}}^T/ \sqrt{d}\right), \quad
    \mathbf{O}_t^{l,h} = \hat{\boldsymbol{\alpha}}_{\text{FAC}}^{l,h} \mathbf{V}_{\mathcal{T}_t} \label{eq:fac_output}
\end{align}
Critically, the original absolute positions of the tokens in $\mathcal{T}_t$ are preserved.
This directly maintains the integrity of their position embeddings and the vital spatial information they encode, preventing the performance degradation associated with positional distortion. 
In essence, the FAC stage functions as a high-fidelity computational filter, restricting full-precision attention to the most salient tokens to achieve a compelling balance between computational efficiency and predictive accuracy.
\subsection{Two Implementations of FASA}
We introduce two specialized, hardware-aware variants of \mymodel that offer a trade-off between memory and speed: (1). \textbf{\mymodel-M (Memory-Optimized)} minimizes its GPU memory footprint by strategically offloading the value cache and non-dominant key components to CPU memory, making it ideal for VRAM-constrained environments. To mitigate the latency from CPU-GPU data transfer, this approach can be effectively paired with prefetching techniques. (2) \textbf{\mymodel-C (Computation-Optimized)} prioritizes inference speed by retaining the full cache on-GPU but accessing only a sparse subset of key states, drastically reducing memory I/O for significant acceleration.
 (See Appendix~\ref{varian of fasa} for details and memory analysis of \mymodel-M).
\subsection{Efficiency Analysis of FASA}

\begin{wrapfigure}{r}{0.4\textwidth}
\vspace{-0.2cm}
\includegraphics[width=\linewidth]{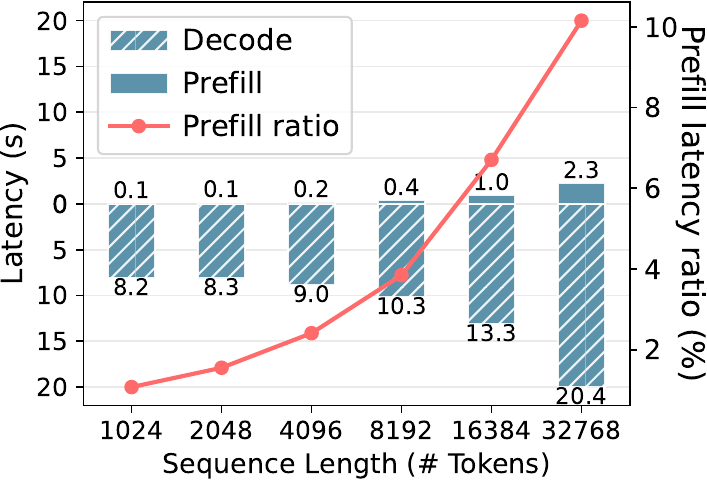} 
% \vspace{-.1in}
\vspace{-0.7cm}
  \caption{\small Decoding latency dominates total latency in auto-regressive generation.}
  \label{fig: prefill_takeup_ratio}
  \vspace{-0.1in}
\end{wrapfigure}
\textbf{Computational Analysis.} At the generation step $t$, the complexity of computing $\mathbf{q}_t\mathbf{K}^{\mathbf{T}}_{1:t}$ is $\mathcal{O}(td)$ and the complexity of multiplying the value states with attention scores is $\mathcal{O}(td)$ per head. For \mymodel, (1) the complexity of the \textbf{TIP} stage is $\mathcal{O}(2tN_{\text{tip}})$ (each FC takes up 2 dimensions), since this stage operates in low-dimensional subspaces, and (2) the \textbf{FAC} stage performs attention on a reduced set of $N_{fac}$ tokens, leading to a complexity of $\mathcal{O}(N_{fac}d)$. Additionally, the detection of dominant frequencies $\mathcal{I}_{dom}$ is offline, one-time, and applicable for various tasks and the burdens from this part could be neglected. Assuming the complexity of selecting the top-k tokens is small, the
overall complexity of \mymodel is $\mathcal{O}(2tN_{tip} + 2N_{fac}d)$. The theoretical speedup at decoding stage is in Equation~\ref{equation: speedup}.
\begin{equation}\label{equation: speedup}
    \operatorname{Speedup} =\frac{2td}{2tN_{tip} + 2N_{fac}d}=\frac{1}{N_{tip}/d+N_{fac}/t},\operatorname{Speedup} \approx\frac{d}{N_{tip}} \hspace{1mm}if \hspace{1mm} N_{fac}\ll t
\end{equation}
% \paragraph{Memory Movement Reduction.} The auto-regressive decode stage is generally memory-bound, as each step requires loading entire KV cache. This is confirmed in Figure \ref{fig: prefill_takeup_ratio}, where decoding consititutes ~90\% of the total latency at a 32K sequence length. \mymodel directly mitigates this bottleneck by trafically reducing memory trffic.  At the generation step $t$, where a standard attention loads $2tm$ of KV cache, \mymodel  loads $2tmN_{tip}/d$ for TIP and $2N_{fac}m$ for FAC. Therefore, \mymodel loads $(2tmN_{tip}/d+2N_{fac}m)/2tm=N_{tip}/d+N_{fac}/t \approx N_{tip}/d (N_{fac}\ll t)$  of the total cache, alleviating the memory-bound constraint.
\textbf{Memory Movement Reduction.}
The auto-regressive decoding stage is notoriously memory-bound, as requiring loading the entire KV cache, creating a significant latency bottleneck. This is confirmed in Figure~\ref{fig: prefill_takeup_ratio}, where decoding constitutes  $90\%$ of the total latency at a 32K context. \mymodel, directly mitigates this bottleneck by drastically reducing memory traffic. At a decoding step $t$, standard attention loads $2tm$ bytes from the KV cache (with $m$ as the byte size per state vector) while \mymodel accesses only $t(2N_{tip}/d*m)$ bytes (only keys) for the TIP and $2N_{\text{fac}}m$ bytes for the FAC. The fraction that \mymodel must load is therefore: $(2tmN_{tip}/d+2N_{fac}m)/2tm=N_{tip}/d+N_{fac}/t \approx N_{tip}/d (N_{fac}\ll t)$, which alleviates the memory-bound constraint of long-context decoding.

\section{Experiments}
\subsection{Experimental Setting}\label{section: Experimental Setting}
% To demonstrate the effectiveness and versatility of \mymodel, we conduct a comprehensive evaluation. The experimental setup is designed to rigorously assess performance across various models, against strong baselines, and on a diverse set of challenging long-context tasks. 
\noindent \textbf{Baselines and Models.}
% \il{Consistent baseline names textsc or no? Consistent benchmark names and model names. Ref for first intro of benchmarks and models.}
To comprehensively evaluate \mymodel's performance, we benchmark it against into two groups of robust baselines: \textbf{(1) State-of-the-art methods:} We compare against leading token eviction methods in efficient KV cache management, including  Stream~\citep{xiao2024efficient},  SnapKV~\citep{li2024snapkv}, RKV~\citep{cai2025r}, Quest~\citep{tang2024quest}, H2O~\citep{zhang2023ho}; \textbf{(2) Upper bounds:} two theoretical bounds, FKV, which represents standard inference with the complete, uncompressed KV cache, serving as the absolute performance ceiling due to no information loss, and Oracle, a more pragmatic upper bound for eviction-based methods, assuming ideal knowledge to retain only the most critical tokens based on full-head scores. Our experiments span a variety of cutting-edge architectures and model sizes, specifically Llama~\citep{touvron2023llama}, Mistral~\citep{jiang2023clip}, and Qwen~\citep{bai2023qwen}. 

\noindent \textbf{Evaluation Benchmarks.}
To rigorously assess the capabilities of \mymodel across diverse long-context scenarios~\citep{liu2025attention}, we conduct comprehensive evaluations spanning three paradigms: (1) \textbf{Long-context understanding:} We use diverse, real-world tasks from LongBench~\citep{bai2024longbench} to assess the ability to identify critical information within lengthy contexts. (2) \textbf{Long-Sequence Modeling:} We measure perplexity on PG-19~\citep{rae2019compressivetransformerslongrangesequence}, WikiText~\citep{merity2017pointer}, and C4~\citep{2019t5} corpus to evaluate generative fidelity over long dependencies. (3) \textbf{Long-CoT Reasoning:} To test performance in long-generation scenarios, we evaluate on complex mathematical reasoning tasks from MATH500~\citep{hendrycks2021measuring} and 
AIME24~\citep{AIME2024} on R1-LLMs.

\begin{table*}[t!]

\fontsize{22}{28}\selectfont
\setlength{\tabcolsep}{5pt}
\centering

\caption{\small Performance of \mymodel on diverse models on LongBench-V1 benchmarks. For baselines, we retain constant token budget (256) and 25\% FCs for \mymodel. $^{\dagger}$FKV and Oracle are full and look-ahead upper bounds.}
\vspace{-0.1in}

% \begin{threeparttable}
\renewcommand{\arraystretch}{0.3}
\scalebox{0.37}{
\begin{tabular}{l|l|cccccccccccccccl@{}}
\specialrule{2pt}{0pt}{2.3pt}
&\multirow{4}{*}{\fontsize{23}{25}\selectfont\textbf{Method}} & \multicolumn{3}{c}{\fontsize{21}{28}\selectfont\textbf{Single-Doc QA}} & \multicolumn{3}{c}{\fontsize{21}{28}\selectfont\textbf{Multi-Doc QA}}& \multicolumn{3}{c}{\fontsize{22}{28}\selectfont\textbf{Summarize}}& \multicolumn{2}{c}{\fontsize{22}{28}\selectfont\textbf{Summarize}}& \multicolumn{2}{c}{\fontsize{22}{28}\selectfont\textbf{Synthetic}} & \multicolumn{2}{c}{\fontsize{22}{28}\selectfont\textbf{Code}}  \\
\cmidrule(lr){3-5}\cmidrule(lr){6-8}\cmidrule(lr){9-11}\cmidrule(lr){12-13}\cmidrule(lr){14-15}\cmidrule(lr){16-17}
&& \rotatebox[origin=c]{45}{NQA} & \rotatebox[origin=c]{45}{Qasp} & \rotatebox[origin=c]{45}{MF-en} & \rotatebox[origin=c]{45}{Hqa} & \rotatebox[origin=c]{45}{2Wiki} & \rotatebox[origin=c]{45}{Musi} & \rotatebox[origin=c]{45}{GovR} & \rotatebox[origin=c]{45}{Qsum} & \rotatebox[origin=c]{45}{Mult} & \rotatebox[origin=c]{45}{Trec} & \rotatebox[origin=c]{45}{Tqa} &  \rotatebox[origin=c]{45}{Pcnt} & \rotatebox[origin=c]{45}{Pre} & \rotatebox[origin=c]{45}{Lcc} & \rotatebox[origin=c]{45}{RB-P} &\rotatebox[origin=c]{45}{AVG.}  \\

% \specialrule{1pt}{2pt}{2pt}
\midrule

\multirow{6}{*}{\rotatebox[origin=c]{90}{\fontsize{18}{100} \textbf{Llama3.2-3B}}}

&  \textsc{FKV}$^{\dagger}$ &  26.0 &  40.7 &  50.4 &  32.2 &  29.6 &  15.1 &  33.5 &  22.9 &  25.3 &  71.5 &  88.9 &  3.5 &  87.8 &  52.0 &  54.2& \underline{42.2}\\

& Oracle$^{\dagger}$& 26.6 & 41.2 & 49.8 & 31.9 & 29.9 & 16.2 & 32.6 & 22.2 & 25.0 & 71.5 & 89.3 & 3.5 & 88.0 & 53.7 & 54.4 & 42.4{$\uparrow$\textsubscript{0.2}}\\
%\cdashline{2-18} 
% {\setlength{\arrayrulewidth}{2pt}\cdashline{2-18}}
& Quest & 8.7 & 19.5 & 23.6 & 12.9 & 15.9 & 6.5 & 23.3 & 18.1 & 25.1 & 34.5 & 52.9 & 6.5 & 38.3 & 53.7 & 43.6 & 25.5\textcolor{red}{$\downarrow$\textsubscript{16.7}}\\
& Stream& 13.2 & 19.7 & 23.6 & 18.1 & 22.7 & 7.8 & 18.2 & 17.9 & 17.9 & 49.0 & 83.7 & 3.5 & 85.7 & 49.3 & 45.9 & 31.8\textcolor{red}{$\downarrow$\textsubscript{10.4}}\\

& SnapKV & 23.5 & 28.9 & 45.6 & 17.7 & 22.9 & 11.8 & 21.7 & 20.9 & 21.1 & 61.0 & 88.5 & 3.5 & 88.0 & 50.7 & 48.6 & 37.0\textcolor{red}{$\downarrow$\textsubscript{5.2}}  \\
% % &TODO \\
 \rowcolor[rgb]{ .867, .922, .969}&\mymodel& \textbf{25.6} & \textbf{38.9} & \textbf{49.9} & \textbf{29.7} & \textbf{31.2} & \textbf{14.8} & \textbf{28.0} & \textbf{24.2} & \textbf{26.1} & \textbf{71.5} & \textbf{89.2} & 3.6 & 86.9 & 53.2 & \textbf{50.5} & \textbf{41.5}\textcolor{blue}{$\downarrow$\textsubscript{0.7}} \\

\specialrule{1pt}{2pt}{10pt}
\specialrule{1pt}{2pt}{2pt}
% \midrule

\multirow{6}{*}{\rotatebox[origin=c]{90}{\fontsize{18}{100}\textbf{Qwen2.5-7B}}}
&  FKV & 24.2 & 43.5 & 52.1 & 55.9 & 46.9 & 28.6 & 31.8 & 23.1 & 23.9 & 71.5 & 89.3 & 7.5 & 92.0 & 60.2 & 66.5 & \underline{47.8}  \\
& Oracle & 24.4 & 43.0 & 52.3 & 57.8 & 46.9 & 30.1 & 31.6 & 23.9 & 24.1 & 72.5 & 89.7 & 8.0 & 100.0 & 60.5 & 65.3 & 48.7$\uparrow$\textsubscript{0.9} \\
%\cdashline{2-18} 
& Quest & 9.1 & 24.5 & 30.4 & 24.7 & 24.1 & 8.8 & 26.8 & 19.9 & 24.4 & 41.8 & 66.7 & 4.4 & 77.6 & 46.5 & 42.0 & 31.4\textcolor{red}{$\downarrow$\textsubscript{16.4}}\\
& Stream& 18.1 & 24.2 & 26.5 & 41.2 & 36.4 & 17.3 & 18.4 & 18.3 & 15.4 & 45.0 & 82.9 & 8.5 & 24.0 & 49.6 & 52.2 & 31.9\textcolor{red}{$\downarrow$\textsubscript{15.9}} \\
& SnapKV & 26.6 & 36.0 & 50.8 & 55.6 & 43.8 & 26.5 & 21.9 & 21.9 & 19.3 & 58.0 & 86.2 & 8.0 & 98.5 & 55.6 & 60.6 & 42.6\textcolor{red}{$\downarrow$\textsubscript{5.2}}\\
% &TODO \\
\rowcolor[rgb]{ .867, .922, .969}& \mymodel& \textbf{28.3} & \textbf{43.8} & \textbf{51.9} & \textbf{57.4} & \textbf{46.0} & \textbf{30.1} & \textbf{31.2} & \textbf{22.8} & 24.3 & \textbf{72.0} & \textbf{89.4} & \textbf{8.0} & \textbf{99.5} &\textbf{ 60.3} & \textbf{64.0} & \textbf{47.9}\textcolor{blue}{$\uparrow$\textsubscript{0.1}}\\
% & $\text{D}_{2}\text{O}$ & \textbf{24.54} & \textbf{25.72} & \textbf{45.07} & \textbf{34.84} & \textbf{24.92} & \textbf{17.29} & \textbf{29.70} & \textbf{21.90} & \textbf{24.06} & \textbf{62.99} & \textbf{84.02}  & \textbf{4.18} & \textbf{86.26} & \textbf{55.17} & \textbf{46.15} \\

\specialrule{1pt}{2pt}{10pt}
\specialrule{1pt}{2pt}{2pt}
% \midrule

\multirow{6}{*}{\rotatebox[origin=c]{90}{\fontsize{16}{100}\selectfont \textbf{Mistral-7B-v0.3}}}
& \textsc{FKV}$^{\dagger}$ & 29.1 & 41.6 & 52.9 & 49.4 & 39.5 & 29.1 & 34.8 & 25.7 & 27.8 & 76.0 & 88.6 & 5.5 & 98.0 & 58.4 & 59.7 & \underline{47.4} \\

& Oracle$^{\dagger}$ & 31.0 & 40.2 & 52.4 & 50.3 & 39.4 & 28.8 & 34.0 & 25.74 & 27.2 & 76.0 & 89.4 & 5.0 & 98.0 & 59.3 & 61.0 & 47.9$\uparrow$\textsubscript{0.5} \\
%\cdashline{2-18} 
% \cline{2-18}
& Quest & 15.7 & 30.7 & 41.0 & 37.4 & 27.1 & 11.9 & 29.3 & 21.3 & 26.6 & 57.0 & 80.7 & 5.0 & 85.5 & 56.9 & 53.0 & 38.6\textcolor{red}{$\downarrow$\textsubscript{8.8}}\\
& Stream & 11.8&15.3&20.9&32.1&27.1&10.6&20.2&17.3&20.1&44.5&69.0&1.6&3.2&56.5&49.8&26.7 \textcolor{red}{$\downarrow$\textsubscript{20.7}} \\
& SnapKV & 25.5&32.6&53.7&48.4&37.3&25.9&22.7&23.6&23.1&62.5&89.4&6.5&94.5&57.3&57.0&44.0\textcolor{red}{$\downarrow$\textsubscript{3.4}}\\
% &TODO \\
\rowcolor[rgb]{ .867, .922, .969}& \mymodel & \textbf{29.9}&\textbf{42.3}&\textbf{53.7}&\textbf{51.1}&\textbf{39.1}&\textbf{28.7}&\textbf{34.0}&\textbf{24.8}&\textbf{28.2}&\textbf{76.0}&\textbf{89.4}&5.0&\textbf{98.0}&\textbf{57.8}&\textbf{58.0}&\textbf{47.8}\textcolor{blue}{$\uparrow$\textsubscript{0.4}} \\

\specialrule{1pt}{2pt}{10pt}
\specialrule{1pt}{2pt}{2pt}
% \midrule

\multirow{6}{*}{\rotatebox[origin=c]{90}{\fontsize{18}{100}\textbf{Llama3.1-8B}}}
&  \textsc{FKV}$^{\dagger}$ & 30.0 & 45.3 & 55.6 & 55.8 & 43.7 & 30.2 & 35.1 & 25.4 & 27.0 & 72.5 & 91.7 & 7.1 & 99.5 & 63.0 & 56.3 & \underline{48.7} \\
& Oracle$^{\dagger}$& 30.3 & 44.5 & 55.0 & 54.9 & 44.6 & 32.0 & 34.8 & 25.1 & 26.9 & 72.5 & 91.5 & 7.0 & 99.5 & 63.3 & 57.4 & 48.7$\downarrow$\textsubscript{0.0} \\
%\cdashline{2-18} 
& Quest & 13.7 & 33.1 & 38.4 & 35.8 & 32.2 & 12.8 & 26.5 & 20.9 & 26.7 & 38.0 & 65.6 & 3.8 & 95.0 & 52.5 & 45.7 & 35.4\textcolor{red}{$\downarrow$\textsubscript{13.3}}\\
& Stream& 21.9 & 23.4 & 31.8 & 45.1 & 36.7 & 24.3 & 20.0 & 21.0 & 19.3 & 45.5 & 87.9 & 6.9 & 99.5 & 59.4 & 49.1 & 38.8\textcolor{red}{$\downarrow$\textsubscript{9.9}} \\
&SnapKV & 27.5 & 34.5 & 51.6 & 52.3 & 44.3 & 28.3 & 23.9 & 24.0 & 22.7 & 62.5 & 90.9 & 7.5 & 99.5 & 60.1 & 52.6 & 45.0\textcolor{red}{$\downarrow$\textsubscript{3.7}}\\
\rowcolor[rgb]{ .867, .922, .969}& \mymodel & \textbf{29.3} & \textbf{43.7} & \textbf{54.1} & \textbf{54.8} & 43.9 & \textbf{30.8} & \textbf{33.5} & \textbf{24.7} & \textbf{27.0} & \textbf{72.0} & \textbf{91.1} & \textbf{7.5} & \textbf{99.5} & \textbf{61.8} & \textbf{52.7} & \textbf{48.2}\textcolor{blue}{$\downarrow$\textsubscript{0.5}} \\

\specialrule{1pt}{2pt}{10pt}
\specialrule{1pt}{2pt}{2pt}
% \midrule

\multirow{5}{*}{\rotatebox[origin=c]{90}{\fontsize{14.5}{100}\textbf{Qwen2.5-14B-1M}}}
&  \textsc{FKV}$^{\dagger}$ & 28.7 & 46.2 & 53.8 & 65.2 & 64.5 & 43.6 & 43.5 & 23.3 & 22.7 & 80.5 & 89.5 & 11.0 & 100.0 & 32.3 & 37.5 & \underline{50.3} \\
& Oracle$^{\dagger}$&28.5 & 46.3 & 54.3 & 64.3 & 63.6 & 44.7 & 31.5 & 22.9 & 22.7 & 81.0 & 88.4 & 10.0 & 100.0 & 33.6 & 39.7 & 49.4$\downarrow$\textsubscript{0.9}   \\
%\cdashline{2-18} 
& Quest &14.5 & 31.9 & 39.1 & 38.8 & 36.6 & 16.2 & 16.2 & 20.1 & 25.2 & 43.5 & 72.7 & 10.0 & 88.8 & 35.0 & 34.0 & 34.9\textcolor{red}{$\downarrow$\textsubscript{15.4}} \\
& Stream&19.6 & 26.9 & 29.4 & 46.5 & 48.3 & 29.6 & 17.8 & 18.4 & 15.0 & 46.5 & 82.5 & 12.5 & 72.1 & 28.7 & 31.2 & 35.3\textcolor{red}{$\downarrow$\textsubscript{15.0}}  \\
&SnapKV &26.3 & 40.5 & 51.2 & 63.2 & 62.2 & 43.3 & 22.5 & 22.0 & 18.3 & 63.5 & 87.5 & 11.5 & 100.0 & 30.4 & 36.0 & 45.9\textcolor{red}{$\downarrow$\textsubscript{4.4}} \\
% &TODO \\
\rowcolor[rgb]{ .867, .922, .969}&\mymodel &\textbf{27.2} & \textbf{45.5} & \textbf{54.5} & \textbf{64.4} & \textbf{63.9} &\textbf{44.5} & \textbf{30.4} & \textbf{22.8} & 21.9 & \textbf{80.0} & \textbf{87.5} & \textbf{15.5} & \textbf{100.0} & 30.5 & \textbf{36.1} & \textbf{49.2}\textcolor{blue}{$\downarrow$\textsubscript{1.1}}\\
\bottomrule
\end{tabular}
}
\label{table: longbench results}
\vspace{-.2in}
\end{table*}
\subsection{Performance Comparison on Long-context Tasks.}
\vspace{-0.08in}
\begin{figure}[h]
    \centering
\includegraphics[width=0.9\textwidth]{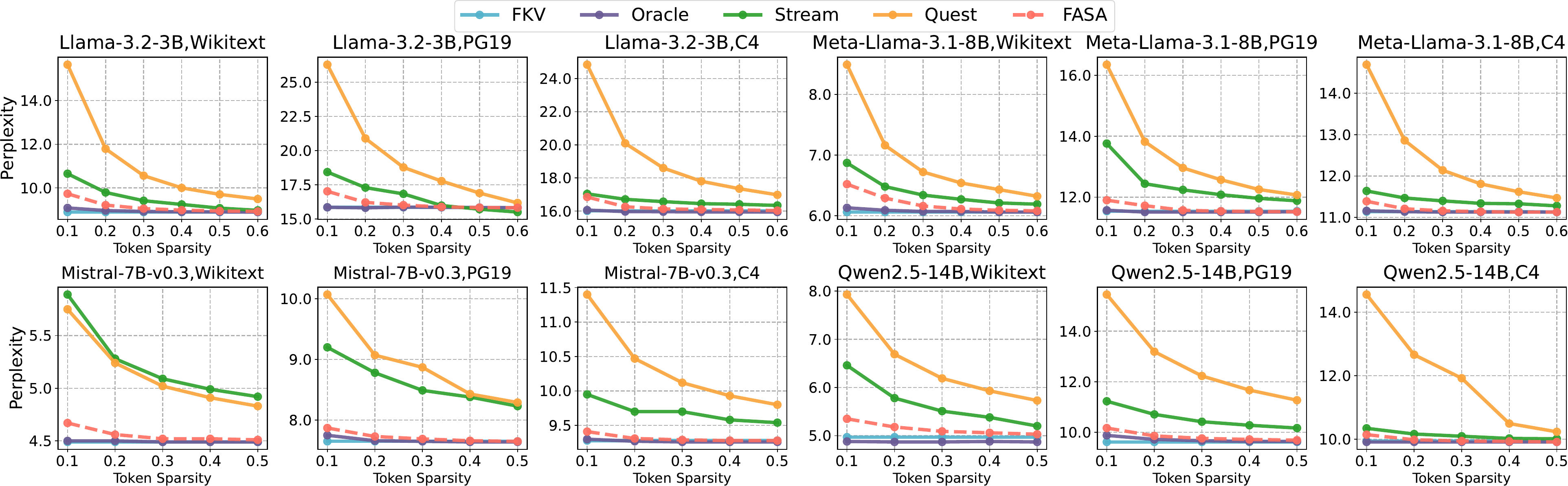}
    \vspace{-0.1in}
    \caption{\small Perplexity results of \mymodel in comparison with FKV, Oracle, Stream, and Quest on Wikitext (\textbf{top}), PG19 (\textbf{middle}), and C4 corpus (\textbf{bottom}). Token sparsity indicates the retained ratio of tokens.}
    \label{fig: ppl main results}
    \vspace{-.1in}
\end{figure}
\begin{wrapfigure}{r}{0.68\textwidth}
\vspace{-0.2in}
% \vspace{-0.5cm}
\includegraphics[width=0.98\linewidth]{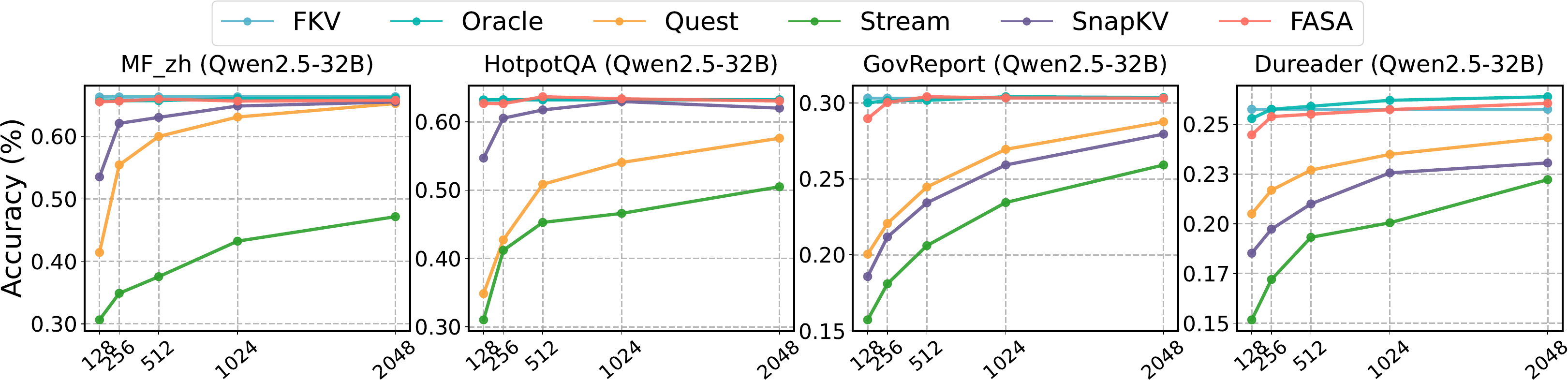} 
\vspace{-.1in}
  \caption{\small \mymodel under various token budgets ($N_{tip}=16$).}
  \label{fig: longbench}
  \vspace{-0.16in}
\end{wrapfigure}

\noindent \textbf{\mymodel achieves near-lossless performance under various budgets.}  \mymodel consistently outperforms all baselines across various budgets (Appendix~\ref{appendix: Performance Analysis on different budgets} and ~\ref{fig: longbench}), preserving contextual integrity even under extreme compression (Table~\ref{table: longbench results}). In stark contrast, existing token-eviction methods suffer catastrophic performance degradation; for instance, Quest’s accuracy plummets by 13.4\% on NarrativeQA, underscoring their inability to retain critical information.
Remarkably, under extreme budgets, \mymodel occasionally surpasses the FKV baseline (e.g., on Mistral-7B). We attribute this phenomenon to the mitigation of attentional distraction from irrelevant tokens. This hypothesis is corroborated by the Oracle baseline, which also outperforms FKV sometimes, thereby validating our frequency-chunk-based framework's efficacy in precisely identifying semantically pivotal regions.
% As shown in Figure ~\ref{appendix: Performance Analysis on different budgets},  \mymodel outperform all baselines across various token budgets. Further, results in Table \ref{table: longbench results}  underscores FASA’s transformative capability in preserving contextual integrity  under extream token budgets since the stress on token importance approximation.    token-eviction baselines, including static and adaptive methods, exhibit catastrophic drops under high compression, for instance, Quest cause even 13.4\% accuracy drop on NQA, underscoring  their failure to preserve critical context. 
% Remarkably, under extreme compression, \mymodel's performance  occasionally surpasses FKV (e.g., on Mistral-7B). We attribute this phenomenon to the mitigation of distraction from irrelevant tokens. This hypothesis is corroborated by the Oracle baseline, which also shows a performance increase over FKV, validating the efficacy of our frequency-chunk-based framework in precisely identifying semantically pivotal context. 
% By dynamically tracking these patterns, \mymodel strikes an effective balance between information retention and computational efficiency, ensuring high semantic fidelity even under extreme constraints.
\begin{table*}[t!]
\centering
\small
\renewcommand\arraystretch{0.35}
\setlength{\tabcolsep}{2pt}
\setlength{\arrayrulewidth}{0.6pt}
\caption{\small Performance and output length of \mymodel compared to baseline models on the MATH500 and AIME24  $N_{tip}\!=\!16$. AIME24 results are reported as pass@1, based on 16 responses per question. \textsc{Pref}\textsuperscript{*} and \textsc{Dec}\textsuperscript{*} denote the prefill and decoding lengths, respectively. $^{\dagger}$FKV and Oracle are full and look-ahead upper bounds.}
\vspace{-.1in}
\scalebox{0.85}{
\begin{tabular}{lp{1.3cm}cccc!{\vrule width 0.6pt}ccc!{\vrule width 0.6pt}ccccc!{\vrule width 0.6pt}ccccc}
% !{\vrule width 0.7pt}
\toprule
 &\multirow{3}{*}{\textbf{Methods}}&
\multicolumn{8}{c}{\textbf{MATH500}} &
\multicolumn{9}{c}{\textbf{AIME24}} \\
\cmidrule(r){3-9} \cmidrule(r){10-17} 
\cmidrule(r){3-9} \cmidrule(r){10-17} 
&&\multicolumn{4}{c}{\textbf{Fixed Budget}}&\multicolumn{3}{c}{\textbf{Len Stats}}&\multicolumn{5}{c}{\textbf{Fixed Budget}}&\multicolumn{3}{c}{\textbf{Len Stats}} \\
\cmidrule(r){3-6} \cmidrule(r){7-9}\cmidrule(r){10-14} \cmidrule(r){15-17}
& & 300 & 500 & 700 & 1000 &\textsc{Pref}\textsuperscript{*}&\textsc{Dec}\textsuperscript{*}&\textsc{Total}.& 500 & 1000 & 1500 & 2000 & 2500 &\textsc{Pref}\textsuperscript{*}&\textsc{Dec}\textsuperscript{*}&\textsc{Total.}\\
\midrule
 % \multirow{7}{*}{\textbf{R1-Llama-8B}}
&\multicolumn{16}{c}{\textbf{DeepSeek-R1-Distill-Llama-8B}} \\
\cmidrule(r){2-17}
&FKV$^{\dagger}$ &72.4&-&-&72.4&\multirow{7}{*}{127}&2977&3104&43.9&-&-&-&43.9&\multirow{7}{*}{161}&13231&13392\\
  &Oracle$^{\dagger}$ &70.4&72.6&74.2&71.8&&3195&3321&30.0&36.7&37.3&39.3&36.0&&15638&15799 \\
  &H2O&6.8&33.0&53.87&42.8&&8244&8370&0.7&4.7&11.3&14.0&20.0&&21099&21260 \\
 &Stream&9.6&24.6&40.4&47.4&&3520&3647&0.0&3.3&8.0&10.7&15.3&&10191&10352 \\
 &SnapKV&21.6&32.6&46.8&54.6&&7047&7174&4.0&8.0&16.0&23.3&29.1&&17359&17520\\ 
 &RKV&24.0&39.4&49.2&57.0&&7005&7132&6.7&10.7&14.0&21.7&23.3&&22916&23077 \\
\rowcolor[rgb]{ .867, .922, .969}&\mymodel&62.2&68.8&69.4&71.8&&3171&3298&20.6&34.4&40.2&35.8&38.0&&17166&17327\\
%%%%%%%%%%%%%%%%%%%%%%%%%%%%%%%%%%%%%%%%%%%%%%%%%%%%%%%%%%%%
\midrule
&\multicolumn{16}{c}{\textbf{DeepSeek-R1-Distill-Qwen-14B}} \\
\midrule
 &
 FKV$^{\dagger}$ &92.4&-&-&92.4&\multirow{7}{*}{127}&2784&2914&66.6&-&-&-&66.6&\multirow{7}{*}{165}&11039&11204\\
  &Oracle$^{\dagger}$ &92.2&92.4&92.4&92.2&&2985&3112&67.9&66.7&67.3&70.7&67.3&&11546&11711\\
  &H2O&29.6&50.2&62.8&77.0&&3413&3540&5.3&20.5&37.3&46.0&52.7&&9519&9684\\
 &Stream&27.8&44.0&57.8&64.4&&2801&2928&2.0&4.0&16.7&22.7&29.3&&8468&8633 \\
 &SnapKV&34.2&55.8&69.4&79.4&&3586&3713&10.0&23.3&40.0&46.0&52.7&&11922&12083\\
 &RKV&57.8&74.0&80.8&86.4&&3865&3992& 20.7&30.0&46.7&55.4&62.0&&16274&16439\\
\rowcolor[rgb]{ .867, .922, .969}&\mymodel&86.6&88.8&90.2&91.2&&3139&3266&54.0&60.6&59.3&62.7&63.3&&11553&11709\\
%%%%%%%%%%%%%%%%%%%%%%%%%%%%%%%%%%%%%%%%%%%%%%%%%%%%%%%%%%%%
\midrule
&\multicolumn{16}{c}{\textbf{DeepSeek-R1-Distill-Qwen-32B}} \\
\midrule
 &
 FKV$^{\dagger}$ &92.6&-&-&92.6&\multirow{7}{*}{127}&2717&2846&72.8&-&-&-&72.8&\multirow{7}{*}{156}&10461&10626\\
  &Oracle$^{\dagger}$ &92.4&91.4&91.4&91.2&&2886&3013&68.0&70.1&70.0&76.7&69.2&&11545&11710\\
  &H2O&47.2&50.0&68.3&74.4&&3841&3968&6.7&16.7&38.4&45.6&55.6&&10904&11069\\
 &Stream&43.6&57.6&65.6&73.4 &&2773&2900&0.7&6.7&18.7&23.3&24.7&&10732&10897\\
 &SnapKV&49.6&66.0&74.8&80.8&&3704&3831&10.0&23.3&40.0&46.0&52.7&&13650&13815\\
 &RKV&75.0&72.2&78.4&83.6&&4229&4356&14.7&32.7&43.3&55.3&61.3&&18078&18243 \\
\rowcolor[rgb]{ .867, .922, .969}&\mymodel&86.4&90.2&90.2&91.2&&2887&3014&60.7&62.0&66.3&70.0&73.2&&11735&11891\\
\bottomrule
% \midrule
\end{tabular}}
\vspace{-.13in}
\label{tab: long-cot-reasoning}
\end{table*}

\begin{figure}[t!]
    \centering
        \centering
        \includegraphics[width=0.9\textwidth]{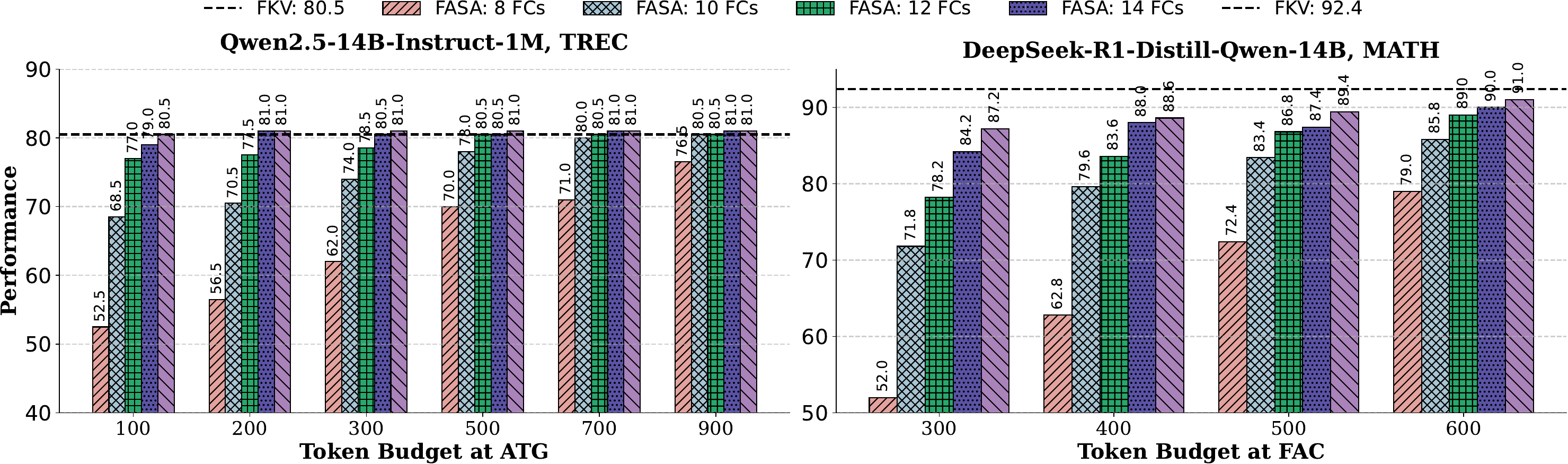}
    \vspace{-0.25cm}
    \caption{\small Evaluation of \mymodel on TREC (left) and MATH (right) datasets. The plots show the synergistic effects under varying numbers of selected FCs and different token budgets.}
    \label{fig: ablation trec}
    \vspace{-.3in}
\end{figure}

\noindent \textbf{\mymodel models complex long-term dependencies.} We simulate a \textit{token-by-token} decoding process wherein the eviction strategy is iteratively applied before token prediction. The fixed-rule approach of Stream~\citep{xiao2024efficient}, which relies on ``attention sinks,'' severely compromises its ability to capture long-range dependencies, leading to a drastic increase in perplexity as shown in Figure~\ref{fig: ppl main results}. Similarly, Quest's coarse, page-level granularity prevents it from adaptively retaining critical, non-contiguous tokens. In contrast, \mymodel's fine-grained, query-dependent mechanism accurately identifies salient tokens, achieving performance comparable to FKV, even under aggressive compression.\looseness-1

\noindent \textbf{\mymodel excels at long-CoT reasoning.} 
The chain of thought in long-form reasoning is a fragile thread, requiring the preservation of dynamically shifting "thought traces", a thread that prominent baselines consistently sever. As shown in Table~\ref{tab: long-cot-reasoning}, their static compression heuristics, blind to the evolving importance of tokens, lead to a precipitous drop in performance. On R1-Llama, SnapKV's accuracy collapses to 21.6, a stark contrast to the FKV's 72.4, demonstrating a fundamental failure to sustain the very logical dependencies required for reasoning.
Conversely, \mymodel operates with surgical precision. It surpasses not only standard baselines but also R-KV, a highly specialized method for CoT compression. It achieves an impressive 86.4\% accuracy on a scant 10\% context budget, narrowly trailing the 92.6\% FKV upper bound. This feat cements its status as a superior framework, one that can navigate the intricate web of complex reasoning without severing the essential threads of logic.\looseness-1
\subsection{In-depth Analysis}
\begin{table*}[h]
\centering
\vspace{-0.15cm}
\begin{minipage}{0.3\textwidth}
   \centering
  % \vspace{-0.4cm} % 调整与上方文本的间距
  \caption{\small Compatibility of \mymodel.}
   \vspace{-.1in}
  \label{tab: ablation_with_compatibility} % 统一了label名称
  \small % 对整个表格使用 small 字体
  \setlength{\tabcolsep}{1pt} % 调整列间距
  \renewcommand{\arraystretch}{0.35}

  % 移除了 \scalebox 和 \arraystretch，以获得更自然、更专业的排版
  \scalebox{0.75}{
  \begin{tabular}{@{}lllll@{}}
    \toprule

    \textbf{Budget}& \textbf{256} & \textbf{512} & \textbf{1024} & \textbf{2048} \\
    \midrule
    \multicolumn{5}{c}{\textbf{Qasp.}} \\
    \midrule
      \textbf{\mymodel} & 43.7 & 44.0 & 44.7 & 45.7 \\
     \textbf{+PyKV} & $44.4_{\textcolor{teal}{\uparrow 0.7}}$ & $44.5_{\textcolor{teal}{\uparrow 0.5}}$ & $45.8_{\textcolor{teal}{\uparrow 1.1}}$ & $45.8_{\textcolor{teal}{\uparrow 0.1}}$ \\
    \midrule
    \multicolumn{5}{c}{\textbf{Lcc}} \\
    \midrule
     \textbf{\mymodel} & 61.8 & 63.4 & 64.4 & 64.8 \\
     \textbf{+PyKV} & $62.2_{\textcolor{teal}{\uparrow 0.4}}$ & $63.6_{\textcolor{teal}{\uparrow 0.2}}$ & $64.7_{\textcolor{teal}{\uparrow 0.3}}$ & $64.9_{\textcolor{teal}{\uparrow 0.1}}$ \\
    \bottomrule
  \end{tabular}}
\end{minipage}
\hfill
\begin{minipage}{0.28\textwidth}
\centering
  \caption{\small Ablation on $\mathcal{K}$.}
  \vspace{-0.1in}
  \label{tab: ablation_with_k}
  \small % 对整个表格使用 small 字体
\renewcommand{\arraystretch}{0.95}
\setlength{\tabcolsep}{1pt}
\scalebox{0.85}{
  \begin{tabular}{
    @{} lcccccl @{} % 最后一列：特殊设置
  }
    \toprule
    \multirow{2}{*}{\textbf{$\mathcal{K}$}} & \multicolumn{5}{c}{\textbf{Token Budget}} & \multirow{2}{*}{\textbf{AVG.}} \\
    \cmidrule(lr){2-6}
    & {\textbf{128}} & {\textbf{256}} & {\textbf{512}} & {\textbf{1024}} & {\textbf{2048}} & \\
    \midrule
    % --- 数据行和均值计算 ---
    $128$ & 42.5 & 43.6 & \textbf{44.9} & \textbf{45.7} & \textbf{45.6} & \textbf{44.5} \\
    $256$ & 42.6 & 43.7 & 44.0 & 44.7 & 45.3 & 44.1 \\
    $512$ & 41.9 & 43.5 & 43.7 & 44.9 & 45.3 & 43.9 \\
    $1024$& 42.2 & 44.2 & 44.3 & 44.7 & 45.0 & 44.1 \\
    \bottomrule
  \end{tabular}}
\end{minipage}
\hfill
\begin{minipage}{0.36\textwidth}
    \centering % 让表格在分配的宽度内居中
  % \vspace{-.2in}
  \caption{\small Ablation of offline calibration.}
  \label{tab:ablation_offline_calibration}
  \renewcommand{\arraystretch}{0.2}
  \vspace{-.1in}
  % --- 核心改动 ---
  % 1. 将字体命令从 \small 改为 \footnotesize
  \footnotesize 
  % 2. 因为字体变小，表格变窄，可以适当增加列间距，让其不那么拥挤
  \setlength{\tabcolsep}{.7pt} % 从 2pt 增加到 3pt
  \scalebox{0.77}{
  \begin{tabular}{@{}lcccccc@{}}
    \toprule
    \multirow{2}{*}{\textbf{Offline}}& \multicolumn{3}{c}{\textbf{S-Doc QA}} & \multicolumn{3}{c}{\textbf{M-Doc QA}} \\
    \cmidrule(lr){2-4} \cmidrule(lr){5-7}
     & \textbf{2Wiki} & \textbf{Musi} & \textbf{Hqa} & \textbf{Qasp.} & \textbf{MF\_en} & \textbf{Nqa} \\
    \midrule
    \textbf{Base}    & 43.7 & 30.2 & 55.8 & 45.3 & 55.6 & 29.9 \\
    \midrule
    \textbf{Nqa }    & 44.5 & 31.6 & 55.0 & 44.2 & 55.8 & 29.2 \\
    \textbf{Qasp.}  & 43.0 & 31.0 & 54.1 & 44.0 & 54.6 & 29.1 \\
    \textbf{Musi} & 43.8 & 30.8 & 55.1 & 44.8 & 54.6 & 29.6 \\
    \textbf{Self}    & 43.5 & 30.8 & 55.3 & 43.9 & 54.4 & 29.2 \\
    \midrule
    \textbf{CV}      & .014 & .012 & .010 & .009 & .011 & .007 \\
    \bottomrule
  \end{tabular}}
\end{minipage}
\vspace{-0.15cm}
\end{table*}
\noindent\textbf{Effect on Generation Length.}
% \il{move this paragraph right after ``FASA excel at long-CoT reasoning.'' either within Section 5.3 or before.}
A neglected aspect of compression methods is the impact on output length. Some compression methods, like H2O, induce generative verbosity, imposing an overlooked computational burden (Table~\ref{tab: long-cot-reasoning}). Conversely, others, such as Stream, prematurely terminate generation, which truncates valid reasoning and degrade performance. In contrast, \mymodel maintains output lengths nearly identical to the FKV  while preserving high performance, demonstrating a superior balance.

\textbf{Compatiblility of \mymodel.} By design, \mymodel is orthogonal to and synergistic with other KV cache optimization paradigms. We demonstrate this by integrating it with PyramidKV~\citep{cai2025pyramidkvdynamickvcache}, which allocates varied budgets across layers. While PyramidKV determines how many tokens to keep per layer, \mymodel{} decides which tokens are most critical. As shown in Table~\ref{tab: ablation_with_compatibility}, this complementary pairing yields consistent performance gains, confirming \mymodel's high compatibility and modularity.

% \input{tables/ablation_diff_calibra}

% \noindent\textbf{Discussion on Design Choices.}
% To further clarify the design principles of \mymodel, we address two potential simplifications and explain why they are infeasible.  \textbf{(1) Can FC-based scores directly replace attention?} No. While our FC-based scores ($\mathbf{S}_t^{l,h}$) are an excellent proxy for identifying token importance, we found that using them directly as attention weights for generation leads to a significant performance collapse. Their utility lies in ranking, not in precise value substitution. \textbf{(2) Can individual dimensions serve as selection units?} Again, no. If we were to select "dominant dimensions" instead of "dominant FCs" and apply a similar pipeline, the model's performance would be completely destroyed. This result strongly suggests that the Frequency Chunk (FC), which groups two dimensions sharing the same frequency, is the fundamental functional unit for this mechanism. This is intrinsically linked to the design of RoPE, where pairs of dimensions are coupled to represent positional information.
\begin{wrapfigure}{r}{0.4\textwidth}
% \vspace{-0.2in}
\vspace{-0.5cm}
\includegraphics[width=0.98\linewidth]{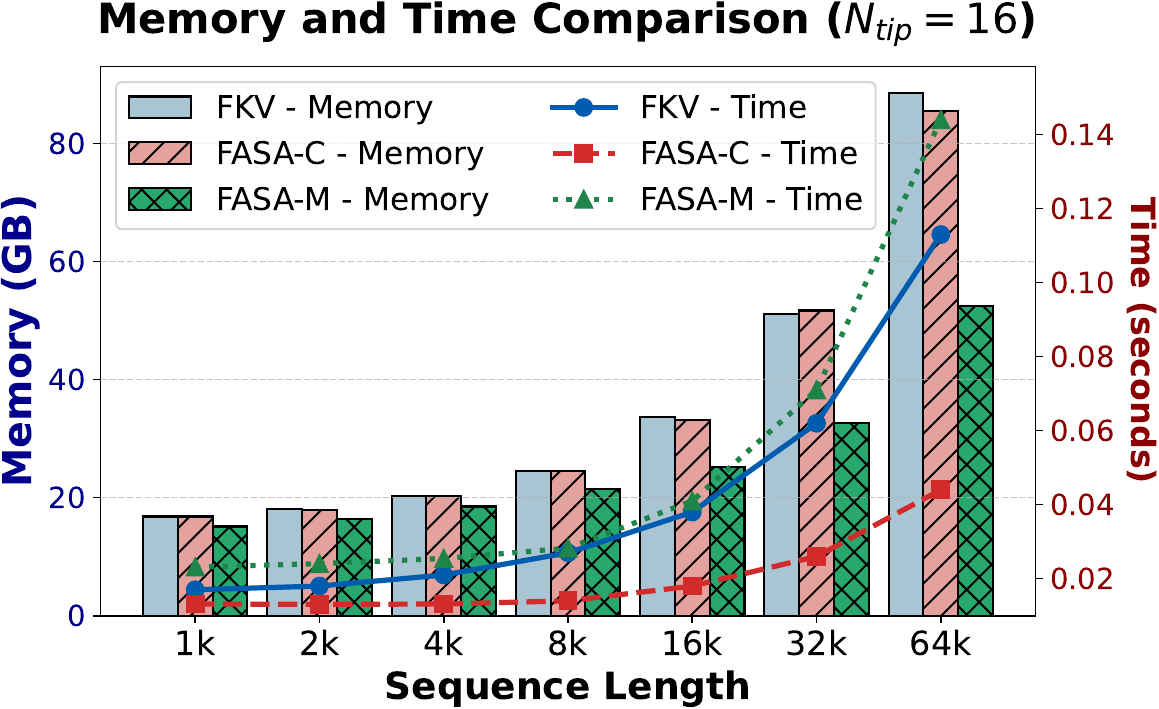} 
\vspace{-.1in}
  \caption{\small Memory vs. latency ($N_{tip}=16$).}
  \label{fig: memory analysis}
  \vspace{-0.25in}
\end{wrapfigure}
\textbf{Efficiency Analysis.}
% \paragraph{Efficiency Assessment}
We assess the efficiency of our two \mymodel variants.
\mymodel-M's memory savings are particularly pronounced in long sequences, as the KV cache's footprint grows to dominate and dwarf the static memory costs of model parameters and activations. While its CPU-GPU data transfer introduces a slight latency overhead, this can be effectively mitigated by prefetching techniques that asynchronously load the required KV pairs in advance.
\mymodel-C, implemented with  Triton (based on ~\cite{ribar2024sparq}),  delivers substantial inference acceleration. The speedup effect intensifies with longer sequences, achieving up to a 2.56$\times$ with $N_{tip} = 16$ under 64K.
\vspace{-.18in}
\subsection{Ablation Studies}
\noindent\textbf{Robustness to Calibration Window $\mathcal{K}$.}
Our method exhibits remarkable robustness to the calibration window size, $\mathcal{K}$. Performance is largely insensitive to $\mathcal{K}$, with smaller $\mathcal{K}$ values often yielding slightly superior results (Table~\ref{tab: ablation_with_k}). This suggests that due to the inherent sparsity of attention, even a small calibration window provides a sufficiently robust signal to identify the dominant FCs.
% \il{reorder ablations so that Table 5 is referenced before Fig. 6 and then Fig. 7?}
% \vspace{-.2in}

\noindent\textbf{Trade-off between $N_{tip}$ and $N_{fac}$.}
The hyperparameters $N_{{tip}}$ (token selection precision) and $N_{fac}$ (retention budget) govern a trade-off between the fidelity of token identification and the volume of retained context. As depicted in Figure~\ref{fig: ablation trec}, optimal performance can be achieved either with high-precision selection (large $N_{{tip}}$) and a small budget, or a more lenient selection (small $N_{{tip}}$) compensated by a larger one. Empirically, on the TREC dataset, we found that using just 10 dominant FCs (15.6\% of dimensions) with  $N_{fac}=500$ is sufficient to match the FKV's performance.

\noindent\textbf{Impact of Offline Calibrated Data.}
As shown in Table~\ref{tab:ablation_offline_calibration}, our method exhibits remarkable robustness to the choice of calibration data. The minimal performance variation across different calibration datasets, as quantified by a low Coefficient of Variation (CV), confirms that our FC detection mechanism is stable and not reliant on a specific calibration source.

\section{Extending FASA to Non-RoPE Models}
In this section, we investigate the generalizability of \mymodel to non-RoPE architectures. The feasibility of this extension hinges on whether functional sparsity is a universal emergent property induced by alternative PE schemes. We first ascertain the presence of functional sparsity in ALiBi and Partial-RoPE, followed by an empirical evaluation of \mymodel’s performance within these frameworks.

\begin{figure}[htbp]
    \centering
    \resizebox{0.83\textwidth}{!}{ % 控制整个 figure 环境的宽度为页面的80%
        % 第一个子图
        \begin{subfigure}[b]{0.49\textwidth}
            \centering
            \includegraphics[width=\textwidth]{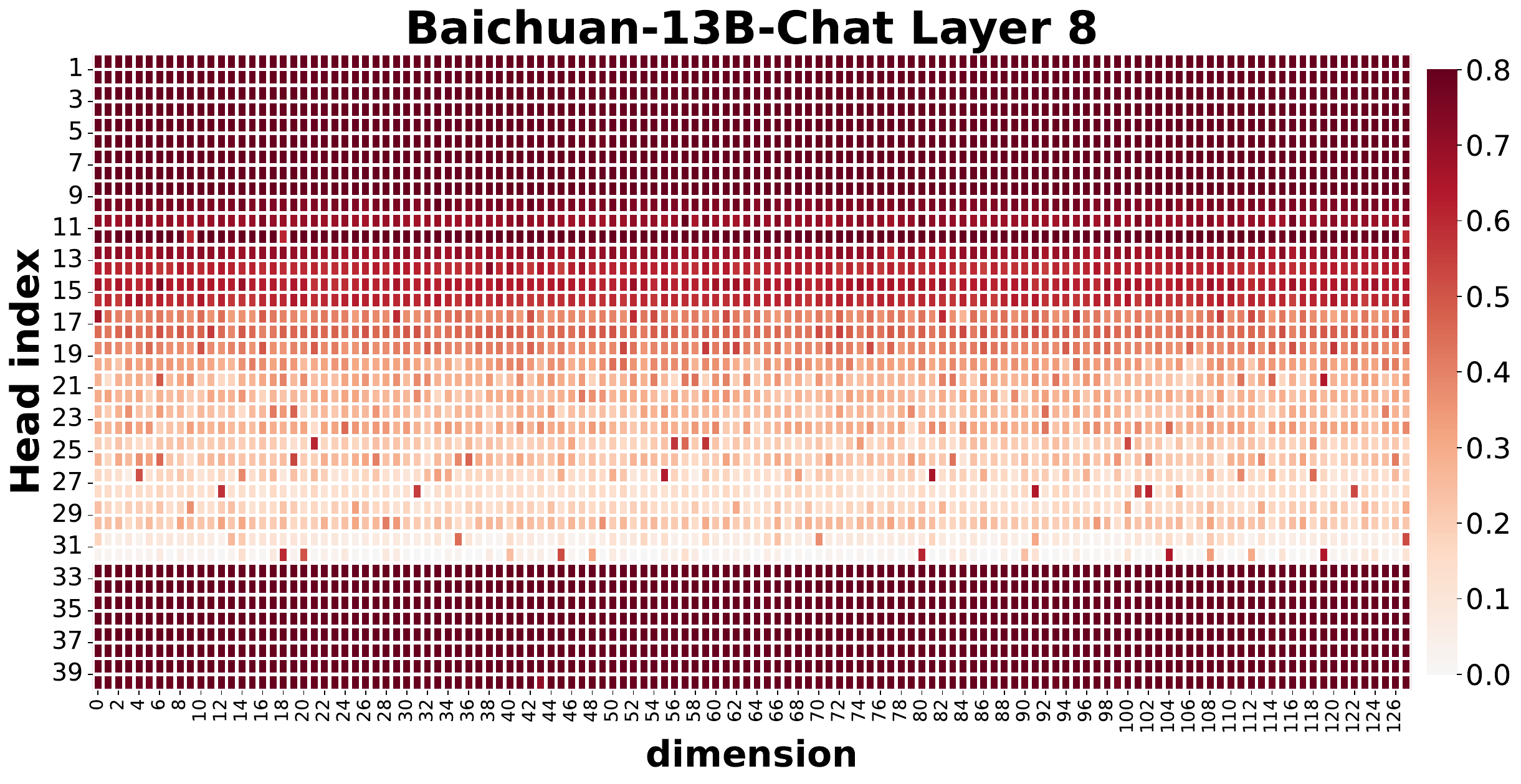}
            \label{fig:layer8} % 为子图添加标签，方便引用
        \end{subfigure}
        \hfill % 横向排列，用 \hfill 分隔
        % 第二个子图
        \begin{subfigure}[b]{0.49\textwidth}
            \centering
            \includegraphics[width=\textwidth]{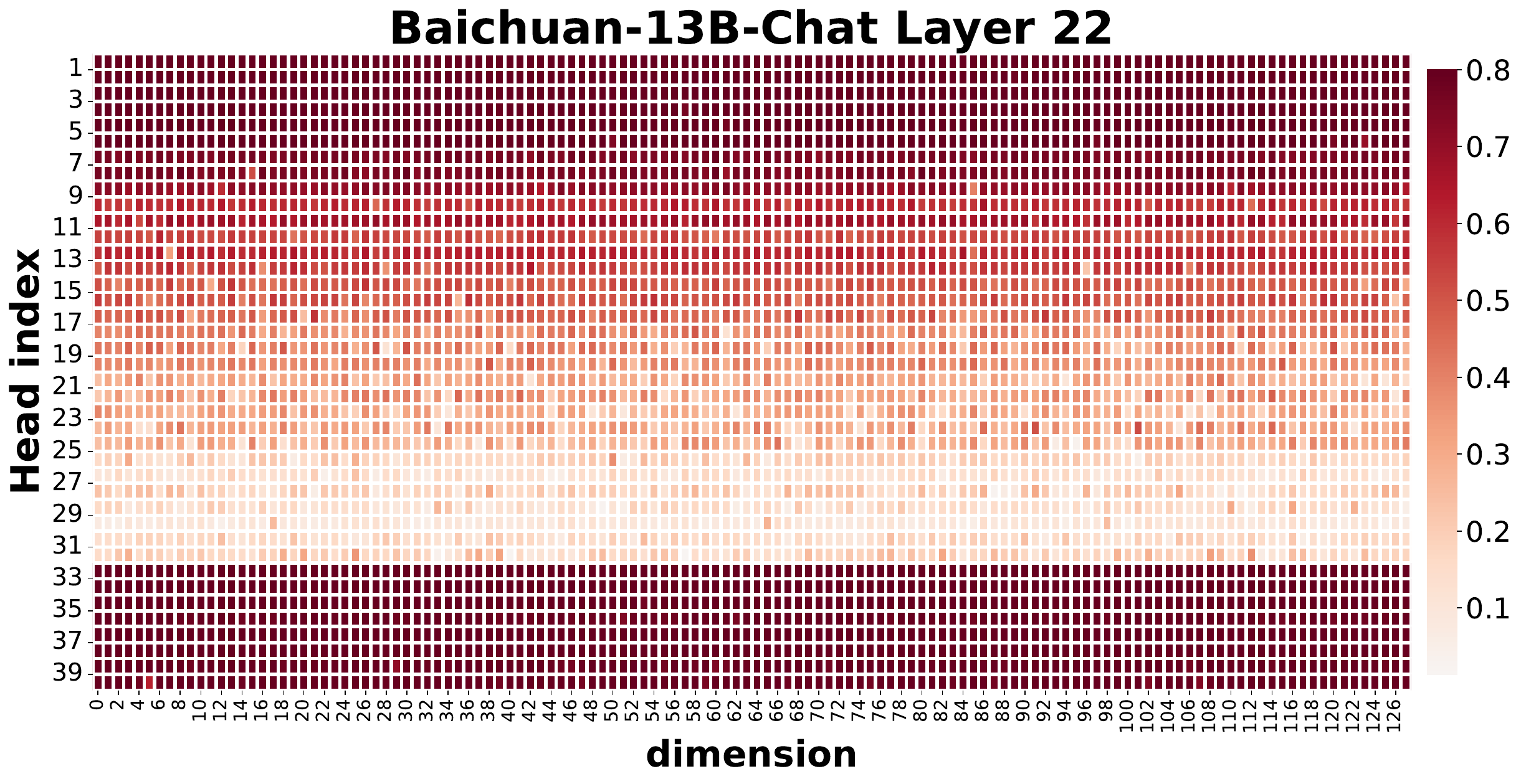}
            \label{fig:layer22}
        \end{subfigure}
    }
    \vspace{-.2in}
    \caption{CA Scores Heatmaps of Baichuan-13B-Chat (ALiBi models).}
    \label{fig: functional sparsity on norope alibi}
\end{figure}

\begin{figure}[htbp]
    \centering
    \resizebox{0.83\textwidth}{!}{ % 控制整个 figure 环境的宽度为页面的80%
        % 第一个子图
        \begin{subfigure}[b]{0.49\textwidth}
            \centering
            \includegraphics[width=\textwidth]{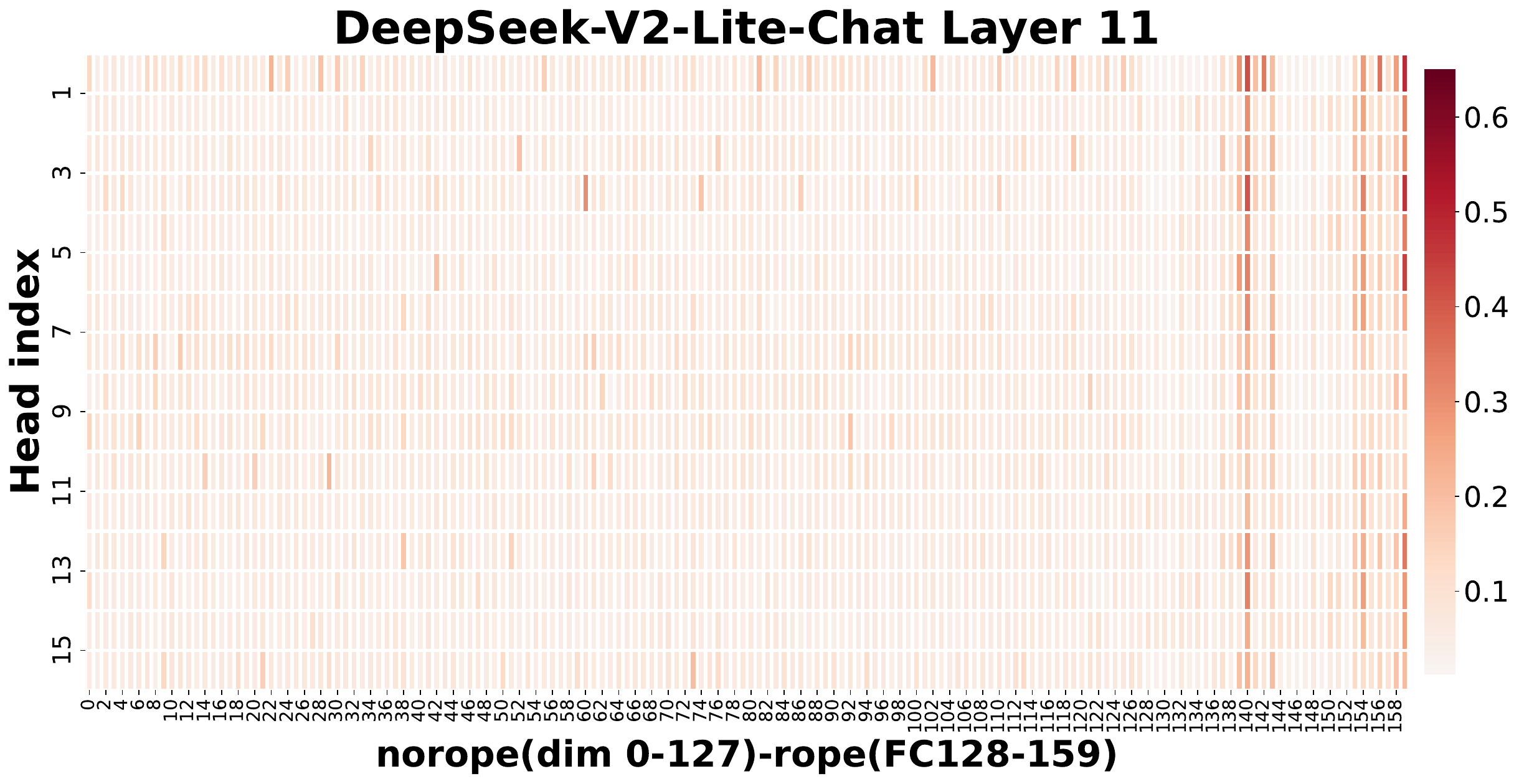}
            \label{fig:layer11} % 为子图添加标签，方便引用
        \end{subfigure}
        \hfill % 横向排列，用 \hfill 分隔
        % 第二个子图
        \begin{subfigure}[b]{0.49\textwidth}
            \centering
            \includegraphics[width=\textwidth]{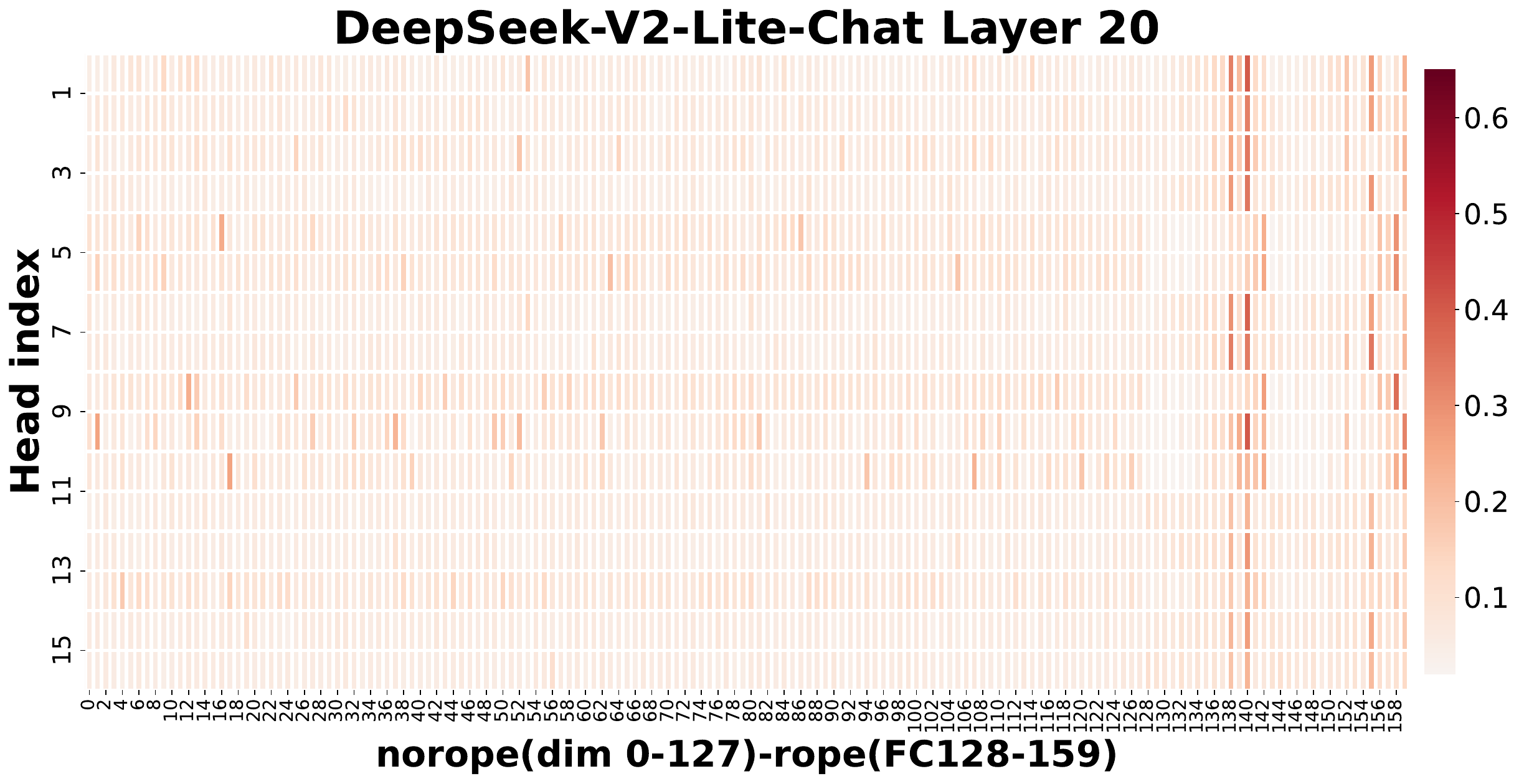}
            \label{fig:layer20}
        \end{subfigure}
    }
    \vspace{-.2in}
    \caption{CA Scores Heatmaps of DeepSeek-V2-Lite-Chat (Partial-RoPE models).}
    \label{fig: functional sparsity on norope mla}
\end{figure}

\paragraph{Functional Sparsity in ALiBi and Partial-RoPE (MLA)}
We extend our analysis to two widely used PE variants: Attention with Linear Biases (ALiBi) and Partial-RoPE (in Multi-head Latent Attention (MLA)). While ALiBi encodes relative distances using head-specific linear biases added to the attention logits, MLA adopts a decoupled approach where RoPE is applied only to a specific partition of the head dimensions. This strategy allows us to explore sparsity behaviors that go beyond the standard full-RoPE configurations. This strategy enables us to examine sparsity behaviors beyond the typical full-RoPE configurations. As illustrated in Figures \ref{fig: functional sparsity on norope alibi} and \ref{fig: functional sparsity on norope mla}, both ALiBi and Partial-RoPE induce functional sparsity at the head dimension level. For ALiBi, heads indexed from 19 to 31 exhibit a regular pattern of functional sparsity, while the remaining dimensions in other heads show high CA scores (approximately 0.7). Therefore, FASA is compatible with ALiBi.

\vspace{-.25in}

\paragraph{FASA Evaluation on Other PEs}
As demonstrated in Tables \ref{table: norope_fasa_eval_mla} and \ref{table: norope_fasa_eval_alibi}, FASA exhibits exceptional generalizability across diverse position encoding architectures, achieving this without incurring any significant performance trade-offs. The results consistently match or surpass those of FKV, \emph{establishing FASA as a robust, high-performance method with broad applicability beyond RoPE.}

\begin{table*}[!h] % 使用 [htbp] 给予 LaTeX 更多放置灵活性
\centering % 这个 \centering 是可选的，如果两个 minipage 总宽度小于 \textwidth，它会让整个组合居中。
\begin{minipage}[t]{0.47\textwidth} % 使用 [t] 确保两个表格的顶部对齐

  \caption{Performance on Partial-RoPE Models.} % 更具描述性的标题
  \vspace{-.1in}
  \label{tab:partial_rope}
  \setlength{\tabcolsep}{1pt} % 设置一个合理的列间距，比 1pt 可读性好
  \renewcommand{\arraystretch}{1.1} % 设置一个合理的行高，0.35太小了
  \small % 使用 small 字体
  \begin{tabular}{lcccccc}
    \toprule
    & \textbf{Qasper} & \textbf{2Wiki} & \textbf{Multi} & \textbf{Passage\_Re} & \textbf{Lcc} & \textbf{Samsum} \\
    \midrule
    FKV     & 33.18           & 19.83             & 47.27                 & 49.00                 & 63.40        & 34.04           \\
    FASA    & 33.46           & 20.25             & 46.50                 & 48.50                 & 62.49        & 32.53           \\
    \bottomrule
  \end{tabular}
  
  \label{table: norope_fasa_eval_mla}
\end{minipage}%  <-- 这个百分号 % 很重要，可以防止产生不必要的空格
\hfill % 在两个 minipage 之间填充弹性空间
\begin{minipage}[t]{0.47\textwidth} % 稍微增加宽度以容纳内容

  \caption{Performance on ALiBi Models.}
  \vspace{-.1in}
  \label{tab:alibi}
  \setlength{\tabcolsep}{1pt} % 根据内容调整列间距
  \renewcommand{\arraystretch}{1.1} % 保持行高一致
  \small
  \begin{tabular}{lccccc} % 注意这里是 6 列
    \toprule
    & \textbf{Qasper} & \textbf{Lsht} & \textbf{Dureader} & \textbf{Trec} & \textbf{Repobench} \\ % 调整了列数
    \midrule
    FKV     & 9.11            & 24.25         & 23.18             & 23.00         & 17.30              \\
    FASA    & 7.80            & 21.25         & 21.70             & 21.50         & 16.46              \\
    \bottomrule
  \end{tabular}
  \label{table: norope_fasa_eval_alibi}
\end{minipage}
\vspace{-.15in}
\end{table*}
\section{Conclusion}
In this work, we addressed the memory footprint and bandwidth introduced by the KV cache in LLMs. Firstly, we cover an intriguing phenomenon: the functional sparsity of  FCs. A subset of dominant FCs could show high contextual awareness. Based on this discovery, we introduce \mymodel, 
a coarse-to-fine two-stage freamwork. The first stage utilizes the dominant FCs to perform dynamic, query-aware token selection without costly training. Then, the second stage perform focused and precise attention computation on this reduced subset.
Our experiments indicate that \mymodel attains performance nearly on par with full KV even under constrained budgets. The memory- and speed-optimized variants of \mymodel offers a practical and effective solution for efficient long-context inference.

\section*{Acknowledgements}

We thank the anonymous reviewers for their insightful comments and suggestions. We also thank the members of the Machine Learning Group at AMAP for their valuable feedback and support throughout the development of this work.

\section*{Ethics Statement}
Our research is focused on enhancing the computational efficiency of Large Language Model (LLM) inference by optimizing KV cache management. The primary positive impact of our work, \mymodel, is to make large-scale models more accessible, affordable, and environmentally sustainable. By significantly reducing memory and computational overhead, our method can enable researchers and institutions with limited resources to develop and deploy powerful long-context models, thereby fostering broader innovation and democratization in the field of AI.

We acknowledge the dual-use nature of efficiency-enhancing technologies. While our goal is positive, lowering the barrier to running large models could inadvertently make it easier for malicious actors to deploy them for harmful purposes, such as generating misinformation or spam at scale. It is important to note, however, that our work is foundational and does not create new capabilities for generating harmful content; it merely optimizes the performance of existing models.

All experiments were conducted on publicly available benchmarks (LongBench, MATH, AIME) and open-source pre-trained models. We did not use any private, sensitive, or user-generated data. We recognize that the foundation models used in our evaluation may reflect and perpetuate societal biases present in their vast training corpora. Our method operates orthogonally to the challenge of model-level bias and does not address it directly, but we encourage users to be mindful of the inherent limitations of the models they deploy with our technique.

\section*{Reproducibility Statement}
To ensure the reproducibility of our work, we provide a detailed account of all models, datasets, experimental setups, and evaluation protocols, all of which are publicly available. An overview of the experiments is provided in \textbf{Section~\ref{section: Experimental Setting}}, with more comprehensive details described across several appendices.
Specifically, the configurations for all baselines and the detailed hyperparameters for \mymodel are presented in \textbf{Appendix~\ref{appendix: Experiment Configurations.}}. The descriptions of all benchmarks and their corresponding evaluation protocols are detailed in \textbf{Appendix~\ref{appendix: benchmark_details}} and \textbf{Appendix~\ref{appendix: Evaluation Metrics}}, respectively. Furthermore, the implementation and design choices for \mymodel{} are explained in \textbf{Appendix~\ref{appendix: Implement Details}}. Finally, the specific algorithms for \mymodel-M and other core functions are provided in \textbf{Appendix~\ref{varian of fasa}} and \textbf{Appendix~\ref{appendix: Algorithm on fasa}}.

\bibliography{iclr2026_conference}
\bibliographystyle{iclr2026_conference}

\clearpage
\appendix
\section{Investigation Results of Dominant Frequency Chunks} \label{part1: sparse fcs investigations}
\subsection{Further Generalization on Model Scales and Architechtures } \label{appendix: fucntional sparsity generalizes model scales and architechtures}
\begin{figure}[h]
    \centering
    % --- 第一行 (4个子图) ---
    % \begin{subfigure}[b]{0.24\textwidth}
    %     \centering
    %     \includegraphics[width=\textwidth]{figures/sparsity_phenomenon/llama3.2-3b-256-22-qasper.pdf}
    % \end{subfigure}%
    \begin{subfigure}[b]{0.48\textwidth}
        \centering
        \includegraphics[width=\textwidth]{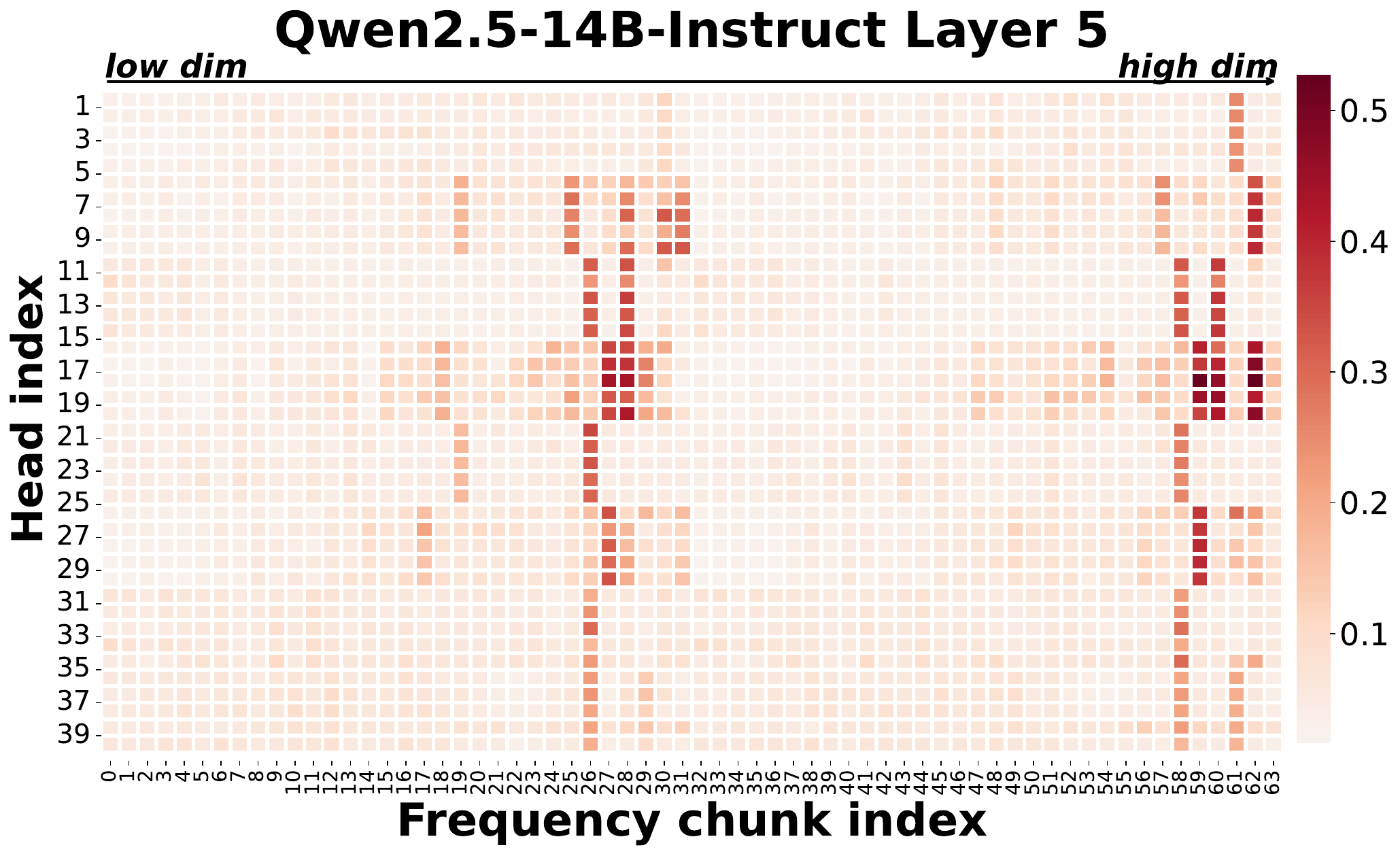}
    \end{subfigure}%
    \begin{subfigure}[b]{0.48\textwidth}
        \centering
        \includegraphics[width=\textwidth]{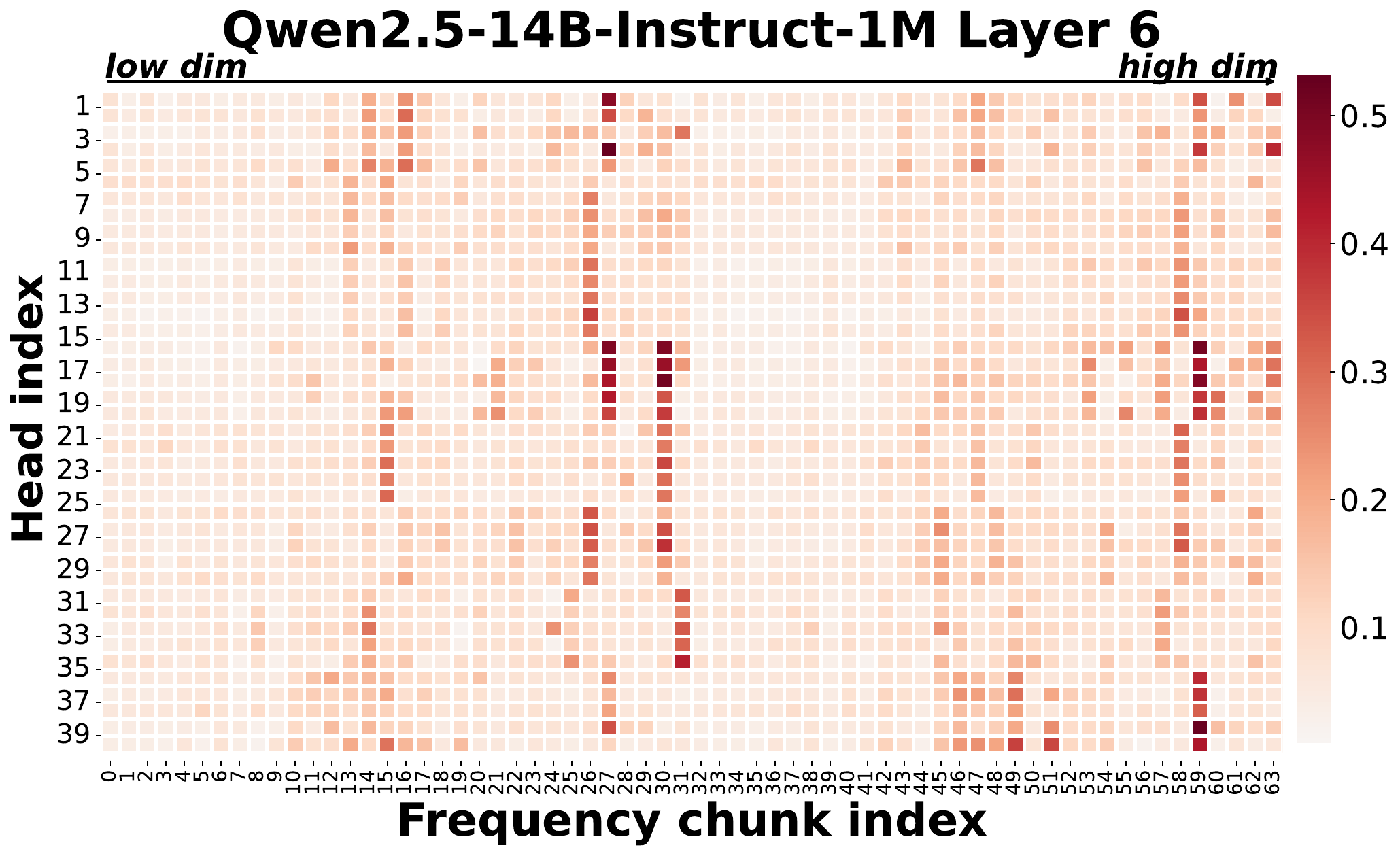}
    \end{subfigure}%
    \vspace{-0.1cm}
    \caption{{Functional sparsity is maintained on Qwen2.5 series models~\citep{DBLP:journals/corr/abs-2412-15115}.} 
        Heatmaps visualize the Mean Contextual Agreement ($\overline{\text{CA}}_{K=256}$) for each Frequency Chunk (FC, x-axis) across all attention heads (y-axis) in a representative layer. 
        We compare the standard \textbf{Qwen2.5-14B-Instruct} model (left) with its long-context variant, \textbf{Qwen2.5-14B-Instruct-1M} (right), both calibrated on the Qasper dataset.  The remarkable similarity between the two heatmaps demonstrates that the functional sparsity of FCs is a robust property, consistently maintained even after long-context fine-tuning.}
\label{fig: parse_frequency_qasper generalization on model architechtures}
%\vspace{-0.6cm}
\end{figure}
\begin{figure}[h]
    \centering
    % --- 第一行 (4个子图) ---
    % \begin{subfigure}[b]{0.24\textwidth}
    %     \centering
    %     \includegraphics[width=\textwidth]{figures/sparsity_phenomenon/llama3.2-3b-256-22-qasper.pdf}
    % \end{subfigure}%
    \begin{subfigure}[b]{0.48\textwidth}
        \centering
        \includegraphics[width=\textwidth]{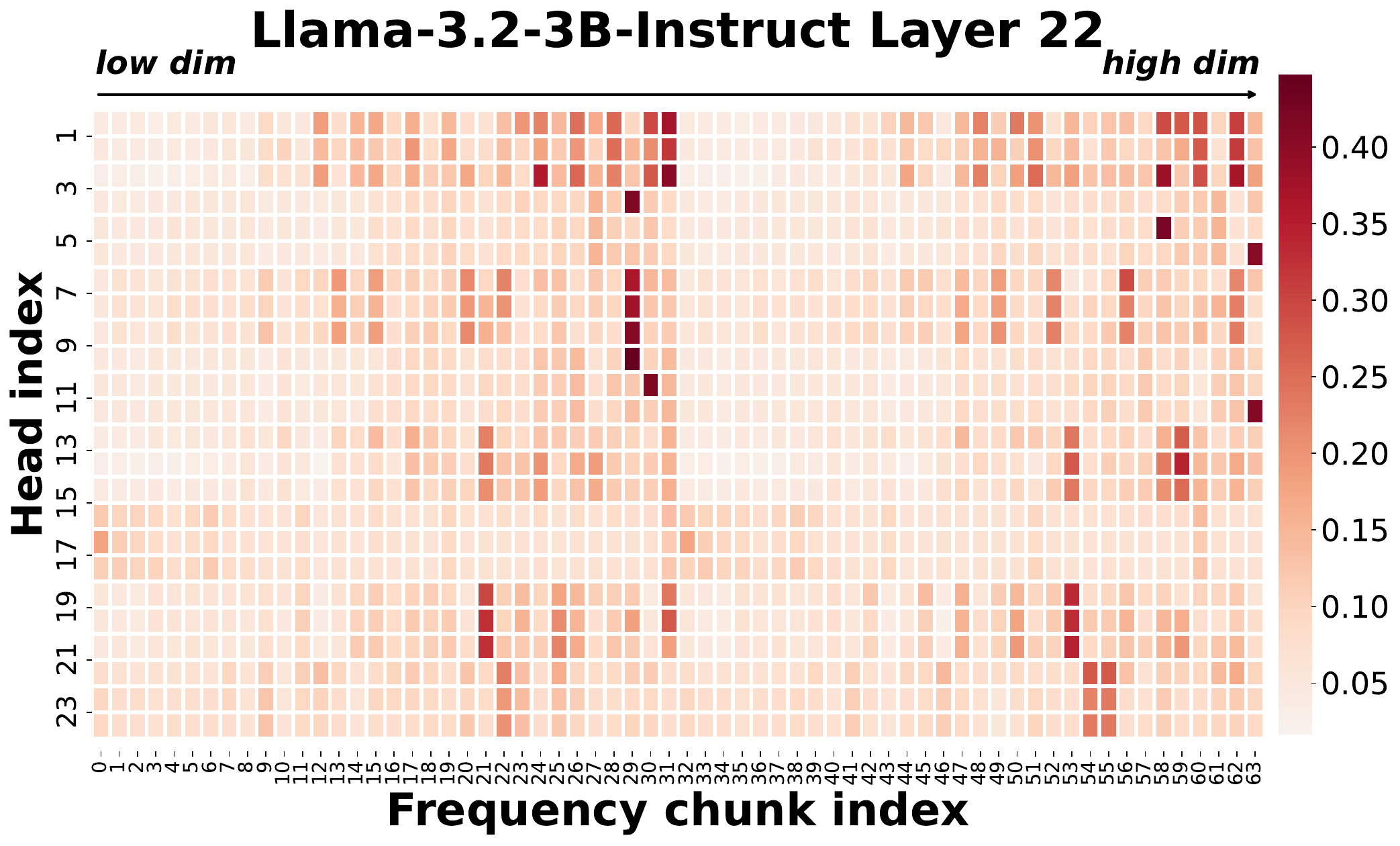}
    \end{subfigure}%
    \begin{subfigure}[b]{0.48\textwidth}
        \centering
        \includegraphics[width=\textwidth]{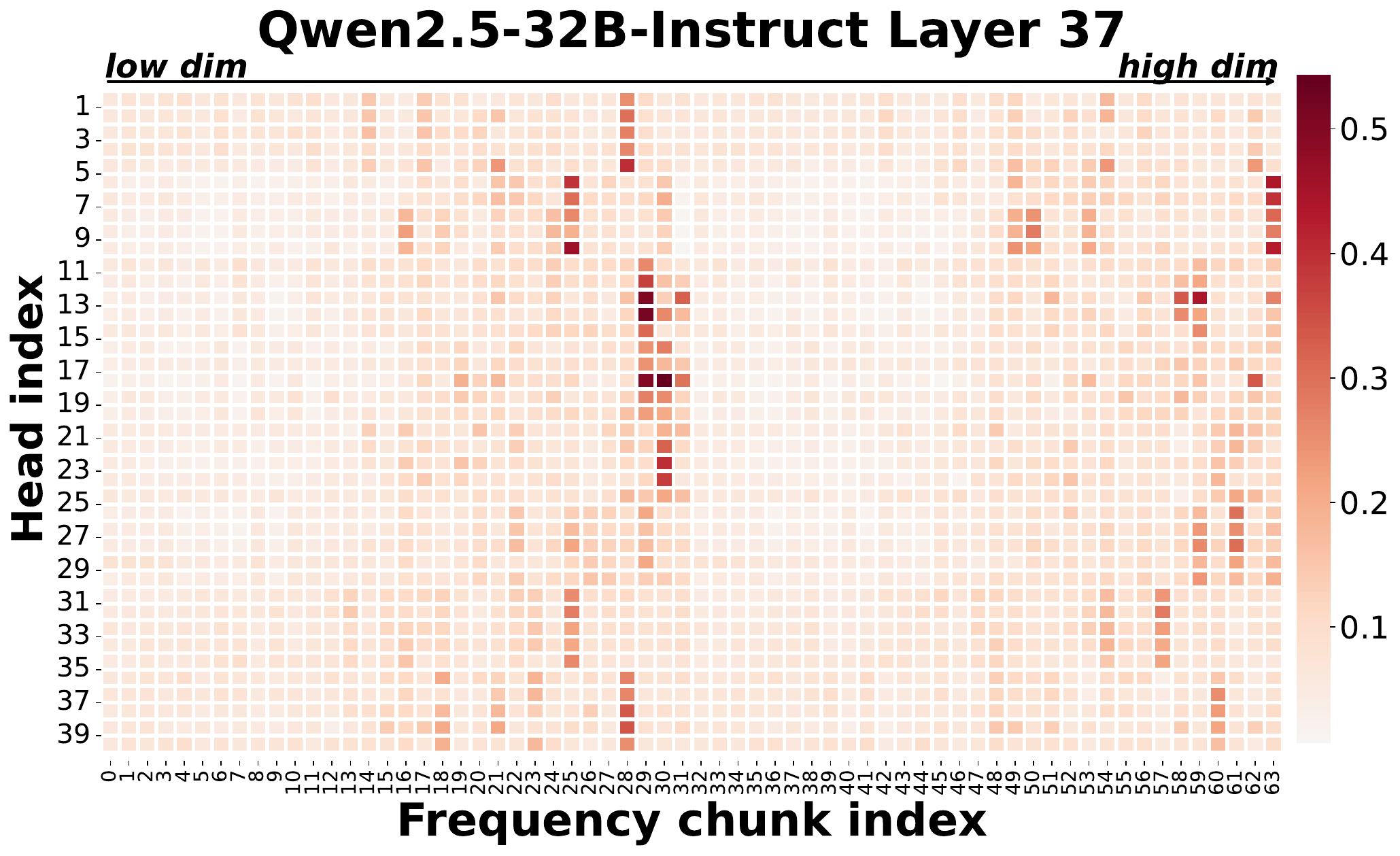}
    \end{subfigure}%
    \vspace{-0.1cm}
    \caption{{Functional sparsity persists across model scales.} Heatmaps show the Mean Contextual Agreement ($\overline{\text{CA}}_{K=256}$) for increasing scale (3B and 32B). 
        The remarkable stability of the dominant FC patterns (bright vertical columns) across these scales demonstrates that functional sparsity is a fundamental and scalable characteristic of RoPE. }
\label{fig: parse_frequency_qasper generalization on model scales}
%\vspace{-0.6cm}
\end{figure}
\textbf{Conclusions:} Our cross-architectural (Figure ~\ref{fig: parse_frequency_qasper generalization on model architechtures}) and cross-scale (Figure \ref{fig: parse_frequency_qasper generalization on model scales}) analysis reveals a striking finding: the functional sparsity of FCs is a universal and stable property. This powerful evidence suggests that the observed functional hierarchy is not an emergent artifact of a specific model's training dynamics or size, but rather an intrinsic characteristic deeply embedded within the RoPE mechanism itself. The roles of different frequencies appear to be fundamental and pre-determined, providing a robust and predictable foundation for developing model-agnostic efficiency optimizations.
\subsection{Task-Invariance Property of Functional Sparsity }
\begin{figure}[ht]
    \centering
    \begin{subfigure}[b]{0.48\textwidth}
        \centering
        \includegraphics[width=\textwidth]{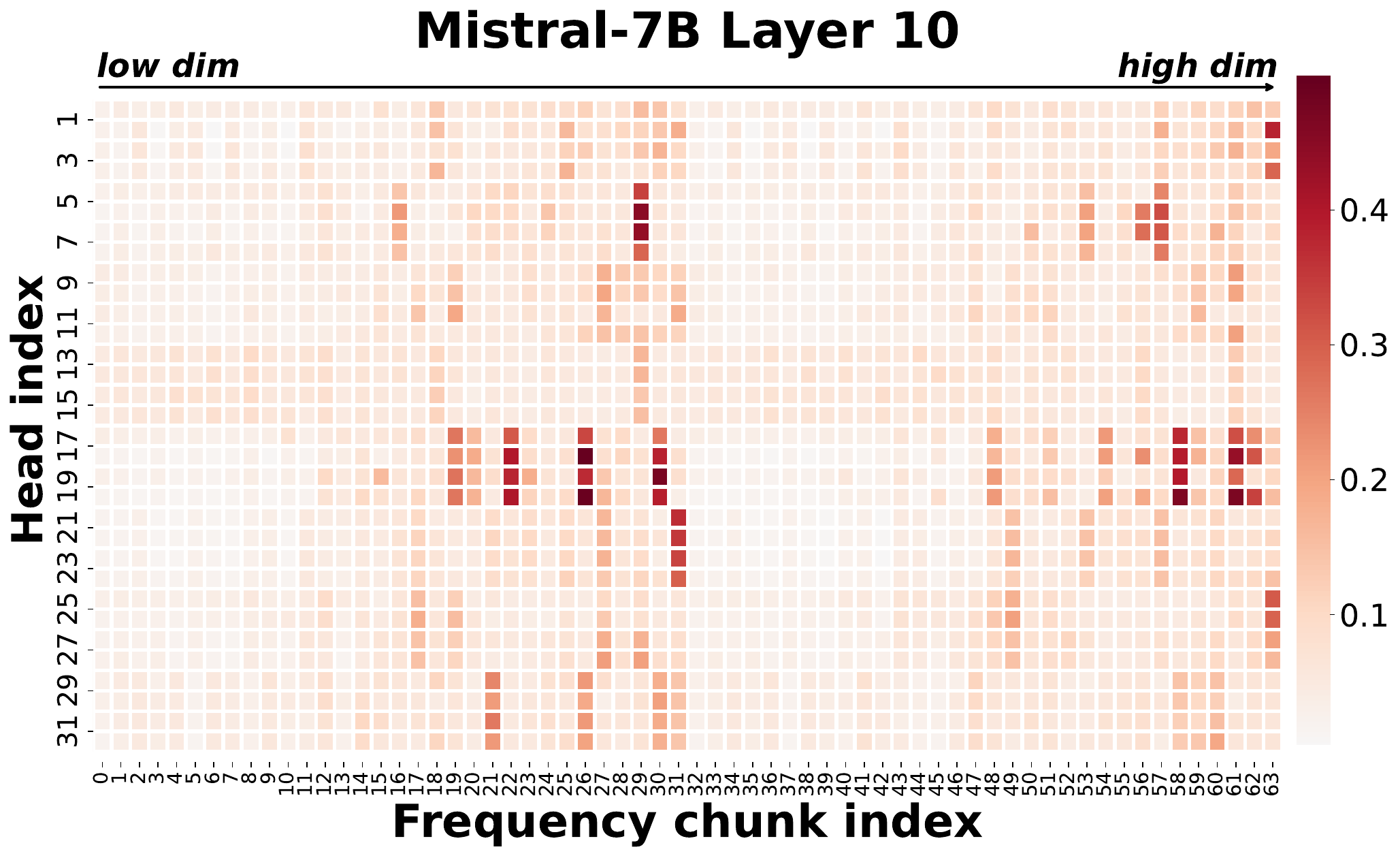}
        \caption{Qasper}
    \end{subfigure}%
    \begin{subfigure}[b]{0.48\textwidth}
        \centering
        \includegraphics[width=\textwidth]{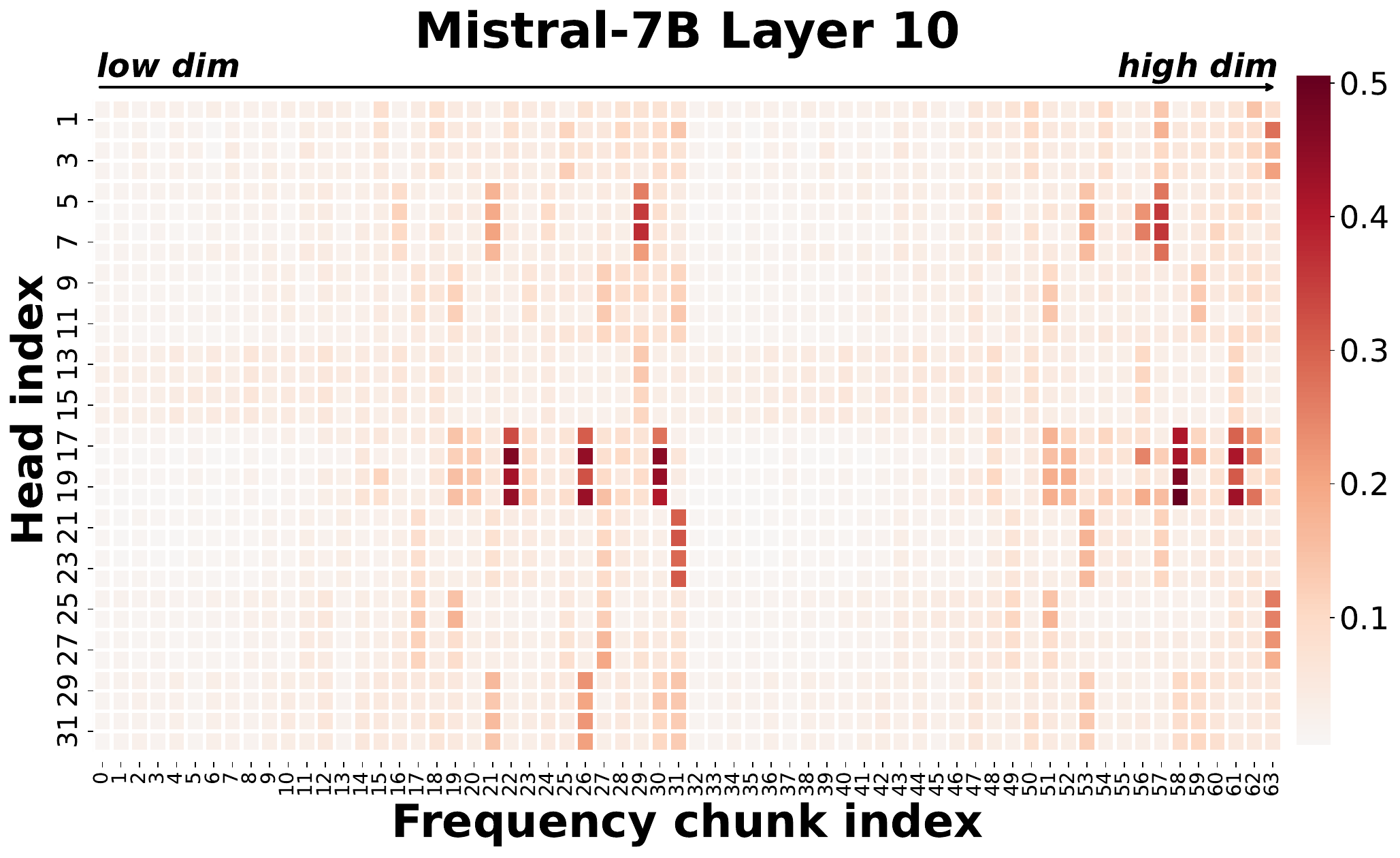}
        \caption{GovReport}
    \end{subfigure}%
    \vspace{-0.1cm}
    \caption{Heatmaps of agreement score ($\overline{\text{CA}},K=256$) across attention heads for the Qasper (Left) and GovReport (Right) from LongBench-V1 (\cite{bai2024longbench}) on Mistral-7B-Instruct-v0.3.}
\label{fig: parse_frequency_qasper task-agnotic}
%\vspace{-0.6cm}
\end{figure}

We find that the saliency of dominant FCs is largely task-agnostic. This property is evidenced by the strong alignment between saliency maps generated for distinct downstream tasks, as shown in Figure \ref{fig: parse_frequency_qasper task-agnotic}. Despite the functional differences between question answering (left) and summarization (right), the resulting importance rankings are highly consistent. This indicates that these FCs perform a fundamental role inherent to the model's architecture, rather than one adapted for a specific task.
\begin{figure}[ht]
    \centering
    \begin{subfigure}[b]{0.2\textwidth}
        \centering
        \includegraphics[width=\textwidth]{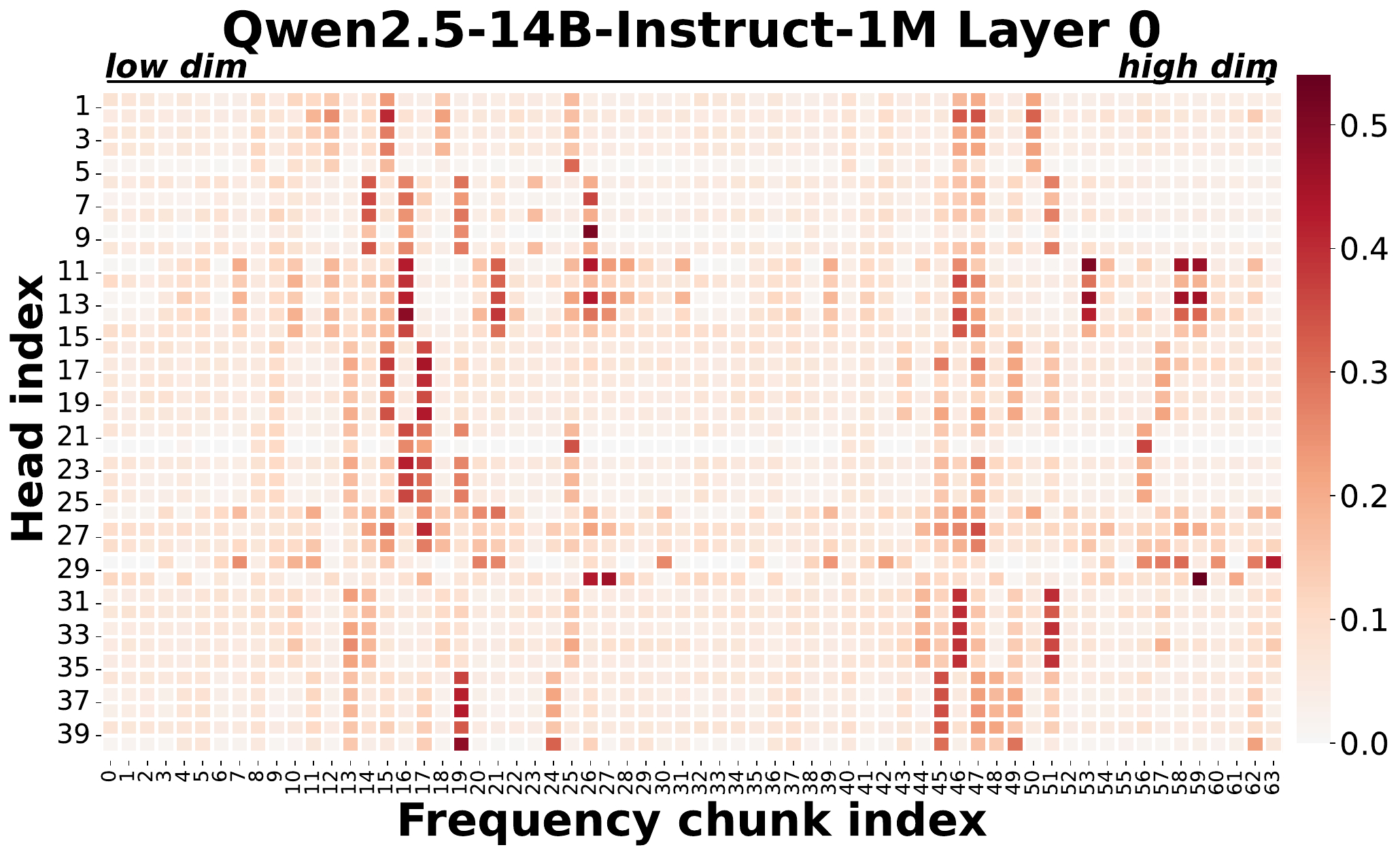}
    \end{subfigure}%
    \begin{subfigure}[b]{0.2\textwidth}
        \centering
        \includegraphics[width=\textwidth]{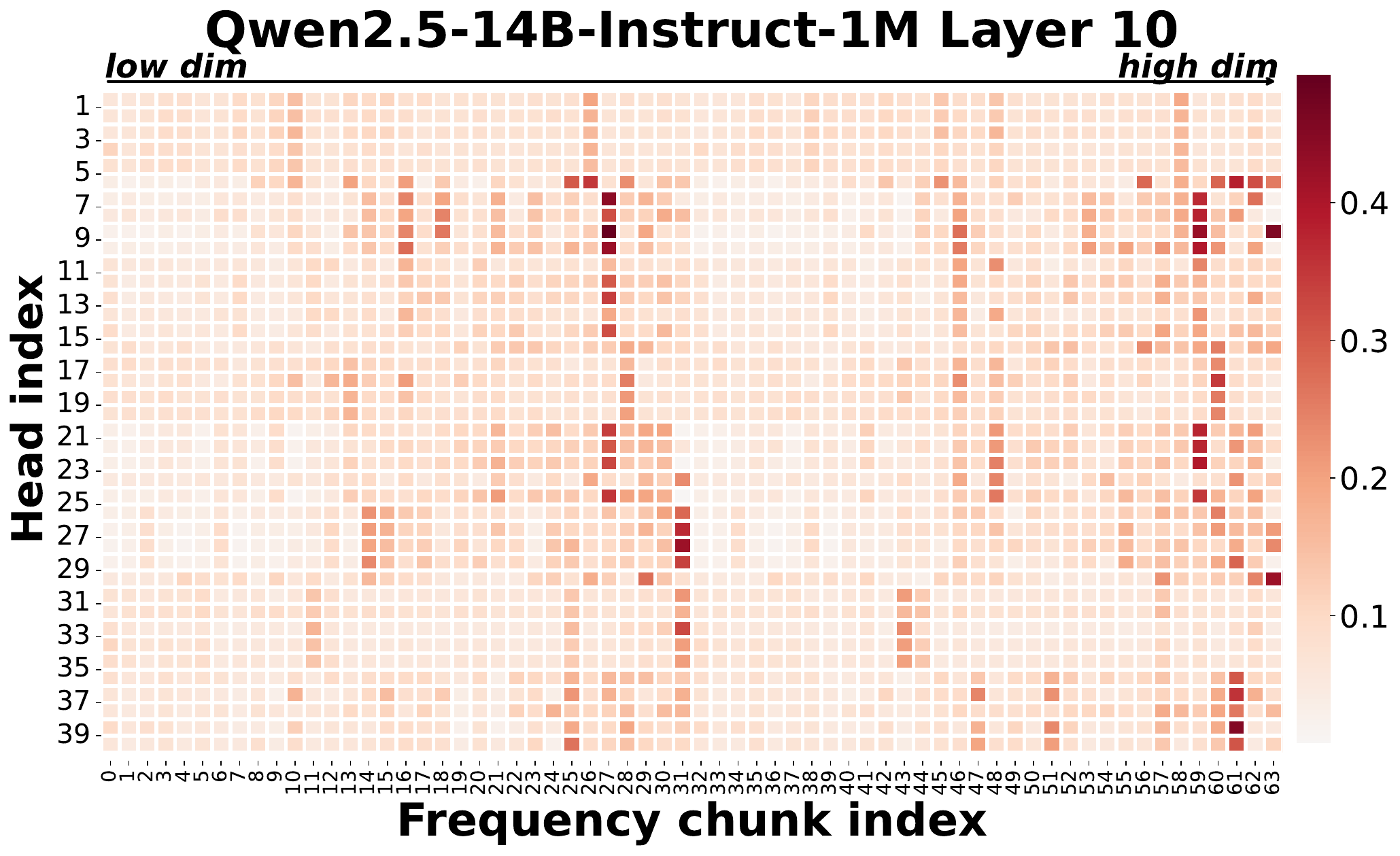}
    \end{subfigure}%
    \begin{subfigure}[b]{0.2\textwidth}
        \centering
        \includegraphics[width=\textwidth]{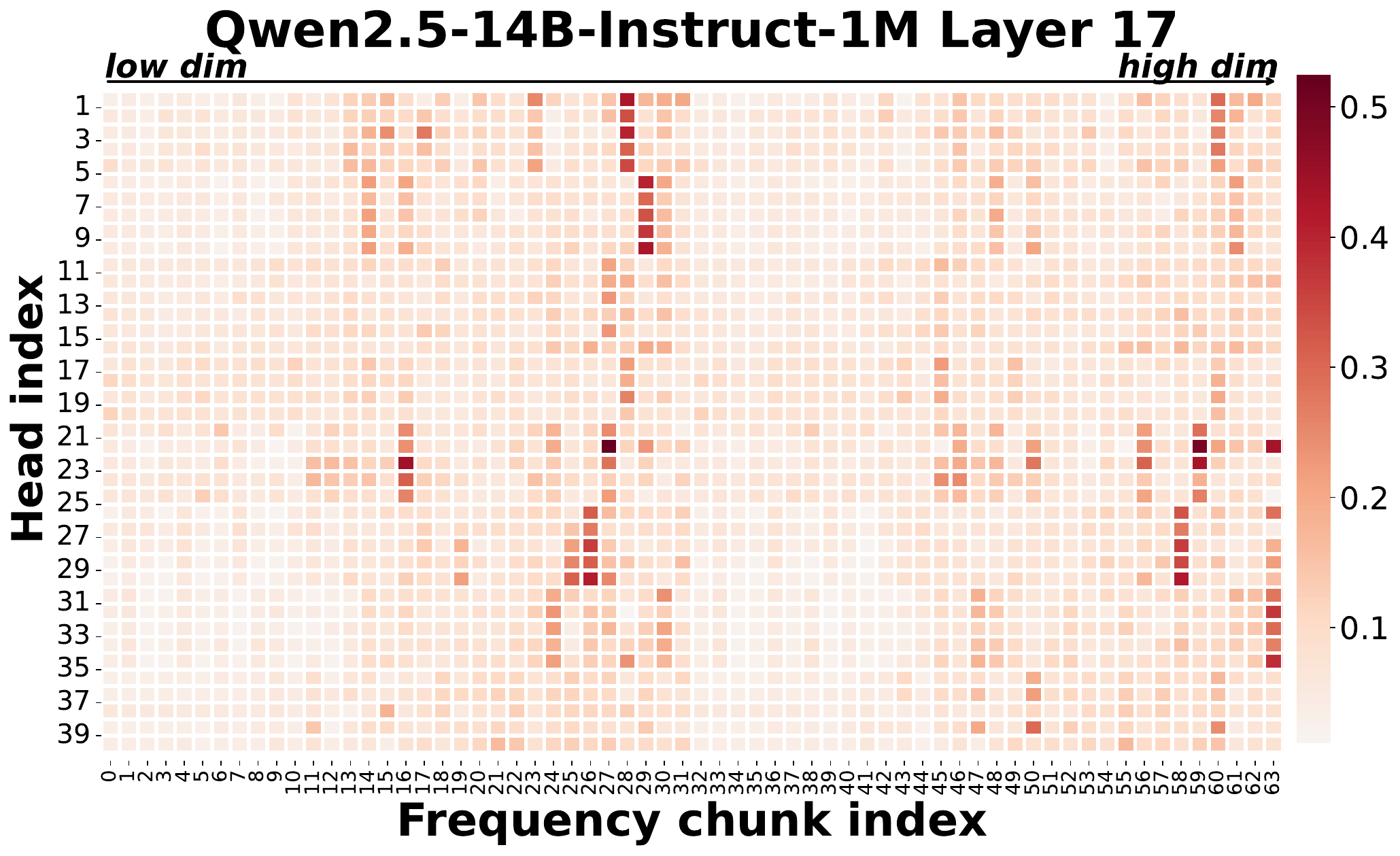}
    \end{subfigure}%
    \begin{subfigure}[b]{0.2\textwidth}
        \centering
        \includegraphics[width=\textwidth]{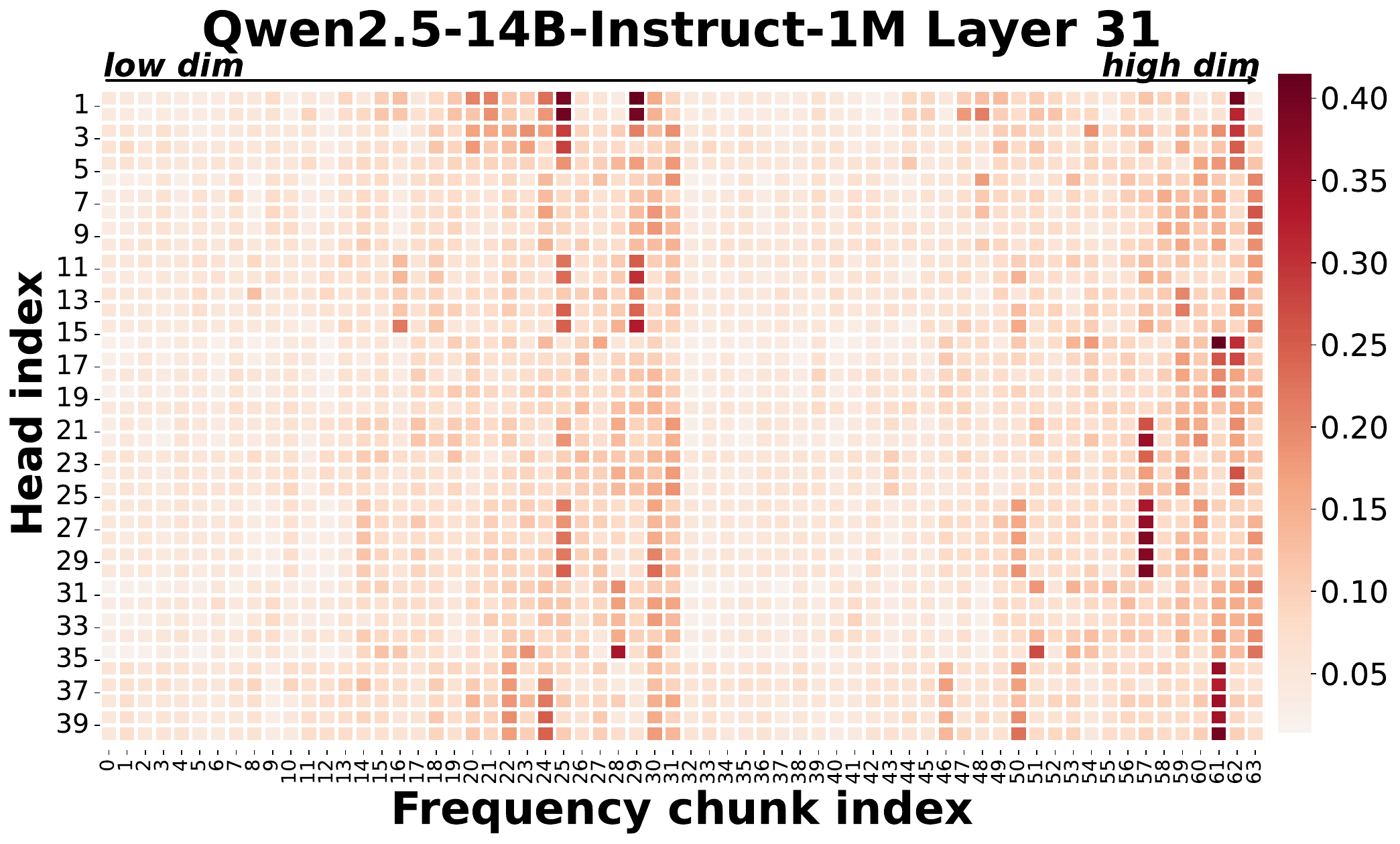}
    \end{subfigure}%
    \begin{subfigure}[b]{0.2\textwidth}
        \centering
        \includegraphics[width=\textwidth]{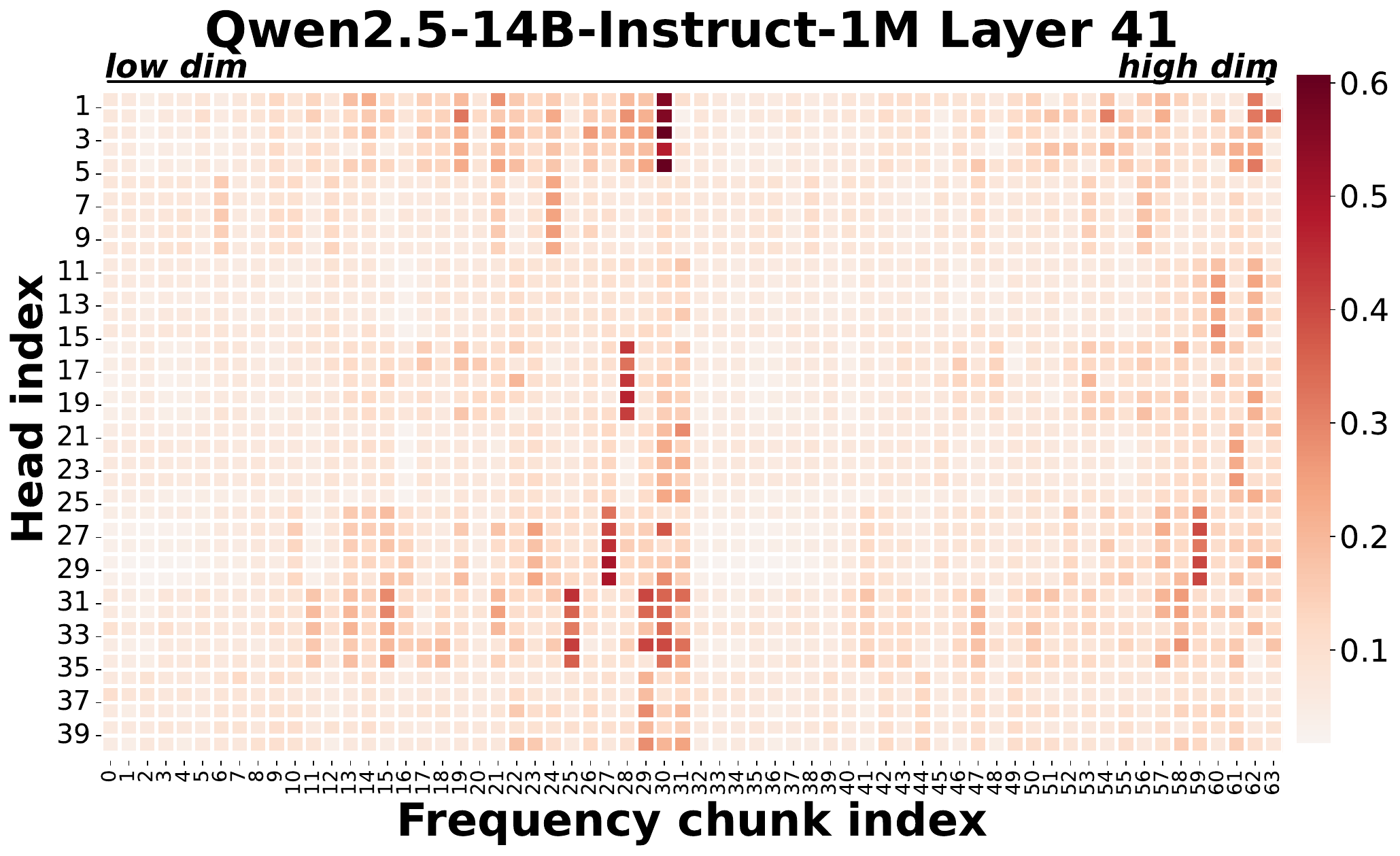}
    \end{subfigure}%
    \vspace{-0.1cm}
    \caption{Heatmaps of agreement score ($\overline{\text{CA}},K=256$) across different layers.}
\label{fig: parse_frequency_qasper across layers}
%\vspace{-0.6cm}
\end{figure}
\subsection{More Analysis Results}
\paragraph{Functional Sparsity across Layers.} While the principle of functional sparsity is universal, the specific set of dominant FCs is far from static in Figure~\ref{fig: parse_frequency_qasper across layers}; instead, it exhibits a high degree of specialization across both model depth and individual attention heads. This dynamic behavior reveals a sophisticated division of labor within the transformer architecture.

\subsection{Quantitative Evidence on Sparsity \& Universality \& Task-
Invariance}
\label{appendix: Quantitative Evidence on Sparsity and Universality and Task-
Invariance}

\begin{table}[h]
\centering
\caption{The ratio of dominant FCs and non-dominant FCs.}
\begin{tabular}{lcc}
\toprule
\textbf{Type of FC} & \textbf{Dominant FCs (\%)} & \textbf{Non-Dom FCs (\%)} \\ 
\midrule
\textbf{Model} & \textbf{CA scores $>$ 0.4} & \textbf{CA score $<$ 0.15} \\
\midrule
Llama-3.2-3B          & 0.54 & 89.6 \\
Meta-Llama-3.1-8B     & 0.68 & 89.6 \\
Mistral-7B-v0.3       & 0.68 & 92.7 \\
Qwen2.5-7B            & 0.17 & 95.5 \\
Qwen2.5-14B           & 0.27 & 94.7 \\
Qwen2.5-14B-1M        & 0.65 & 90.5 \\
Qwen2.5-32B-Instruct  & 0.52 & 91.2 \\
R1-Distill-Llama-8B  & 0.79 & 89.5 \\
R1-Distill-Qwen-14B  & 0.76 & 90.2 \\
R1-Distill-Qwen-32B  & 0.67 & 90.9 \\
\bottomrule
\end{tabular}
\label{tab: quantitative analysis about sparsity}
\end{table}
\begin{table*}[h]
\centering
\caption{Cross-task overlap matrix of dominant FCs (\%). Each sub-table shows the percentage of intersection between dominant FCs identified on a "row" dataset  and a "column" dataset. }
\resizebox{\textwidth}{!}{% This command scales the entire table to fit the text width
\begin{tabular}{@{}ll cccccc@{}}
\toprule
\textbf{Model} & \textbf{Overlap of Dom-FCs} & \textbf{Qasper} & \textbf{Gov\_Report} & \textbf{Musique} & \textbf{Narrativeqa} & \textbf{2Wikimqa} & \textbf{Avg.} \\
\midrule
% Llama-3.2-3B Block
\multirow{5}{*}{\textbf{Llama-3.2-3B}} & \textbf{Qasper} & 100.00 & 75.90 & 82.30 & 70.50 & 83.20 & 82.38 \\
& \textbf{Gov\_Report }& 75.90 & 100.00 & 82.10 & 70.80 & 81.90 & 82.14 \\
& \textbf{Musique} & 82.30 & 82.10 & 100.00 & 73.60 & 96.50 & 86.90 \\
& \textbf{Narrativeqa} & 70.50 & 70.80 & 73.60 & 100.00 & 73.10 & 77.60 \\
& \textbf{2Wikimqa} & 83.20 & 81.90 & 96.50 & 73.10 & 100.00 & 86.94 \\
\midrule
% Mistral-7B Block
\multirow{5}{*}{\textbf{Mistral-7B}} & \textbf{Qasper} & 100.00 & 71.10 & 77.10 & 67.30 & 77.00 & 78.50 \\
& \textbf{Gov\_report} & 71.10 & 100.00 & 79.40 & 65.50 & 78.90 & 78.98 \\
& \textbf{Musique} & 77.10 & 79.40 & 100.00 & 67.80 & 97.90 & 84.44 \\
& \textbf{Narrativeqa} & 67.30 & 65.50 & 67.80 & 100.00 & 67.30 & 73.58 \\
& \textbf{2Wikimqa} & 77.00 & 78.90 & 97.90 & 67.30 & 100.00 & 84.22 \\
\midrule
% Qwen2.5-7B Block
\multirow{5}{*}{\textbf{Qwen2.5-7B}} & \textbf{Qasper} & 100.00 & 70.60 & 80.90 & 68.70 & 81.30 & 80.30 \\
& \textbf{Gov\_Report} & 70.60 & 100.00 & 79.40 & 68.20 & 78.70 & 79.38 \\
& \textbf{Musique} & 80.90 & 79.40 & 100.00 & 71.70 & 96.60 & 85.72 \\
& \textbf{Narrativeqa} & 68.70 & 68.20 & 71.70 & 100.00 & 71.10 & 75.94 \\
& \textbf{2Wikimqa} & 81.30 & 78.70 & 96.60 & 71.10 & 100.00 & 85.54 \\
\midrule
% Qwen2.5-14B Block
\multirow{5}{*}{\textbf{Qwen2.5-14B}} & \textbf{Qasper} & 100.00 & 69.20 & 84.30 & 71.80 & 84.50 & 81.96 \\
& \textbf{Gov\_Report} & 69.20 & 100.00 & 75.00 & 67.60 & 74.80 & 77.32 \\
& \textbf{Musique} & 84.30 & 75.00 & 100.00 & 74.30 & 98.40 & 86.40 \\
& \textbf{Narrativeqa} & 71.80 & 67.60 & 74.30 & 100.00 & 73.90 & 77.52 \\
& \textbf{2Wikimqa} & 84.50 & 74.80 & 98.40 & 73.90 & 100.00 & 86.32 \\
\bottomrule
\end{tabular}
}
\label{tab:cross-task overlap}
\end{table*}
\begin{table*}[h]
\centering
\caption{Predictive distribution of dominant FCs across different attention score ranges.}
\label{tab:performance_distribution}
\resizebox{\textwidth}{!}{% This command scales the table to fit the text width
\begin{tabular}{@{}ll ccccc@{}}
\toprule
& & \multicolumn{5}{c}{\textbf{Prediction accuracy across varying attention scale ranges}} \\
\cmidrule(l){3-7}
\textbf{Model} & \textbf{Type of FCs} & \textbf{Top 20\%} & \textbf{Top 20-40\%} & \textbf{Top 40-60\%} & \textbf{Top 60-80\%} & \textbf{Top 80-100\%} \\
\midrule
\multirow{2}{*}{\textbf{Llama-3.2-3B-Instruct}} & Dom & $82.4^{*}$ & 79.1 & 72.1 & 59.2 & 44.9 \\
& Non-dom & 4.6 & 5.3 & 5.3 & 5.4 & 5.4 \\
\midrule
\multirow{2}{*}{\textbf{Mistral-7B-Instruct-v0.3}} & Dom & 81.1 & 80.7 & 78.7 & 72.5 & 56.4 \\
& Non-dom & 3.6 & 4.2 & 4.9 & 4.4 & 4.5 \\
\midrule
\multirow{2}{*}{\textbf{Qwen2.5-7B-Instruct}} & Dom & 81.9 & 82.4 & 76.9 & 63.7 & 49.3 \\
& Non-dom & 6.1 & 5.7 & 5.4 & 5.6 & 5.5 \\
\midrule
\multirow{2}{*}{\textbf{Qwen2.5-14B-Instruct}} & Dom & 74.3 & 66.4 & 56.6 & 44.9 & 34.7 \\
& Non-dom & 4.1 & 4.6 & 4.5 & 4.9 & 4.9 \\
\bottomrule
\end{tabular}
}
\label{tab: predictive distribution of dominant fcs}
\end{table*}

\newpage
\section{Experiments Details}
\label{appendix: experiments details}
\subsection{Experiment Configurations.}
\label{appendix: Experiment Configurations.}
\paragraph{Baseline Configurations.}
As \mymodel is designed to optimize the decode phase, we forgo any KV cache optimizations during prefilling for all methods under evaluation. This experimental design isolates the performance impact of decode-stage acceleration, ensuring that our comparisons are direct and fair. For all baselines, we adopted configurations that are either standard in their original papers or represent a fair and strong setup for comparison.

\begin{itemize}[leftmargin=*, topsep=2pt, itemsep=2pt, parsep=0pt]
    \item \textbf{Oracle}: serves as an oracle baseline to demonstrate the upper-bound performance of Top-k sparse attention. This method operates under the ideal assumption that the k most important KV tokens for each query can be identified perfectly and at no computational cost. Consequently, a given token budget directly corresponds to this optimal Top-k set.
    \item \textbf{Stream}~\citep{xiao2024efficient}: 
    This method is based on the "attention sink" phenomenon, preserving a fixed number of initial tokens and a sliding window of recent tokens. Following its standard setup, we set the initial "start\_size" to 8 and the "recent\_size" to "budget - 8".
    \item \textbf{SnapKV}~\citep{li2024snapkv}: 
    SnapKV estimates token importance based on accumulated attention scores within a observation window during prefilling. We adopted its "maxpool" strategy with a window size of 32 and a kernel size of 7. As its original design performs a one-time filtering, it is not directly suited for long-generation tasks. We therefore adapted it, following the methodology in~\citep{cai2025r}, by re-applying the filtering mechanism every $n$ generated tokens.
    \item \textbf{Quest}~\citep{tang2024quest}: 
    Quest organizes the KV cache into pages and retrieves them based on a coarse-grained query-page similarity. We set the page size to 16, a value reported as near-optimal, to balance the trade-off between retrieval granularity and overhead.
    \item \textbf{RKV}~\citep{cai2025r}: 
    RKV is a state-of-the-art method for reasoning tasks that also employs a retrieval mechanism. We set its core hyperparameter $\lambda$, which balances between recent and important tokens, to 0.1 as recommended for optimal performance.
\end{itemize}

\paragraph{\mymodel Configurations.}
Our configuration for \mymodel is designed for both effectiveness and practical efficiency. Unless otherwise specified, the following setup was used across all experiments.

\begin{itemize}[leftmargin=*, topsep=2pt, itemsep=2pt, parsep=0pt]
    \item \textbf{Dominant FC Identification:} 
    A core principle of \mymodel is that the set of dominant FCs is a universal, task-agnostic property of the model architecture itself. Consequently, these indices ($\mathcal{I}_{dom}$) can be determined via a highly efficient, one-time offline calibration. 
    For our \textbf{LongBench} experiments, this calibration was performed on just a single data sample from the Qasper dataset. We found this minimal setup to be remarkably robust, as the generated response provides sufficient signal to identify the dominant FCs. The universality of these calibrated indices is empirically validated by \mymodel's strong performance across diverse tasks, from summarization to code completion. For \textbf{Long-CoT reasoning}, a similar single-instance calibration was performed on a question from the MATH500 dataset.
    
    \item \textbf{Hyperparameter Settings:} 
    For architectural simplicity and to maximize computational parallelism, we employ a uniform configuration across all  heads and layers. The number of dominant FCs to retain, denoted as $N_{\text{tip}}$, was consistently set to 16. This choice represents a balance between preserving sufficient contextual information and maximizing computational. 
    \item  \textbf{Task Configurations:} We configured the maximum sequence length to 32k for the AIME24 benchmark, reflecting its higher reasoning complexity, and to 16k for MATH500. For the LongBench benchmark, we set the maximum prompt length to 127.5k for Llama3/Qwen2.5 series models  and 31.5k for Mistral-7B-Instruct-v0.2.
    \end{itemize}
\subsection{Benchmark Details}\label{appendix: benchmark_details}

\textbf{LongBench~\citep{bai2024longbench}} is a comprehensive, multi-task benchmark designed to evaluate the long-context understanding capabilities of Large Language~\citep{wang-etal-2024-unveiling,wang-etal-2025-uncertainty}. It comprises a diverse set of tasks, including single-document QA, multi-document QA, summarization, few-shot learning, synthetic tasks, and code completion. In our experiments, we report the average performance across all relevant tasks to provide a holistic measure of a model's ability to process and reason over extended contexts, with sequence lengths ranging from 4K to over 100K tokens.

\textbf{MATH500 ~\citep{hendrycks2021measuring}} is a challenging benchmark for evaluating mathematical reasoning. It consists of 12,500 problems sourced from high school math competitions, spanning subjects like Algebra, Geometry, Number Theory, and Precalculus. Each problem is accompanied by a step-by-step solution, making it highly suitable for assessing CoT reasoning capabilities. We utilize the MATH500 subset for our long-CoT generation experiments, where models must produce detailed reasoning chains to arrive at the final answer. 

\textbf{AIME~\citep{AIME2024}} represents a significant step-up in reasoning complexity compared to the MATH dataset. It consists of problems from the AIME competition, which are known for their non-routine, multi-step solutions requiring deep mathematical insight and creativity~\citep{li2025adacurl,dai2026harder}. These problems serve as a stress test for a model's most advanced reasoning and long-chain generation abilities~\citep{xiong2025hs}. Following standard practice, we evaluate performance using the pass@k metric, specifically reporting pass@1 based on 16 generated responses per question.

\textbf{C4}~\citep{2019t5} is a massive, general-domain English text dataset derived from the Common Crawl web scrape. The "clean" version is created by applying a series of heuristics to filter out boilerplate content, code, and offensive language, resulting in a high-quality, natural language corpus. 

\textbf{PG19}~\citep{rae2019compressivetransformerslongrangesequence} is a long-form text dataset derived from books in the Project Gutenberg library. It is specifically curated for evaluating long-range sequence modeling.  Each example in the dataset is a full book text, making it an ideal benchmark for assessing a model's ability to handle and maintain coherence over very long dependencies, often exceeding the context windows of LLMs.

\textbf{WikiText}\citep{merity2017pointer}
 is a large-scale language modeling corpus sourced from high-quality "Good" and "Featured" articles on Wikipedia. Unlike raw web text, WikiText is well-formatted, grammatically correct, and retains its original punctuation and case. It is split into training, validation, and test sets at the article level. 
\subsection{Evaluation Protocols}\label{appendix: Evaluation Metrics}
To provide a comprehensive and rigorous assessment of model performance, we employ a set of standard metrics tailored to each evaluation paradigm.

\paragraph{Long-Context Understanding (LongBench).}
For the diverse tasks within the LongBench~\citep{bai2024longbench}, we follow its official evaluation protocol. Specifically, we use:
\begin{itemize}[leftmargin=*, nosep]
    \item \textbf{f1 score} for question-answering tasks.
    \item \textbf{rouge\_score} for summarization tasks.
    \item \textbf{code\_sim\_score} for code completion tasks.
\end{itemize}
The final reported score for LongBench is the average performance across all constituent tasks.

\paragraph{Long-Sequence Modeling.}
To evaluate a model's ability to maintain generative fidelity over long dependencies, we use perplexity (PPL). Perplexity measures how well a probability model predicts a sample. For a sequence of tokens $W = (w_1, w_2, \dots, w_N)$, PPL is defined as the exponential of the average negative log-likelihood in Equation~\ref{eq:perplexity}. A lower PPL indicates a better model, as it signifies higher confidence and accuracy in predicting the next token.
\begin{equation}
\label{eq:perplexity}
\text{PPL}(W) = \exp\left( -\frac{1}{N} \sum_{i=1}^{N} \log P(w_i | w_{<i}) \right)
\end{equation}

\paragraph{Long CoT Reasoning.}
For complex mathematical reasoning tasks such as MATH500 and AIME2024, we evaluate the model's performance in a long-generation setting. This paradigm is distinct from conventional long-context understanding tasks. Instead of processing a long static input, the model must maintain logical coherence and track thought traces across an extended, auto-regressive generation process to produce the correct final answer. Performance is reported as pass@1~\citep{dai2026harder}.
\begin{itemize}[leftmargin=*, nosep]
    \item For MATH500, we report pass@1, where a single generation is sampled for each problem.
    \item For AIME2024, which features more challenging problems, we also report pass@1, but the result is determined by checking if at least one correct answer exists within $k=16$ independent generations for each question. This sampling strategy is standard for estimating performance on complex reasoning benchmarks.
\end{itemize}
\begin{figure}[h]
    \centering
    \includegraphics[width=0.83\linewidth]{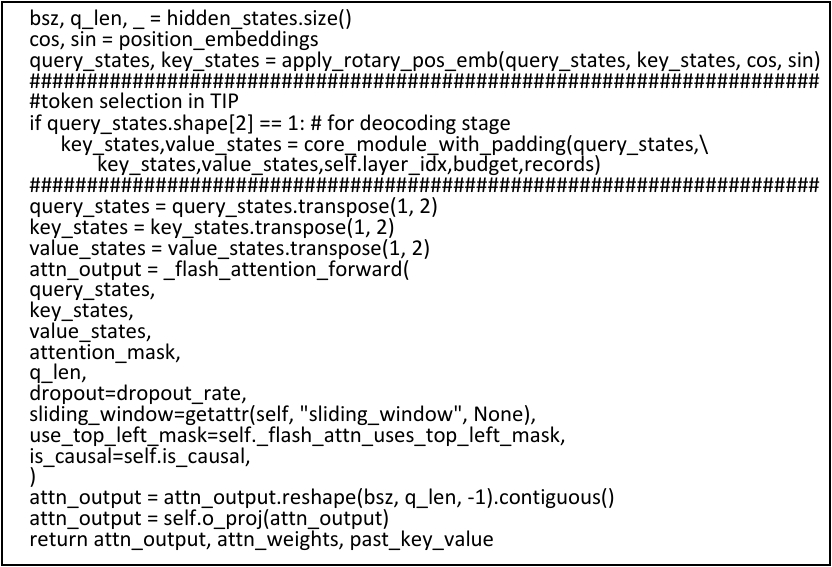}
    \caption{The \mymodel Pipeline: An Efficient, FlashAttention-Compatible Approach. The algorithm details our two-stage process. A key design feature is that the FAC stage seamlessly integrates with the standard FlashAttention API, leveraging its performance while enabling sparse computation.}
    \label{fig: appendix_pesudo_code}
\end{figure}
\subsection{Implement Details}\label{appendix: Implement Details}
\paragraph{Implementation Details}
Our implementation of \mymodel is built upon the HuggingFace Transformers library~\citep{wolf-etal-2020-transformers}. We employ a non-invasive monkey patching approach to integrate our logic. Specifically, we intercept the forward pass of the FlashAttention2 class within the model's modeling.py file.
The core of our method resides in two components. First, leveraging the universal nature of dominant FCs, their pre-computed indices are stored in a globally accessible dictionary, shared across all layers and heads. Second, the Token Importance Prediction (TIP) logic, which performs the critical token selection, is encapsulated within our core\_module\_with\_padding function.
A key advantage of our design is its simplicity and minimal intrusion. The integration requires inserting just a single line of code, the token selection logic, into the original attention function, making \mymodel easy to deploy and adapt. This minimal intrusion makes \mymodel highly portable and easy to adapt. The corresponding pseudocode is provided in Figure~\ref{fig: appendix_pesudo_code}.

\section{Additional Experimental Results}
\subsection{Performance Analysis on different budgets}
\label{appendix: Performance Analysis on different budgets}
\begin{figure}[h!]
    \centering
        \centering
        \includegraphics[width=0.99\textwidth]{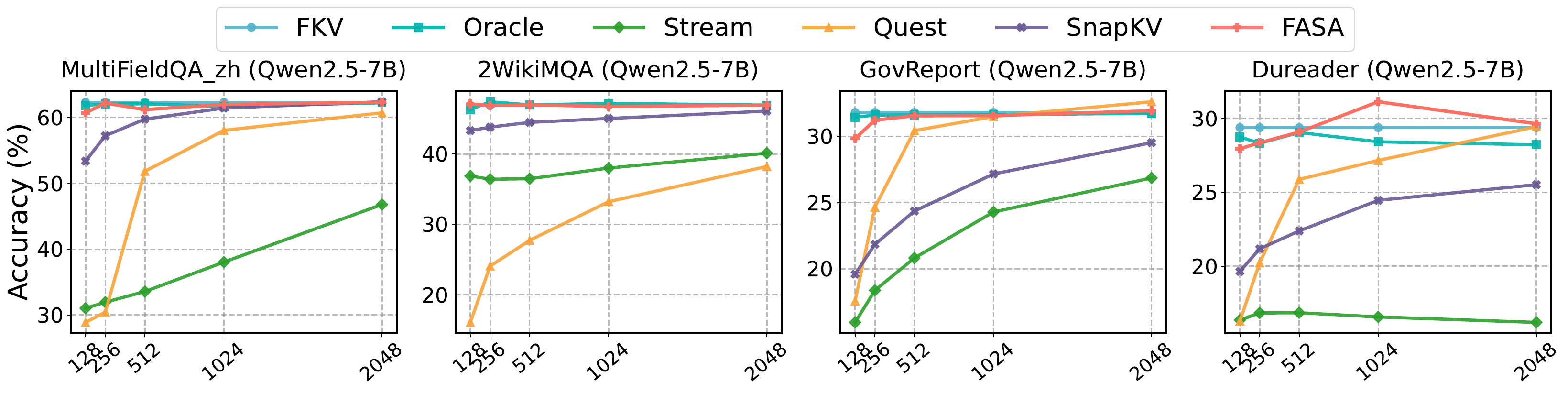}
    \vspace{-0.25cm}
    \caption{\small \mymodel on Qwen2.5-7B-Instruct under various token budgets ($N_{\text{tip}}=16$).}
    \label{fig: various budgets qwen 7b}
    \vspace{-0.3cm}
\end{figure}
\begin{figure}[h!]
    \centering
        \centering
        \includegraphics[width=0.9\textwidth]{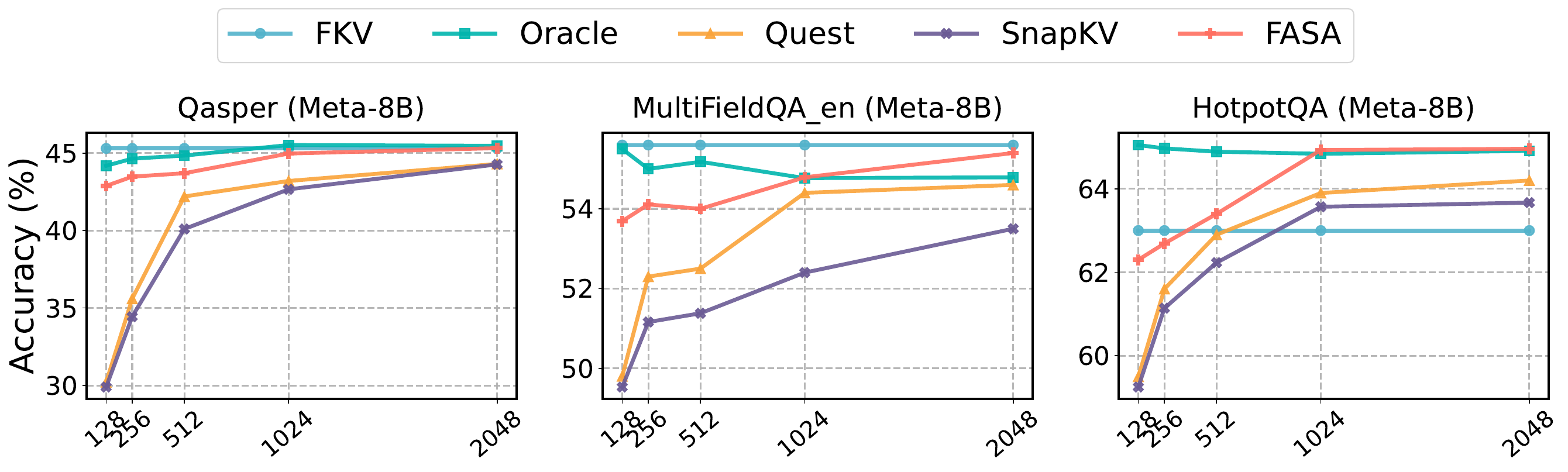}
    \vspace{-0.25cm}
    \caption{\small \mymodel on Meta-3.1-Llama-8B-Instruct under various token budgets ($N_{\text{tip}}=16$).}
    \label{fig: various budgets qwen 7b}
    \vspace{-0.3cm}
\end{figure}
\begin{figure}[h!]
    \centering
    \includegraphics[width=0.5\linewidth]{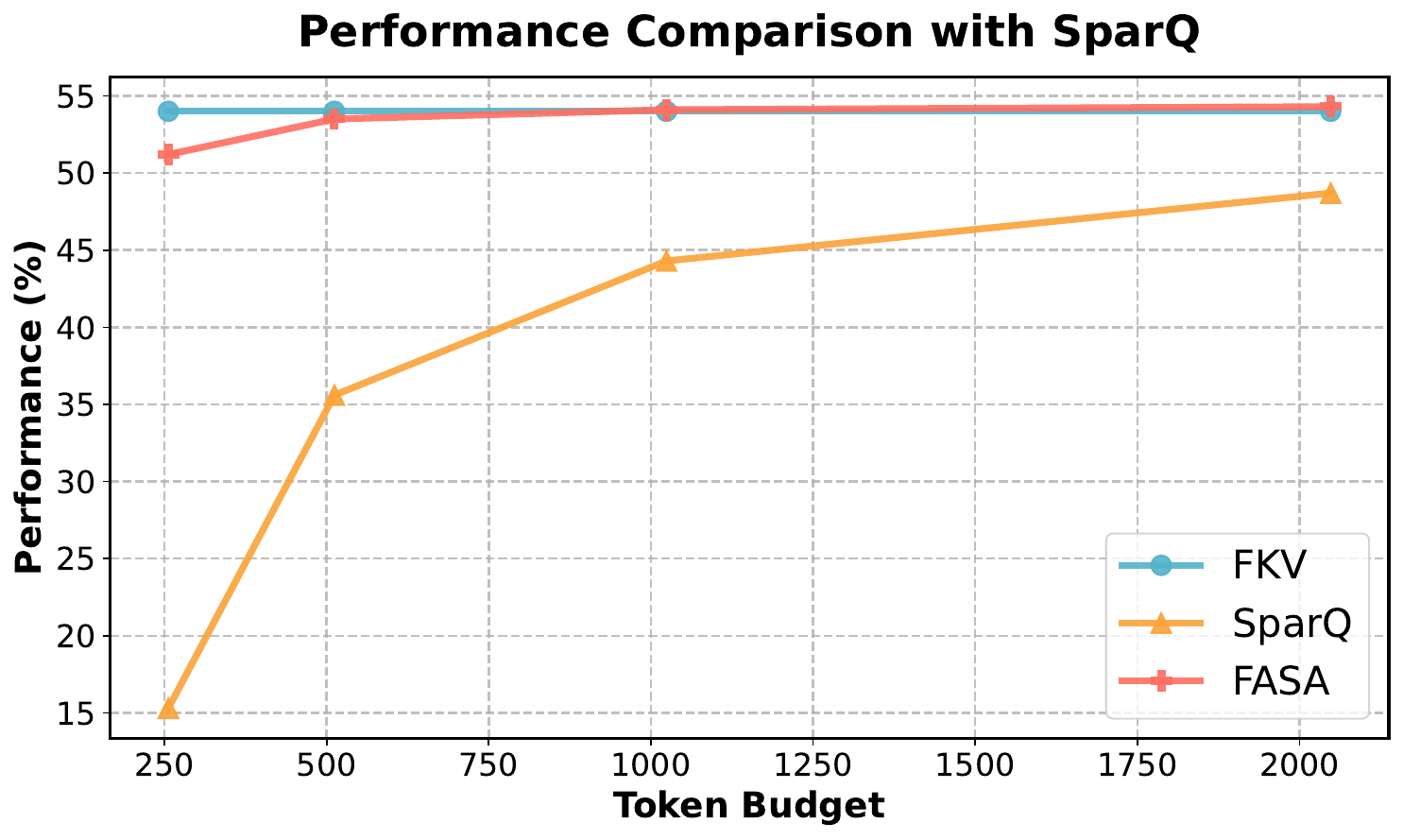}
    \caption{Comparision with SparQ on LongBench.}
    \label{fig: comparision with sparq}
\end{figure}

\paragraph{Comparison with Low-Rank Methods}
A closely related work to \mymodel is SparQ~\citep{ribar2024sparq}, which also performs a form of dimension selection. SparQ operates on the heuristic that high-magnitude dimensions in a query vector are the most indicative of importance, and thus selects corresponding key dimensions as a proxy for token prediction. However, as our experiments in Figure~\ref{fig: comparision with sparq} demonstrate, this heuristic proves to be a poor substitute for true contextual awareness. Under a constrained budget of 256 tokens, SparQ's performance collapses, indicating its inability to reliably identify critical tokens based solely on query magnitudes.
Furthermore, from an efficiency standpoint, SparQ incurs significant overhead as it must re-evaluate high-magnitude dimensions for every new query. In stark contrast, \mymodel leverages a one-time, offline calibration, making its per-token inference cost substantially lower.

% The end-to-end latency of \mymodel is primarily constrained by two factors: complex indexing operations within the KV-cache. While we have employed custom Triton kernels adapted from SparQ~\cite{ribar2024sparq}, a discernible gap between theoretical and wall-clock speedups remains.  exploring fine-grained kernel fusion and other low-level hardware optimizations will be a key focus of our future work.
% \section{Algorithms}
% \begin{algorithm}[h]
%    \caption{Offline Calibration Algorithm}
% \KwIn{ empty dicts $M$ for FC CA  scores savings, calibration datasets $\Omega$, Empty dict $R$ recoding average CA, final dominant FCs $F$}
% \KwOut{dominant FC recoreds $F$}
% \tcp{Stage 1: collecting CA scores for each FC}
% \For{each example in $\Omega$}{
% \For{each token generation step $t$}{
%   \For{each layer $l$}{
%     \For{each head $h$}{
%         Generating full attention scores $\alpha_{l,h}(\textbf{q}_t,\mathbf{K}_{1:t})$ as in Equation ~\ref{equation of full attentions scores}\;
%         \For{each FC $i$}{
%         Generating single-FC scores $\alpha_{l,h}^{(i)}$\;
%         Compute CA scores $\text{CA}_{\mathcal{K}}^{l,h,i}$ in Equation~\ref{eq:ca_k}\;
%         Saving CA into records $M[l][h][i]$.append($\text{CA}_{\mathcal{K}}^{l,h,i}$);\
%     }
%   }

% }}

% \tcp{Stage two: selecting the dominant FCs from $M$}
% Computes the average CA scores,$R=torch.mean(M[l,h,i],dim=-1)$\;
% dominant FCs are selected by $F=torch.topk(R,dim=-1)$ ;\
% }
% \label{Offline Calibration Algorithm1}
% \end{algorithm}

\section{Discussion on FASA}
\label{appendix: fasa-realted}
\subsection{Variants of FASA}
\label{varian of fasa}
\paragraph{FASA-M (Memory-Optimized)}
The memory-optimized variant, \mymodel-M, is specifically engineered for scenarios with constrained GPU memory, such as consumer-grade hardware. As detailed in Algorithm~\ref{alg:fasa-m}, its core strategy is to minimize the on-GPU memory footprint by strategically keeping only the most essential data on the GPU. 

Specifically, only the dominant parts of the Key cache ($C_{key}^{dom}$), which are required for the initial token importance prediction, are retained in GPU memory. The non-dominant parts of the Key cache ($C_{key}^{nondom}$) and the entire Value cache ($C_{val}$) are offloaded to and managed in the much larger CPU memory. During the Focused Attention Computation (FAC) stage, once the critical token indices ($\mathcal{T}_t$) are identified, only the small, required subsets of the non-dominant key and value caches are transferred from the CPU to the GPU for the final attention calculation. This "just-in-time" data transfer ensures that the GPU memory is primarily occupied by the most critical components, leading to substantial memory savings.

\paragraph{Memory Footprint Analysis}
The GPU memory footprint of the KV cache in \mymodel-M can be formulated as follows. Let $L$ be the total sequence length, $b$ the token budget, $d$ the model's hidden dimension, and $N_{layers}$ the number of layers. Let $d_{dom}$ be the dimension of the dominant FCs and $d_{nondom}$ be the dimension of the non-dominant FCs ($d = d_{dom} + d_{nondom}$). The memory occupied by the KV cache on the GPU is:
\begin{equation}
\label{eq:memory_fasa_m}
\text{Mem}_{\text{GPU}} \approx N_{layers} \times \left( \underbrace{L \times d_{dom}}_{\text{Dominant Keys}} + \underbrace{b \times d_{nondom}}_{\text{Non-dominant Keys}} + \underbrace{b \times d}_{\text{Values}} \right) \times \text{bytes\_per\_param}
\end{equation}
Compared to a full KV cache, which occupies $N_{layers} \times L \times 2d \times \text{bytes\_per\_param}$, \mymodel-M significantly reduces the memory burden, especially when the non-dominant and value components constitute a large portion of the cache. For instance, if $d_{dom}$ is 25\% of $d$ and the budget $b$ is 10\% of $L$, the memory savings can be substantial, approaching an 8$\times$ reduction in typical configurations.
\subsection{Design Choices}\label{appendix: design choices}

\begin{itemize}[leftmargin=*, nosep] 
    \item \textbf{On the Role of FC-Scores: A Proxy for Ranking, Not a Substitute for Attention.}
A crucial design principle we validated is that our FC-based scores ($\mathbf{S}_t^{l,h}$)  are not calibrated to function as direct attention weights. Although they provide a remarkably accurate relative ranking of token importance, their direct substitution for attention probabilities leads to a catastrophic performance degradation. This reveals their fundamental role as a selector—a mechanism to identify salient tokens rather than an approximator of the final attention distribution.

    % \item \textbf{Can individual dimensions serve as selection units?}
    \item \textbf{On the Indivisibility of Frequency Chunks.}
We investigated whether individual dimensions could serve as selection units, and the answer is a definitive no. A pipeline based on selecting "dominant dimensions" suffers a catastrophic performance degradation. This empirically validates that the Frequency Chunk (FC) is an indivisible functional unit for this process. This principle is not coincidental but is a direct corollary of RoPE's core mechanism, which encodes position by applying rotations to coupled pairs of dimensions. Disrupting these pairs severs the positional encoding, leading to model failure.
\end{itemize}
In summary, these two findings underscore two core design principles of \mymodel. First, an efficient proxy for token importance does not necessarily serve as a valid substitute for attention weights. Second, any optimization for RoPE-based models must respect the inherent coupling of dimension pairs, treating the Frequency Chunk as an indivisible functional unit.

\subsection{Algorithm on FASA}
\label{appendix: Algorithm on fasa}
See the algorithm of offline calibration in Algorithm ~\ref{alg:offline_calibration}; see the algorithm of \mymodel-M in Algorithm~\ref{alg:fasa-m}.
\begin{algorithm}[h!]
\caption{Offline Calibration for Dominant FCs}
\label{alg:offline_calibration}
\DontPrintSemicolon
\SetKwComment{Comment}{// }{}

\KwIn{A calibration dataset $\Omega$; number of dominant FCs to select $k$.}
\KwOut{The set of dominant FC indices, $\mathcal{I}_{dom}$.}

\BlankLine
\Comment{Stage 1: Collect Contextual Agreement (CA) scores}
Initialize an empty map $M$ to store CA scores for each $(l, h, i)$ triplet\;
\ForEach{example in $\Omega$}{
    \ForEach{token generation step $t$}{
        \ForEach{layer $l$}{
            \ForEach{head $h$}{
                Compute full attention scores $\boldsymbol{\alpha}_{l,h}(\mathbf{q}_t, \mathbf{K}_{1:t})$\;
                \ForEach{FC index $i$}{
                    Compute single-FC scores $\boldsymbol{\alpha}_{l,h}^{(i)}(\mathbf{q}_t, \mathbf{K}_{1:t})$\;
                    Calculate the CA score $\text{CA}_{\mathcal{K}}^{l,h,i}$ using Eq.~\ref{eq:ca_k}\;
                    Store $\text{CA}_{\mathcal{K}}^{l,h,i}$ in $M[l][h][i]$\;
                }
            }
        }
    }
}

\BlankLine
\Comment{Stage 2: Select Dominant FCs}
Initialize an empty map $\overline{M}$ for mean CA scores\;
\ForEach{$(l, h, i)$ in $M$}{
    $\overline{M}[l][h][i] \leftarrow \text{Mean}(M[l][h][i])$\;
}
$\mathcal{I}_{dom} \leftarrow \text{TopK-Indices}(\overline{M}, k)$ \Comment{Select top-k indices based on $\overline{\text{CA}}$}

\KwRet{$\mathcal{I}_{dom}$}\;
\end{algorithm}

\begin{algorithm}[h!]
\caption{Inference with \mymodel-M (Memory-Optimized Variant)}
\label{alg:fasa-m}
\DontPrintSemicolon
\SetKwComment{Comment}{// }{}

\KwIn{
    Current query $\mathbf{q}_t$; Current key $\mathbf{k}_t$; Current value $\mathbf{v}_t$\;
    Dominant FC indices $\mathcal{I}_{dom}$\;
    Token budget $b$\;
    Past KV cache: $C_{key}^{dom}$ (GPU), $C_{key}^{nondom}$ (CPU), $C_{val}$ (CPU)\;
}
\KwOut{
    Next hidden state $\mathbf{h}_{t+1}$\;
    Updated KV cache: $C_{key}^{dom}$, $C_{key}^{nondom}$, $C_{val}$\;
}

\BlankLine
\Comment{Stage 1: Token Importance Prediction (TIP)}
\Comment{Split key by dominant FCs}
$\mathbf{k}_t^{dom}, \mathbf{k}_t^{nondom} \leftarrow \text{Split}(\mathbf{k}_t, \mathcal{I}_{dom})$\;
\Comment{Select corresponding query dimensions}
$\mathbf{q}_t^{dom} \leftarrow \text{Select}(\mathbf{q}_t, \mathcal{I}_{dom})$\;

$K_{1:t}^{dom} \leftarrow \text{UpdateCache}(C_{key}^{dom}, \mathbf{k}_t^{dom})$\;
\Comment{Approximate scores using dominant parts}
$\hat{\mathbf{S}}_t \leftarrow \mathbf{q}_t^{dom} (K_{1:t}^{dom})^\top$\;
\Comment{Identify indices of $b$ most salient tokens}
$\mathcal{T}_t \leftarrow \text{TopK-Indices}(\hat{\mathbf{S}}_t, b)$\;

\BlankLine
\Comment{Stage 2: Focused Attention Computation (FAC)}
\Comment{Select dominant key parts on GPU}
$K_{\mathcal{T}_t}^{dom} \leftarrow \text{SelectTokens}(K_{1:t}^{dom}, \mathcal{T}_t)$\;

\Comment{Update non-dominant cache on CPU}
$C_{key}^{nondom} \leftarrow \text{UpdateCache}(C_{key}^{nondom}, \mathbf{k}_t^{nondom})$\;
$K_{1:t}^{nondom} \leftarrow \text{LoadFromCPU}(C_{key}^{nondom})$\;
\Comment{Select non-dominant key parts on CPU}
$K_{\mathcal{T}_t}^{nondom} \leftarrow \text{SelectTokens}(K_{1:t}^{nondom}, \mathcal{T}_t)$\;

\Comment{Update value cache on CPU}
$C_{val} \leftarrow \text{UpdateCache}(C_{val}, \mathbf{v}_t)$\;
$V_{1:t} \leftarrow \text{LoadFromCPU}(C_{val})$\;
\Comment{Select values on CPU}
$V_{\mathcal{T}_t} \leftarrow \text{SelectTokens}(V_{1:t}, \mathcal{T}_t)$\;

\BlankLine
\Comment{Offload required non-dominant keys to GPU}
$K_{\mathcal{T}_t}^{nondom} \leftarrow \text{TransferToGPU}(K_{\mathcal{T}_t}^{nondom})$\;
\Comment{Offload required values to GPU}
$V_{\mathcal{T}_t} \leftarrow \text{TransferToGPU}(V_{\mathcal{T}_t})$\;

\Comment{Reconstruct full keys for selected tokens}
$K_{\mathcal{T}_t} \leftarrow \text{Combine}(K_{\mathcal{T}_t}^{dom}, K_{\mathcal{T}_t}^{nondom}, \mathcal{I}_{dom})$\;

\Comment{Compute full attention on the subset}
$\boldsymbol{\alpha}_{\text{fac}} \leftarrow \text{Softmax}(\mathbf{q}_t K_{\mathcal{T}_t}^\top / \sqrt{d_k})$\;
$\mathbf{h}_{t+1} \leftarrow W_O (\boldsymbol{\alpha}_{\text{fac}} V_{\mathcal{T}_t})$\;

\BlankLine
\KwRet{$\mathbf{h}_{t+1}$ and updated caches}\;
\end{algorithm}
% \begin{algorithm}[h!]
% \caption{Implementation on \mymodel-M}
% \label{alg: fasa-m}
% \KwIn{At generation step $t$, $C_{key}^{dom}$, $C_{key}^{nondom}$, $C_{value}^{full}$; $\mathbf{q}_t \in \mathbb{R}^d$, $\mathbf{k}_t \in \mathbb{R}^{ d}$, $\mathcal{I}_{dom}$, token budget $b$}
% \tcp{step1: split the key into two parts}
% $\mathbf{k}_t^{dom}$,$\mathbf{k}_t^{nondom}$ = split($\mathbf{k}_t$,$\mathcal{I}_{dom}$)\;
% $K_{1:t}^{dom}$ = updatedKV($C_{key}^{dom}$,$\textbf{k}_t^{dom}$)\;
% $K_{1:t}^{nondom}$ = updatedKV($C_{key}^{nondom}$,$\textbf{k}_t^{nondom}$)\;
% $V_{1:t}^{full}$ = updatedKV($C_{value}^{full}$,$\textbf{v}_t$)\;
% $\mathbf{q}_t^{dom}$ = select($\mathbf{q}_t$,$\mathcal{I}_{dom}$)\;
% $\textbf{S}_t$ = $\textbf{q}_t^{dom}{K_{1:t}^{dom}}^T$\;
% $\mathcal{T}_t$ = Topk-I($\textbf{S}_t$,$N_{fac}$)\;
% \tcp{stage2: FAC}
% $K_{\mathcal{T}_t}^{dom}$=select($K_{{1:t}}^{dom}$,$\mathcal{T}_t$)\;
% $K_{\mathcal{T}_t}^{nondom}$=select($K_{{1:t}}^{nondom}$,$\mathcal{T}_t$)\;
% $V_{\mathcal{T}_t}$= select($V_{1:t}^{full}$,$\mathcal{T}_t$)\;
% $K_{\mathcal{T}_t}^{nondom}$=$K_{\mathcal{T}_t}^{nondom}$.to(cuda)\;
% $K_{\mathcal{T}_t}$=concate($K_{\mathcal{T}_t}^{dom}$,$K_{\mathcal{T}_t}^{nondom}$, $\mathcal{I}_{dom}$)\;
% $\alpha_{fac}$=$\textbf{q}_tK_{\mathcal{T}_t}^T$
% $h_{t+1}=W_O(\alpha_{fac}V_{\mathcal{T}_t})$
% \end{algorithm}

\section{LLM Usage}
During the preparation of this manuscript, we utilized the AI-based language model ChatGPT, developed by OpenAI. Its use was strictly limited to language refinement, including grammar correction, stylistic enhancement, and rephrasing for clarity. All scientific concepts, experimental designs, data analyses, and conclusions presented herein are the original work of the authors and were conceived and executed without any substantive contribution from the language model.

\end{document}